\pdfoutput=1
\documentclass[aps,twocolumn,superscriptaddress, notitlepage,longbibliography]{revtex4-1}

\usepackage{amsmath} 
\usepackage{amsthm} 
\usepackage{amssymb}	
\usepackage{physics}
\usepackage[pdftex]{graphicx}
\usepackage{hyperref} 
\hypersetup{
    colorlinks,
    citecolor=blue,
    filecolor=black,
    linkcolor=red,
    urlcolor=blue
}
\usepackage{xcolor}
\usepackage{soul}
\usepackage{url}

\newcommand{\E}[1][]{\mathrm{E}_{#1}}
\newcommand{\Var}[1][]{\mathrm{Var}_{#1}}
\newcommand{\Cov}[1][]{\mathrm{Cov}_{#1}}
\newcommand{\Bias} {\mathrm{Bias}}
\newcommand{\test}{\mathcal{E}_{\mathrm{test}}}
\newcommand{\train}{\mathcal{E}_{\mathrm{train}}}

\newcommand{\etest}{\expval{\mathcal{E}_{\mathrm{test}}}}
\newcommand{\etrain}{\expval{\mathcal{E}_{\mathrm{train}}}}

\newcommand{\vbx}{\vec{\mathbf{x}}}
\newcommand{\vby}{\vec{\mathbf{y}}}
\newcommand{\vbz}{\vec{\mathbf{z}}}

\newcommand{\vbbeta}{\vec{\boldsymbol{\beta}}}
\newcommand{\vbeps}{\vec{\boldsymbol{\varepsilon}}}
\newcommand{\vbeta}{\vec{\boldsymbol{\eta}}}
\newcommand{\hbbeta}{\hat{\boldsymbol{\beta}}}
\newcommand{\hbw}{\hat{\mathbf{w}}}
\newcommand{\hby}{\hat{\mathbf{y}}}

\newcommand{\hbu}{\hat{\mathbf{u}}}
\newcommand{\hbh}{\hat{\mathbf{h}}}

\newcommand{\qqc}{,\qquad}

\begin{document}

\title{Memorizing without overfitting:\\  Bias, variance, and interpolation in over-parameterized models}

\author{Jason W. Rocks}
\affiliation{Department of Physics, Boston University, Boston, Massachusetts 02215, USA}

\author{Pankaj Mehta}
\affiliation{Department of Physics, Boston University, Boston, Massachusetts 02215, USA}
\affiliation{Faculty of Computing and Data Sciences, Boston University, Boston, Massachusetts 02215, USA}

\begin{abstract}
The bias-variance trade-off is a central concept in supervised learning. In classical statistics, increasing the complexity of a model (e.g., number of parameters) reduces bias but also increases variance. Until recently, it was commonly believed that optimal performance is achieved at intermediate model complexities which strike a balance between bias and variance. Modern Deep Learning methods flout this dogma, achieving state-of-the-art performance using ``over-parameterized models'' where the number of fit parameters is large enough to perfectly fit the training data. As a result, understanding bias and variance in over-parameterized models has emerged as a fundamental problem in machine learning. Here, we use methods from statistical physics to derive analytic expressions for bias and variance in two minimal models of over-parameterization (linear regression and two-layer neural networks with nonlinear data distributions), allowing us to disentangle properties stemming from the model architecture and random sampling of data.
In both models, increasing the number of fit parameters leads to a phase transition where the training error goes to zero and the test error diverges as a result of the variance (while the bias remains finite).
Beyond this threshold, the test error of the two-layer neural network decreases due to a monotonic decrease in \emph{both} the bias and variance in contrast with the classical bias-variance trade-off.
We also show that in contrast with classical intuition, over-parameterized models can overfit even in the absence of noise and exhibit bias even if the student and teacher models match. We synthesize these results to construct a holistic understanding of generalization error and the bias-variance trade-off in over-parameterized models and relate our results to random matrix theory. 
\end{abstract}

\maketitle

\section{Introduction}

Machine Learning (ML) is one of the most exciting and fastest-growing areas of modern research and application.  Over the last decade, we have witnessed  incredible 
progress in our ability to learn statistical relationships from large data sets and make accurate predictions. Modern ML techniques have now made it possible to 
automate tasks such as speech recognition, language translation, and visual object recognition, with wide-ranging implications for fields such as genomics, physics, and even 
mathematics. These techniques  -- in particular the Deep Learning methods that underlie many of the most prominent recent advancements  -- are  especially successful at tasks that can be recast as supervised learning problems~\cite{Lecun2015}. In supervised learning, the goal is to learn statistical relationships from labeled data (e.g., a collection of pictures labeled as 
containing a cat or not containing a cat). Common examples of supervised learning tasks include classification and regression. 

A fundamental concept in supervised learning is the bias-variance trade-off.
In general, the out-of-sample, generalization, or test error, of a statistical model can be decomposed into three sources: bias (errors resulting from erroneous assumptions which can hamper a statistical model's ability to fully express the patterns hidden in the data),
variance (errors arising from over-sensitivity to the particular choice of training set), 
and noise. 
This bias-variance decomposition  provides a natural intuition for understanding  how complex a model must be in order to make accurate predictions on unseen data. 
As model complexity (e.g., the number of fit parameters) increases, 
bias decreases as a result of the model becoming more expressive and better able to 
capture complicated statistical relationships in the underlying data distribution.
However, a more complex model may also exhibit higher variance as it begins to overfit, becoming less constrained and therefore more sensitive to the quirks of the training set (e.g., noise) that do not generalize to other data sets.
This trade-off is reflected in the generalization error in the form of a classical  ``U-shaped'' curve: the test error first decreases with model complexity until it reaches a minimum before increasing dramatically as the model overfits the training data.
For this reason, it was commonly believed until recently that optimal performance is  achieved at intermediate model complexities which strike a balance between bias (underfitting) and variance (overfitting).

Modern Deep Learning methods defy this understanding, achieving state-of-the-art performance using ``over-parameterized models'' 
where the number of fit parameters is so large -- often orders of magnitude larger than the number of data points~\cite{Canziani2016} -- that one would expect a model's accuracy to be overwhelmed by overfitting.
In fact, empirical experiments show that
convolutional networks commonly used in image classification are so overly expressive that they can easily fit training data with randomized labels, or even images generated from random noise, with almost perfect accuracy~\cite{Zhang2017}. 
Despite the apparent risks of overfitting, these models seem to perform at least as well as, if not better than, traditional statistical models. 
As a result, modern best practices in Deep Learning recommend using highly over-parameterized models that are expressive enough to achieve zero error on the training data~\cite{Mehta2019}.

\begin{figure*}[t!]
\centering
\includegraphics[width=1.0\linewidth]{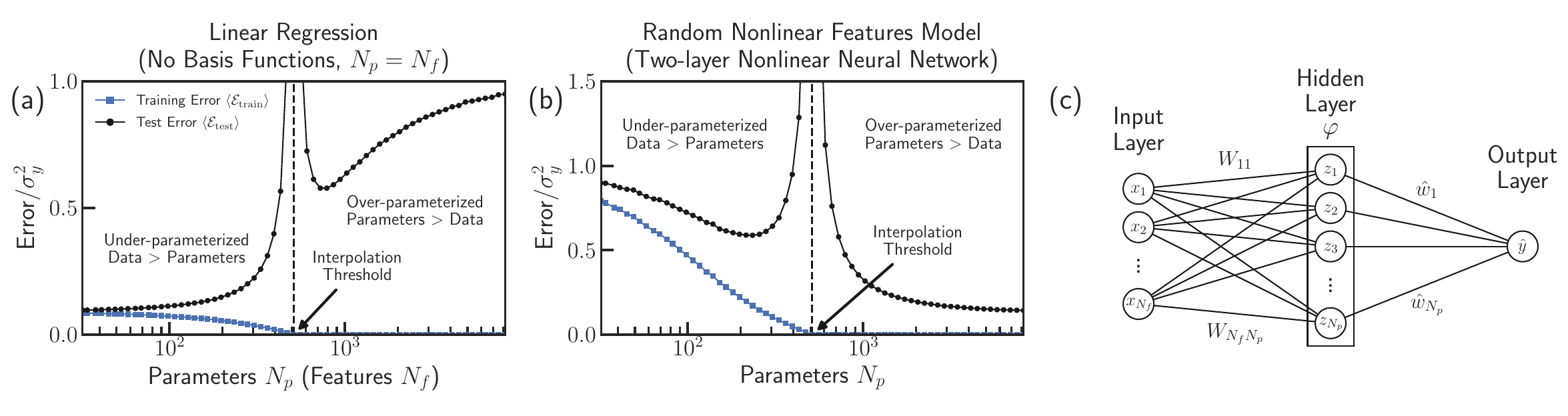} 
\caption{
{\bf Double-descent phenomenon.} {\bf (a)-(b)} Examples of the average training error (blue squares) and test error (black circles) for two different models calculated via numerical simulations.
In both models, the test error diverges when the training error reaches zero at the interpolation threshold,
located where the number of parameters $N_p$ matches the number of points in the training data set $M$ (indicated by a black dashed vertical line).
{\bf (a)} In  linear regression without basis functions, the number of features in the data $N_f$ matches the number of fit parameters $N_p$. 
{\bf(b)} The random nonlinear features model (two-layer neural network where the parameters of the middle layer are random but fixed) decouples the number of features $N_f$ and the number of fit parameters $N_p$  by incorporating an additional ``hidden layer'' and transforms the data using a nonlinear activation function (e.g., ReLU), resulting in the canonical double-descent behavior.
{\bf(c)} Schematic of the model architecture for the random nonlinear features model.
Numerical results are shown for a linear teacher model $y(\vbx) = \vbx\cdot \vbbeta + \varepsilon$, a signal-to-noise ratio of $\sigma_\beta^2\sigma_X^2 / \sigma_\varepsilon^2 = 10$, and a small regularization parameter of $\lambda = 10^{-6}$.
The $y$-axes have been scaled by the variance of the training set labels $\sigma_y^2 = \sigma_\beta^2\sigma_X^2 + \sigma_\varepsilon^2$.
Each point is averaged over at least $1000$ independent simulations trained on $M=512$ data points with small error bars indicating the error on the mean.
In (b), there are less features than data points $N_f = M/4$.
See Sec.~\ref{sec:setup} for precise definitions and Sec.~S4 of Supplemental Material~\cite{supporting_info} for additional simulation details.
 }\label{fig:schematics}
\end{figure*}

Clearly, the classical picture provided by the bias-variance trade-off is incomplete. 
Classical statistics largely focuses on under-parameterized models which are simple enough that they have a nonzero training error.
In contrast, Modern Deep Learning methods push model complexity past the \textit{interpolation threshold},
the point at which the training error reaches zero~\cite{Belkin2019, Geiger2019, Spigler2019, Geiger2020, Bahri2020}.
In the classical picture, approaching the interpolation threshold coincides with a large increase, or even divergence, in the test error via the variance.
However, numerical experiments suggest that the predictive performance of over-parameterized models is better described by ``double-descent'' curves which extend the classic U-shape past the interpolation threshold to account for over-parameterized models with zero training error~\cite{Belkin2019,Kobak2020,Loog2020}.
Surprisingly, if model complexity is increased past the interpolation threshold,  the test error
once again decreases, often resulting in over-parameterized models with even better out-of-sample performance than their under-parameterized counterparts [see Fig.~\ref{fig:schematics}(b)].

This double-descent behavior stands in stark contrast with the classical statistical intuition based on the bias-variance trade-off; both bias \textit{and} variance appear to decrease past the interpolation threshold.
Therefore, understanding this phenomenon requires generalizing the bias-variance decomposition to over-parameterized models.
More broadly, explaining the unexpected success of over-parameterized models represents a fundamental problem in ML and modern statistics.

\subsection{Relation to Previous Work}

In recent years, many attempts have been made to understand the origins of this double-descent behavior via numerical and/or analytical approaches.
While much of this work has relied on well constructed numerical
experiments on complex Deep Learning models~\cite{Zhang2017, Belkin2019,  Spigler2019, Nakkiran2020},
many theoretical studies have focused on a much simpler setting: the so-called ``lazy training'' regime.
Previously, it was observed that in the limit of an infinitely wide network,
the learning process appears to mimic that of an approximate kernel method in which the kernel used by the model to express the data -- the Neural Tangent Kernel (NTK) --  remains fixed~\cite{Jacot2018, Lee2019}.
This stands in contrast to the so-called ``feature training'' regime in which the kernel evolves over time as the model learns the most informative way to express the relationships in the data~\cite{Geiger2020b}.

Making use of the observation that the kernel remains approximately fixed in the lazy regime,
many analytical studies have derived training and test errors for neural networks where the top layer is trained, but the middle layer(s) remained fixed,
effectively reducing these models to linearized versions of regression or classificaiton with various types of nontrivial basis functions~\cite{Adlam2020, Advani2020, Ba2020, Barbier2019, Bartlett2020, Belkin2020, Bibas2019, Deng2019, DAscoli2020, DAscoli2020a, Derezinski2020, Dhifallah2020, Gerace2020, Hastie2019, Jacot2020, Kini2020, Lampinen2019, Li2020, Liang2020, Liang2020b, Liao2020, Lin2020, Mitra2019, Mei2019, Muthukumar2019, Nakkiran2019, Xu2019, Yang2020}.
Furthermore, some of these studies have considered nonlinear data distributions~\cite{Dhifallah2020, Mei2019}.
Importantly, such studies have typically combined a fixed-kernel approach with specific choices of loss functions (e.g., mean-squared error) to guarantee convexity,
while the loss landscapes of neural networks in practical settings are often highly non-convex~\cite{Dauphin2014}.
Despite being limited to the lazy regime and convex loss landscapes, the closed-form solutions for the training and test error obtained for these models exhibit the double-descent phenomenon,
demonstrating that many of the key features of more complex Deep Learning architectures can arise in much simpler settings.

A smaller subset of these studies have
also attempted to extend these calculations to compute the bias-variance decomposition~\cite{Adlam2020, Ba2020, DAscoli2020, DAscoli2020a, Derezinski2020, Hastie2019, Jacot2020, Li2020, Lin2020, Mei2019, Nakkiran2019, Yang2020}. 
However, this literature is rife with qualitative and quantitative disagreements, 
and as a result, a consensus has not formed regarding many of the basic properties of bias and variance in over-parameterized models.
Underlying these disagreements is the ubiquitous use of non-standard and varying definitions of bias and variance (see Sec.~\ref{sec:nonstandard} for an in-depth discussion).

For example, some studies consider a fixed design matrix for the training data set in the definitions of bias and variance, but not for the test set,
resulting in an effective mismatch in their data distributions~\cite{Adlam2020, Ba2020, Derezinski2020, Hastie2019, Jacot2020, Li2020, Mei2019}.
As a result, these studies do not even reproduce the classical bias-variance trade-off expected in the under-parameterized regime.

Meanwhile, other studies do not distinguish between sources of randomness stemming from the model architecture (e.g., due to initialization) and sampling of the training data set, inadvertently leading these analyses to derive the bias of ensemble models rather than the models actually under investigation~\cite{Adlam2020, DAscoli2020, DAscoli2020a, Jacot2020, Lin2020, Yang2020}. 
Consequently, these studies have found that in the absence of regularization, the bias reaches a minimum at the interpolation threshold and then remains constant into the over-parameterized regime.

In fact, of these studies, closed-form expressions using the standard definitions of bias and variance have only been obtained for the simple case of linear regression without basis functions and a linear data distribution~\cite{Nakkiran2019}. 
While this setting captures some qualitative aspects of the double-descent phenomenon [see Fig.~\ref{fig:schematics}(a)], 
it requires a one-to-one correspondence between features in the data and fit parameters and a perfect match between the data distribution and model architecture, making it difficult
to understand, if and how these results generalize to more complicated statistical models.

In line with previous studies, in this work, we also focus on the lazy regime with a convex loss landscape, considering two different linear models that emulate many properties of more complicated neural network architectures (see next section).
However, our approach differs in that we utilize the traditional definitions of bias and variance, allowing us to clear up much of the confusion surrounding the bias-variance decomposition in over-parameterized models.
In this way, we connect Modern Deep Learning to the statistical literature of the last century and in doing so, gain proper intuition for the origins of the double-descent phenomenon.

\subsection{Overview of Approach}

In this work, we use methods from statistical physics to derive analytic results for bias and variance in the over-parameterized regime for two minimal model architectures. 
These models, whose behavior is depicted in Fig.~\ref{fig:schematics}, are \emph{linear regression} (ridge regression without basis functions and in the limit where the regularization parameter goes to zero -- often called ``ridge-less regression'' in the statistics and ML literature) 
and the \emph{random nonlinear features model} (a two-layer neural network with an arbitrary nonlinear activation function where the top layer is trained and parameters for the intermediate layer are chosen to be random but fixed). 
We generate the data used to train both models using a non-linear ``teacher model'' where the labels are related to the features through a non-linear function (usually with additive noise). 
Using similar terminology, we often refer to the details of a model's architecture as the ``student model.''
Crucially, the differences between the two models we consider allow us to disentangle the effects of model architecture on bias and variance versus effects arising from randomly sampling the underlying data distribution.

Linear regression is one of the simplest models in which the test error diverges at the interpolation threshold but then decreases in the over-parameterized regime  [Fig.~\ref{fig:schematics}(a)]. 
Because this model uses the features in the data directly without modification (i.e., it lacks a hidden layer),
it provides evidence that the process of randomly sampling the data itself plays an integral part in the double-descent phenomena. 
The random nonlinear features model  [Figs.~\ref{fig:schematics}(b) and (c)] provides insight into the effects of filtering the features through an additional transformation,
in effect, changing the way the model views the data. 
This disconnect between features and their representations in the model is crucial for understanding bias and variance in more complex over-parameterized models.

To treat these models analytically, we make use of the ``zero-temperature cavity method'' which has a long history in the physics of disordered systems and statistical learning theory~\cite{engel2001statistical, Mezard2003, Ramezanali2015}.
In particular, our calculations follow the style of Ref.~\onlinecite{Mehta2019a}
and assume that the solutions can be described using a Replica-Symmetric Ansatz, which we confirm numerically.
Our analytic results are exact in the thermodynamic limit where the number of data points $M$, the number of features in the data $N_f$, and the number of fit parameters (hidden features) $N_p$ all tend towards infinity. 
Crucially, when taking this limit, the ratios between these three quantities are assumed to be fixed and finite, allowing us to ask how varying these ratios and other model properties (such as linear versus nonlinear activation functions) affect generalization. 
We confirm that all our analytic expressions agree extremely well with numerical results, even for relatively small system sizes ($N_f,N_p, M \sim 10-1000$).

\subsection{Summary of Major Results}

Before proceeding further, we briefly summarize our major results:
\begin{itemize}
\item We derive analytic expressions for the test (generalization) error, training error, bias, and variance for both models using the zero-temperature cavity method.
\item We show that both models exhibit a phase transition at an interpolation threshold to an interpolation regime where the training error is zero.
\item In the under-parameterized regime, we find that the variance diverges as it approaches the interpolation threshold, leading to extremely large generalization error, while the bias either remains constant (linear regression) or decreases monotonically in a classical bias-variance trade-off (random nonlinear features model).
\item In the over-parameterized regime, we find that the test error either decreases non-monotonically due to a decrease in variance and an increase in bias (linear regression) or decreases monotonically due to a monotonic decrease in both variance and bias, even in the absence of regularization (random nonlinear features model). 
\item We show that bias in over-parameterized models has two sources: error resulting from mismatches between the student and teacher models (i.e., the model is incapable of fully capturing the full data distribution) and incomplete sampling of the data's feature space. 
We show that as a result of the latter case, one can have nonzero bias even if the student and teacher models are identical. We also show that bias decreases in the interpolation regime only if the number of features in the data remains fixed. 
\item We show that biased models can overfit even in the absence of noise. In other words, biased models can interpret signal as noise. 
\item We show that the zero-temperature susceptibilities that appear in our cavity calculations measure the sensitivity of a fitted model to small perturbations.
We discuss how these susceptibilities can be used to identify phase transitions and are each related to different aspects of the double-descent phenomena. 
\item We combine these observations to provide a comprehensive intuitive explanation of the double-descent curves for test error observed in over-parameterized models, making connections to random matrix theory. 
In particular, we discuss how diverging test error stems from small eigenvalues in the Hessian,
corresponding to poorly sampled directions in the space of input features for linear regression or the space of hidden features for the random nonlinear features model.
\item We discuss why using the standard definitions of bias and variance are necessary to properly connect the double-descent phenomenon to the classical bias-variance trade-off.
\end{itemize}

\subsection{Organization of Paper}
In Sec.~\ref{sec:setup}, we start by providing the theoretical setup for both models and briefly summarize the methods we use to derive analytic expressions. In Sec.~\ref{sec:bvdefs}, we provide precise definitions of bias and variance, taking great care to distinguish between different sources of randomness.
In Sec.~\ref{sec:results}, we report our analytic results and compare them to numerical simulations. 
In  Sec.~\ref{sec:discuss}, we use these analytic expressions to understand how bias and variance generalize to over-parameterized models and also discuss the roles of the susceptibilities that arise as part of our cavity calculations.
Finally, in Sec.~\ref{sec:conclude}, we conclude and discuss the implications of our results for modern ML methods.

\section{Theoretical Setup}\label{sec:setup}

\subsection{Supervised Learning Task}

In this work, we consider data points $(y, \vbx)$, each consisting of a continuous label $y$ paired with a set of $N_f$ continuous features $\vbx$.
To distinguish the features in the data from those in the model, we refer to $\vbx$ as the ``input features.''
We frame the supervised learning task as follows: using the relationships learned from a training data set, construct a model to accurately predict the labels $y$ of new data points based on their input features $\vbx$.

\subsection{Data Distribution (Teacher Model)}\label{sec:teacher}

We assume that the relationship between the input features and labels (the data distribution or teacher model) can be expressed as 
\begin{align}
y(\vbx) &= y^*(\vbx; \vbbeta) + \varepsilon\label{eq:teacher}
\end{align}
where $\varepsilon$ is the label noise  and $y^*(\vbx; \vbbeta)$ is an unknown function representing the ``true'' labels.
This function takes the features as arguments 
and combines them with a set of $N_f$ ``ground truth'' parameters $\vbbeta$ 
which characterize the correlations between the features and labels.

We draw the input features for each data point independently and identically from a normal distribution with zero mean and variance $\sigma_X^2/N_f$.
Normalizing the variance by $N_f$ ensures that the magnitude of each feature vector is independent of the number of features and that a proper thermodynamic limit
exists.
Note that in this work, we consider features that do not contain noise.
We also choose each element of the ground truth parameters $\vbbeta$ and the label noise $\varepsilon$ to be independent of the input features and mutually independent from one another, drawn from normal distributions with zero mean and variances $\sigma_\beta^2$ and $\sigma_\varepsilon^2$, respectively.
 
In this work, we restrict ourselves to a teacher model of the form
\begin{equation}
y^*(\vbx;\vbbeta) =  \frac{\sigma_\beta\sigma_X}{\expval*{f'}} f\qty(\frac{\vbx\cdot\vbbeta}{\sigma_X\sigma_\beta} )\label{eq:true_labels}
\end{equation}
where the function $f$ is an arbitrary nonlinear function and ${\expval*{f'} = \frac{1}{\sqrt{2\pi}}\int_{-\infty}^\infty \dd he^{-\frac{h^2}{2}}f'(h)}$ is a normalization constant chosen for convenience with prime notation used to indicate a derivative (see Sec.~S1D of Supplemental Information~\cite{supporting_info}).
We place a factor of $1/(\sigma_X\sigma_\beta)$ inside the function $f$ so that its argument has unit variance,
while the pre-factor $\sigma_\beta\sigma_X$ ensures that $y^*$ reduces to a linear teacher model  ${y^*(\vbx) = \vbx\cdot\vbbeta}$ when $f(h) = h$. 
Furthermore, we assume both the labels and input features are centered so that $f$ has zero mean with respect to its argument.
While the results we report hold for a general $f$ of this form, all figures show numerical simulations for a linear teacher model unless otherwise specified.

\subsection{Model Architectures (Student Models)}

We consider two different student models that we discuss in detail below:  linear regression and the random nonlinear features model.  
A schematic of the network architecture for the latter model is depicted in Fig.~\ref{fig:schematics}(c).
Both student models take the general form
\begin{align}
\hat{y}(\vbx) &= \vbz(\vbx)\cdot\hbw.\label{eq:student}
\end{align}
 where $\hbw$ is a vector of fit parameters and $ \vbz(\vbx)$ is a vector of ``hidden'' features which each may depend on a combination of the input features $\vbx$. 
The hidden features $\vbz$ are effectively the representations of the data points from the perspective of the model.

Since we only fit the top layer of the network in both models, the number of fit parameters $N_p$ equals the number of hidden features,
and Eq.~\eqref{eq:student} is equivalent to a linear model with basis functions.
Despite its simplicity, we will show that this model reproduces much of the interesting behaviors observed in more complicated neural networks.

\subsubsection{Linear Regression}

For linear regression without basis functions [Fig.~\ref{fig:schematics}(a)], the representations of the features from the perspective of the model are simply the input features themselves so that $\vbz(\vbx)  = \vbx$. 
In other words, the hidden and input features are identical, leading to exactly one fit parameter for each feature, $N_p = N_f$. 

\subsubsection{Random Nonlinear Features Model}

In the random nonlinear features model [Fig.~\ref{fig:schematics}(b) and (c)],
the hidden features for each data point are related to the input features via a random matrix $W$ of size $N_f\times N_p$ and a nonlinear activation function $\varphi$,
\begin{align}
\vbz(\vbx) &= \frac{1}{\expval*{\varphi'}} \frac{\sigma_W\sigma_X}{\sqrt{N_p}}\varphi\qty(\frac{\sqrt{N_p}}{\sigma_W\sigma_X} W^T\vbx),\label{eq:rnlfm_model}
\end{align}
where $\varphi$ acts separately on each element of its input and ${\expval*{\varphi'} = \frac{1}{\sqrt{2\pi}}\int_{-\infty}^\infty \dd he^{-\frac{h^2}{2}}\varphi'(h)}$ is a normalization constant chosen for convenience (see Sec.~S1D of Supplemental Information~\cite{supporting_info}).
We take each element of $W$ to be independent and identically distributed, drawn from a normal distribution with zero mean and variance $\sigma_W^2/N_p$. 
The normalization by $N_p$ is chosen so that the magnitude of each hidden feature vector only depends on the ratio of the number of input features to parameters which we always take to be finite.
We place a factor of $\sqrt{N_p}/(\sigma_W\sigma_X)$ inside the activation function so that its argument has approximately unit variance, and a second pre-factor of $\sigma_W\sigma_X/\sqrt{N_p}$ in front of $\varphi$ to ensure that $\vbz$ reduces to a linear model $\vbz(\vbx) = W^T\vbx$ when the activation function is linear [i.e., $\varphi(h) = h$]. 
We note that while this model is technically a linear model with a specific choice of basis functions, it is equivalent to a two-layer neural network where the weights of the middle layer are chosen to be random and only the top layer is trained. 
Although our analytic results  hold for an arbitrary nonlinear activation function $\varphi$, all figures show results for ReLU activation where $\varphi(h) = \max(0, h)$.

\subsection{Fitting Procedure}

We train each model on a training data set consisting of $M$ data points, ${\mathcal{D}=\{(y_a, \vbx_a)\}_{a=1}^M}$.
For convenience, we organize the vectors of input features in the training set into an observation matrix $X$ of size $M\times N_f$ and define the length-$M$ vectors of training labels $\vby$, training label noise $\vbeps$, and label predictions for the training set $\hby$.
We also organize the vectors of hidden features evaluated on the input features of the
training set, $\{\vbz(\vbx_a)  \}_{a=1}^M$, into the rows of a hidden feature matrix $Z$ of size $M\times N_p$.

Given a set of training data $\mathcal{D}$, we solve for the optimal values of the fit parameters $\hbw$ by minimizing the standard loss function used for ridge regression,
\begin{align}
L(\hbw; \mathcal{D}) &= \frac{1}{2}\norm*{\Delta \vby}^2 + \frac{\lambda}{2}\norm*{\hbw}^2,\label{eq:loss}
\end{align}
where the notation $\norm*{\cdot}$ indicates an $L_2$ norm and  ${\Delta \vby = \vby - \hby}$ is the vector of residual label errors for the training set.
The first term is simply the mean squared error between the true labels and their predicted values, 
while the second term imposes standard $L_2$ regularization with regularization parameter $\lambda$. 
We will often work in the ``ridge-less limit'' where we take the limit $\lambda \rightarrow 0$.

\subsection{Model Evaluation}
To evaluate each model's prediction accuracy, we measure the training and test (generalization) errors. 
We define the training error as the mean squared residual label error of the training data,
\begin{align}
\train  &= \frac{1}{M}\norm*{\Delta \vby}^2.
\end{align}
We define the interpolation threshold as the model complexity at which the training error becomes exactly zero (in the thermodynamic and ridge-less limits). 
Analogously, we define the test error as the mean squared error evaluated on a test data set, ${\mathcal{D}'=\{(y_a', \vbx_a')\}_{a=1}^{M'}}$, composed of $M'$ new data points drawn independently from the same data distribution as the training data,
\begin{align}
\test &=  \frac{1}{M'}\norm*{\Delta \vby'}^2,\label{eq:test_err}
\end{align}
where $\Delta \vby' = \vby'-\hby'$ is a length-$M'$ vector of residual label errors between the vector of labels $\vby'$ and their predicted values $\hby'$ for the test set.

\subsection{Exact Solutions}

To solve for unique optimal solution, we set the gradient with respect to the fit parameters to zero, giving us a set of $N_p$ equations with $N_p$ unknowns,
\begin{align}
0 &= \pdv{L(\hbw)}{\hbw} =  -Z^T\Delta \vby + \lambda \hbw.\label{eq:grad}
\end{align}
Solving this set of equations results in a unique solution for the fit parameters,
\begin{align}
\hbw &= \qty[\lambda I_{N_p} + Z^TZ]^{-1}Z^T\vby.\label{eq:exact}
\end{align}

For simplicity, we also take the ridge-less limit where $\lambda$ is infinitesimally small ($\lambda \rightarrow 0$).
While our calculations do provide exact solutions for finite $\lambda$,
the solutions in the limits of small $\lambda$ are much more insightful.
In this limit, Eq.~\eqref{eq:exact} and  Eq.~\eqref{eq:student} approximate to
\begin{equation}
\hbw \approx Z^+  \vby\qqc \hat{y}(\vbx) \approx \vbz(\vbx)\cdot Z^+ \vby \label{eq:exactpseudo}
\end{equation} 
where $^+$ denotes a Moore-Penrose inverse, or pseudoinverse.

\subsection{Hessian Matrix}

We note that the solution for the fit parameters in Eq.~\eqref{eq:exact} depends on the matrix $Z^TZ$, which we refer to in the ridge-less limit as the Hessian matrix.
The matrix $Z^TZ/M$ can be interpreted as an empirical covariance matrix of the hidden features sampled by the training set when the hidden features are centered. 
The authors of Ref.~\cite{Advani2018} showed that for ridge regression, the divergence of the test error at the interpolation threshold can be naturally understood in terms of the spectrum of the Hessian. Inspired by this observation, we also explore the relationship between the eigenvalues of the Hessian and the double-descent phenomenon in our more general setting. 
 
To do so, we reproduce the known eigenvalue distribution for the Hessian for both models. 
We note that in linear regression (no basis functions), the Hessian is simply a Wishart matrix whose eigenvalues follow the Marchenko-Pastur distribution~\cite{Marchenko1967}.
The eigenvalue spectrum for the Hessian for the random nonlinear features model was explored in Ref.~\onlinecite{Pennington2019}.
For both models, we provide an alternative derivation of these spectra using the zero-temperature cavity method, allowing us to directly relate the eigenvalues of the Hessian matrix to the double-descent phenomenon.

\subsection{Derivation of Closed-Form Solutions}
While the expressions in the previous section are quite general, they hide much of the complexity of the problem and are difficult to analyze carefully.
 For this reason,  we make use of the zero-temperature cavity method to find closed-form solutions for all quantities of interest. 
 The zero-temperature cavity method has a long history in the physics of disordered systems and statistical learning theory and has been used to analyze the Hopfield model~\cite{Mezard2017} and more recently, compressed sensing~\cite{Ramezanali2019, Barbier2019}. 
 The cavity method is an alternative to the more commonly used Replica Method or analyses based on Random Matrix Theory.
 
 Like the Replica Method, finding closed-form solutions requires some additional assumptions. 
 In particular, we assume that the solutions satisfy a Replica-Symmetric Ansatz (an assumption we confirm numerically by showing remarkable agreement between our analytic results and simulations).
 Furthermore, we work in the thermodynamic limit, where $N_f, M, N_p \rightarrow \infty$ and keep terms to leading order in these quantities.
  Our results are exact under these assumptions.

To apply the zero-temperature cavity method, we start by defining the ratio of the number of input features to training data points $\alpha_f = N_f/M$ and the ratio of fit parameters to training data points $\alpha_p=N_p/M$. 
Next, we take the thermodynamic limit  $N_f, M, N_p \rightarrow \infty$, while keeping the ratios $\alpha_f$ and $\alpha_p$ finite.  
The essence of the cavity method is to expand the solutions of Eq.~\eqref{eq:grad} with $M+1$ data points, $N_f+1$ features and $N_p+1$ parameters about the solutions where one quantity of each type has been removed: ${(M+1, N_f+1, N_p+1)\rightarrow(M, N_f, N_p)}$. These two solutions are then related using generalized susceptibilities. 
The result is a set of algebraic self-consistency equations that can be solved for the distributions of the removed quantities.
The central limit theorem then allows us to approximate any quantity defined as a sum over a large number of random variables (e.g., the training and test errors) using just distributions for the removed quantities. 
Furthermore, using the procedure described in  Ref.~\cite{Cui2020}, 
we use the susceptibilities resulting from the cavity method to reproduce the known closed-form solutions for the eigenvalues spectra of the Hessian matrices for both models.
We refer the reader to the Supplemental Material~\cite{supporting_info} for further details on these calculations.

\section{Bias-Variance Decomposition}\label{sec:bvdefs}

The bias-variance decomposition separates test error into components stemming from three distinct sources: bias, variance, and noise. 
Informally, bias captures a model's tendency to underfit, reflecting the erroneous assumptions made by a model that limit its ability to fully express the relationships underlying the data. 
On the other hand, variance captures a model's tendency to overfit, capturing characteristics of the training set that are not a reflection of the data's true relationships, but rather a by-product of random sampling (e.g., noise).
As a result, a model with high variance  may not generalize well to other data sets drawn from the same data distribution.
Noise simply refers to an irreducible error inherent in generating a set of test data (i.e., the label noise in the test set).

Formally, bias represents the extent to which the label predictions $\hat{y}(\vbx)$ differs from the true function underlying the data distribution $y^*(\vbx)$ when evaluated on an arbitrary test data point $\vbx$ and averaged over all possible training sets $\mathcal{D}$~\cite{Bishop2006},
\begin{align}
\Bias\qty[\hat{y}(\vbx)] &= \E[\mathcal{D}]\qty[\hat{y}(\vbx)] - y^*(\vbx).\label{eq:raw_bias}
\end{align}
Likewise, variance formally measures the extent to which solutions of $\hat{y}(\vbx)$ for individual training sets $\mathcal{D}$ vary around the average~\cite{Bishop2006},
\begin{align}
\Var\qty[\hat{y}(\vbx)] &= \E[\mathcal{D}]\qty[\hat{y}^2(\vbx)] - \E[\mathcal{D}]\qty[\hat{y}(\vbx)]^2.\label{eq:raw_var}
\end{align}
Finally, the noise is simply the mean squared label noise associated with an arbitrary test data point $\vbx$,
\begin{align}
\mathrm{Noise} &= \E\qty[\varepsilon^2] = \sigma_\varepsilon^2.
\end{align}
The standard bias-variance decomposition relates these three quantities to the test error (averaged over all possible training sets $\mathcal{D}$).
In addition, we must take into account the fact that the test error is evaluated on $M'$ test data points, while the bias and variance only consider a single test point.
Since each test point is drawn from the same distribution, averaging the test error over all possible test sets $\mathcal{D}'$ is equivalent to averaging the bias and variance over the point $\vbx$.
This gives us the canonical bias-variance decomposition~\cite{Bishop2006},
\begin{align}
\E[\mathcal{D}', \mathcal{D}]\qty[\test] &= \E[\vbx]\qty[\Bias^2[\hat{y}(\vbx)]] + \E[\vbx]\qty[\Var[][\hat{y}(\vbx) ]] + \sigma_\varepsilon^2.
\label{eq:classicbvdec}
\end{align}
In this work, we also consider other sources of randomness (e.g., $\vbbeta$ and $W$).  To incorporate these random variables, we define the more general ensemble-averaged squared bias and variance, respectively, as
\begin{align}
\begin{split}
\expval*{\Bias^2[\hat{y}]} &= \E[\vbbeta, W, \vbx]\qty[\Bias[\hat{y}(\vbx)]^2]\\
&  = \E[\vbbeta, W, \vbx]\qty[\qty(\E[X, \vbeps][\hat{y}(\vbx)] - y^*(\vbx)) ^2]
\end{split}\\
\begin{split}
\expval*{\Var[][\hat{y} ]} &=\E[\vbbeta, W, \vbx]\qty[\Var[][\hat{y}(\vbx)]] \\
&= \E[\vbbeta, W, \vbx]\qty[\E[X, \vbeps][\hat{y}^2(\vbx)] - \E[X, \vbeps][\hat{y}(\vbx)]^2]\label{eq:ens_var}
\end{split}
\end{align}
where we have explicitly included all random variables considered in this work.
All analytic expressions we report are ensemble-averaged (denoted by angle brackets $\langle \cdot \rangle$) and utilize the ensemble-averaged bias-variance decomposition of the test error,
\begin{align}
\etest &= \expval*{\Bias^2[\hat{y} ]} +\expval*{\Var[][\hat{y} ]} +\sigma_\varepsilon^2.\label{eq:bvdecomp}
\end{align}
By fixing any parameters that do not pertain to the random sampling process of the test or training data (in this case, $\vbbeta$ and $W$), 
this formula properly reduces to the canonical bias-variance decomposition in Eq.~\eqref{eq:classicbvdec}.

\section{Results}\label{sec:results}

In this section, we provide analytic results for the training error, test error, bias, and variance, 
along with partial comparisons to numerical results. 
We limit ourselves to simply discussing major features of our closed-form solutions, deferring
a discussion of the implications of these results to the next section. 
Analytic derivations and complete comparisons to numerical results are left to the Supplemental Material~\cite{supporting_info}.

\subsection{General Solutions}\label{sec:res_gen}

We first report the forms of the solutions for arbitrary student and teacher models.
We find that the training error, test error, bias, and variance take the general forms
\begin{align}
\etrain &= \expval*{\Delta y^2}\label{eq:gen_train}
\\
\etest &= \sigma_X^2 \expval*{\Delta \beta^2} + \sigma_{\delta z}^2 \expval*{\hat{w}}^2 + \sigma_{\delta y^*}^2 + \sigma_\varepsilon^2\label{eq:gen_test}
\\
\expval*{\Bias^2[ \hat{y}]} &= \sigma_X^2 \expval*{\Delta\beta_1\Delta\beta_2} + \sigma_{\delta z}^2\expval*{\hat{w}_1\hat{w}_2} + \sigma_{\delta y^*}^2\label{eq:gen_bias}
\\
\begin{split}
\expval*{\Var[][\hat{y}]}  &= \sigma_X^2 \qty\big[\expval*{\Delta\beta^2}-\expval*{\Delta\beta_1\Delta\beta_2}]\\
&\qquad + \sigma_{\delta z}^2 \qty\big[\expval*{\hat{w}^2} - \expval*{\hat{w}_1\hat{w}_2}],\label{eq:gen_var}
\end{split}
\end{align}
which depend on five key ensemble-averaged quantities: $\expval*{\Delta y^2}$, $\expval*{\hat{w}^2}$, $\expval*{\Delta \beta^2}$, $\expval*{\hat{w}_1\hat{w}_2}$, and $\expval*{\Delta \beta_1\Delta \beta_2}$ (see Sec.~S1E of Supplemental Material~\cite{supporting_info} for detailed derivation). 
The first two quantities are the average of the squared training label errors $\expval*{\Delta y^2}$
and the average of the squared fit parameters $\expval*{\hat{w}^2}$.
The third quantity $\expval*{\Delta \beta^2}$ measures a model's accuracy in identifying the ground truth parameters $\vbbeta$.
To see this, we note that an estimate of the ground truth parameters for each model can be constructed from the fit parameters via the expression $\hbbeta \equiv W \hbw$,
with residual parameter errors $\Delta \vbbeta \equiv  \vbbeta - \hbbeta$.
Given these definitions, $\expval*{\Delta \beta^2}$ is then the average of the squared residual parameter errors.
Finally, the quantities $\expval*{\hat{w}_1\hat{w}_2}$ and $\expval*{\Delta \beta_1\Delta \beta_2}$ measure the average covariances of a pair fit parameters or residual parameter errors, respectively, that have the same index but derive from models trained on different training sets drawn independently from the same data distribution.

In addition to these five ensemble averages, the above expressions also depend on the quantities $\sigma_{\delta y^*}^2$ and $\sigma_{\delta z}^2$
which characterize the degree of nonlinearity of the labels and hidden features, respectively.
To define these quantities, we note that the nonlinear labels and hidden features considered in this work can each be decomposed into linear and nonlinear parts. 

First, we decompose the true labels in Eq.~\eqref{eq:true_labels} into two components that are statistically independent with respect to the distribution of input features,
allowing us to express the teacher model as
\begin{equation}
y(\vbx) = \vbx\cdot\vbbeta + \delta y_{\mathrm{NL}}(\vbx) + \varepsilon.\label{eq:ydecomp}
\end{equation}
The first term $\vbx\cdot\vbbeta$ captures the linear correlations between the input features $\vbx$ and the true labels $y^*$ via the ground truth parameter $\vbbeta$,
while the second term $\delta y^*_{\mathrm{NL}}(\vbx)$ captures the nonlinear behavior of $y^*$ [defined as ${\delta y^*_{\mathrm{NL}}(\vbx) \equiv y^*(\vbx) - \vbx\cdot\vbbeta}$].
The nonlinear component has zero mean (since the labels are centered with zero mean) and we define its variance as $\sigma_{\delta y^*}^2$.
Previously, this decomposition was implemented by noting that $\delta y^*_{\mathrm{NL}}(\vbx)$ behaves like an independent Gaussian process~\cite{Dhifallah2020, Mei2019}.
Here, we note that this approximation follows naturally in the thermodynamic limit from the relationship  $\vbbeta = \Sigma_{\vbx}^{-1}\Cov[\vbx][\vbx, y^*(\vbx)]$ where ${\Sigma_{\vbx}\equiv\Cov[\vbx][\vbx, \vbx^T]}$ is the covariance matrix of the input features (see Sec.~S1D of Supplemental Material~\cite{supporting_info}).

We also decompose the hidden features in Eq.~\eqref{eq:rnlfm_model} into three statistically independent components with respect to the distribution of input features,
\begin{equation}
\vbz(\vbx) = \frac{\mu_z}{\sqrt{N_p}}\vec{\mathbf{1}} +  W^T\vbx + \delta \vbz_{\mathrm{NL}}(\vbx).\label{eq:zdecomp}
\end{equation}
The first term $\mu_Z/\sqrt{N_p}\vec{\mathbf{1}}$ is the mean of each hidden feature where $\vec{\mathbf{1}}$ is a length-$N_p$ vector of ones.
Analogously to the label decomposition, the second term $W^T\vbx$ captures the linear correlations between the input features $\vbx$ and the hidden features $\vbz(\vbx)$ via the matrix of parameters $W$,
while the third term $ \delta \vbz_{\mathrm{NL}}(\vbx)$ captures the remaining nonlinear behavior of $\vbz(\vbx)$ [defined as ${\delta \vbz_{\mathrm{NL}}(\vbx) \equiv \vbz(\vbx) - \mu_z\vec{\mathbf{1}}/\sqrt{N_p} -  W^T\vbx}$].
The nonlinear component has zero mean and we define its total variance as $\sigma_{\delta z}^2$.
Like the nonlinear teacher model, it was previously observed that the nonlinear component of the hidden features behaves like an independent Gaussian process~\cite{Goldt2019}, and this decomposition has since been used as a common trick to obtain closed-form solutions for nonlinear models.
Here, we again note that this approximation follows naturally in the thermodynamic limit from the relationship  $W = \Sigma_{\vbx}^{-1}\Cov[\vbx][\vbx, \vbz(\vbx)^T]$ (see Sec.~S1D of Supplemental Material~\cite{supporting_info}).

We find the variance of the nonlinear components of the labels and hidden features, respectively, to be
\begin{align}
\sigma_{\delta y^*}^2 &= \sigma_\beta^2\sigma_X^2 \Delta f, &\Delta f &= \frac{\expval*{f^2}-\expval*{f'}^2}{\expval*{f'}^2}\\
\sigma_{\delta z}^2 &= \sigma_W^2\sigma_X^2 \Delta \varphi, & \Delta \varphi &= \frac{\expval*{\varphi^2}-\expval*{\varphi}^2 - \expval*{\varphi'}^2}{\expval*{\varphi'}^2},
\end{align}
where the quantities $\expval*{f^2}$, $\expval*{f'}$, $\expval*{\varphi^2}$, $\expval*{\varphi}$, $\expval*{\varphi'}$ are integrals of the form
\begin{align}
\expval*{g} &= \frac{1}{\sqrt{2\pi}}\int\limits_{-\infty}^\infty \dd h e^{-\frac{1}{2}h^2} g(h)\label{eq:integral}
\end{align}
with derivatives indicated via prime notation [e.g., ${f'=df(h)/dh}$].
The differences $\Delta f$ and $\Delta \varphi$ measure the ratio of the variances of each nonlinear component to its linear counterparts and go to zero in the linear limit.
For ReLU activation, $\varphi(h) = \max(h, 0)$, we find  $\expval*{\varphi^2}= 1/2$, $\expval*{\varphi} = 1/\sqrt{2\pi}$, and $\expval*{\varphi'} = 1/2$, resulting in $\Delta \varphi=1-2/\pi$. 

We derive Eqs.~\eqref{eq:gen_test}-\eqref{eq:gen_var} by decomposing the labels and hidden features of the test data (see Sec.~S1E in Supplemental Information~\cite{supporting_info}).
As a result, we can identify what elements of the test data lead to each term in these expressions.
We observe that terms proportional to $\sigma_X^2$ capture error arising from the linear components of the test data's labels and hidden features.
More precisely, these terms measure error due to the mismatch between the linear components of the test labels  and a model's predictions based solely on the linear components of the test data's hidden features ${\vbx\cdot \Delta\beta = \vbx\cdot\vbbeta - \vbx\cdot W\hbw}$.
In contrast, terms proportional to $\sigma_{\delta y^*}^2$ and $\sigma_{\delta z}^2$ represent errors due to the nonlinear components of the test data's labels and hidden features, respectively.
Since the model is linear in the fit parameters, it cannot fully capture nonlinear relationships in the test data that deviate from those observed in the training set.

Finally, we note that the decomposition of the labels in Eq.~\eqref{eq:ydecomp} suggests that each key quantity, along with total train error, test error, bias, and variance, decompose into contributions from the different parts of the training labels. 
Since each term is statistically independent, the contribution of each is proportional to its respective variance, 
allowing us to simply read off the sources of each type of error from our analytic results.
In particular, the linear and nonlinear components of the labels give rise to terms proportional to $\sigma_\beta^2\sigma_X^2$ and $\sigma_{\delta y^*}^2$, respectively,
while the label noise gives rise to terms proportional to  $\sigma_\varepsilon^2$.

\begin{figure*}[t!]
\centering
\includegraphics[width=1.0\linewidth]{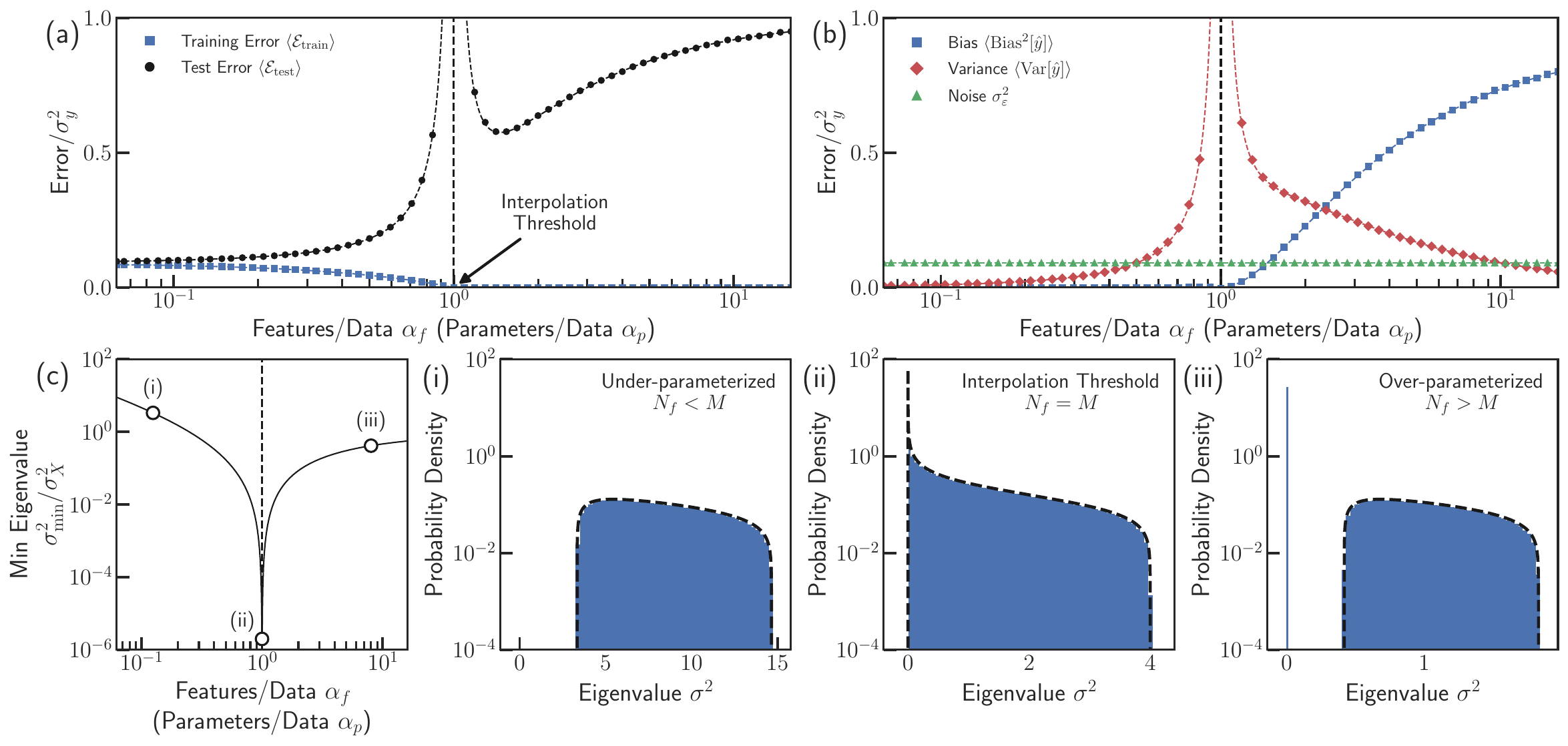} 
\caption{{\bf Linear Regression (No Basis Functions).}
Analytic solutions for the ensemble-averaged
{\bf(a)}  training error (blue squares) and test error (black circles), and
{\bf(b)} bias-variance decomposition of test error with contributions from the squared bias (blue squares), variance (red squares), and test set label noise (green triangles), plotted as a function of $\alpha_f=N_f/M$ (or equivalently, $\alpha_p=N_p/M$).
Analytic solutions are indicated as dashed lines with numerical results shown as points with small error bars indicating the error on the mean.
In each panel, a black dashed vertical line marks the interpolation threshold $\alpha_f = 1$. 
{\bf(c)} Analytic solution for the minimum eigenvalue $\sigma_{\min}^2$ of the Hessian matrix $Z^TZ$. 
Examples of the eigenvalue distributions are shown  {\bf(i)} in the  under-parameterized regime with $\alpha_f = 1/8$,  {\bf(ii)} at the interpolation threshold, $\alpha_f = 1$, and {\bf(iii)} in the  over-parameterized regime with $\alpha_f = 8$.
Analytic solutions for the distributions are depicted as black dashed curves with numerical results shown as blue histograms.
See Sec.~S4 of Supplemental Material~\cite{supporting_info} for additional simulation details.
}
\label{fig:linreg}
\end{figure*}

\subsection{Linear Regression}

Here, we present results for linear regression (no basis functions). 
Generally, our solutions are most naturally expressed in terms of ${\alpha_f = N_f/M}$, the ratio of input features to training data points, and $\alpha_p=N_p/M$, the ratio of fit parameters to training data points. However, in this case, the input and hidden features coincide ($N_f=N_p$), so all expressions depend only on $\alpha_f$. The ensemble-averaged training error, test error, bias, and variance for linear regression are:
\begin{widetext}
\begin{alignat}{2}
\etrain &= \left\{\begin{array}{c}
(\sigma_\varepsilon^2 + \sigma_{\delta y^*}^2) (1-\alpha_f)\\
0
\end{array}\right. &\quad &\begin{array}{l}
\qif N_f < M\\
 \qif N_f > M
\end{array} \label{eq:linreg_train}\\
\etest &= \left\{\begin{array}{c}
(\sigma_\varepsilon^2 + \sigma_{\delta y^*}^2) \frac{1}{(1-\alpha_f)}\\
\sigma_\beta^2\sigma_X^2 \frac{(\alpha_f-1)}{\alpha_f} +  (\sigma_\varepsilon^2+\sigma_{\delta y^*}^2) \frac{\alpha_f}{(\alpha_f-1)}
\end{array} \right.&\quad &\begin{array}{l}
\qif N_f < M\\
 \qif N_f > M
\end{array} \label{eq:linreg_test}\\
\expval*{\Bias^2[ \hat{y}]}  &= \left\{
\begin{array}{c}
\sigma_{\delta y^*}^2\\
\sigma_\beta^2\sigma_X^2 \frac{(\alpha_f-1)^2}{\alpha_f^2} + \sigma_{\delta y^*}^2 
\end{array}
\right.&\quad &\begin{array}{l}
\qif N_f < M\\
 \qif N_f > M
\end{array} \label{eq:linreg_bias2}\\
\expval*{\Var[][ \hat{y}]} &= \left\{\begin{array}{c}
(\sigma_\varepsilon^2 + \sigma_{\delta y^*}^2) \frac{\alpha_f}{(1-\alpha_f)}\\
\sigma_\beta^2\sigma_X^2 \frac{(\alpha_f-1)}{\alpha_f^2} +   (\sigma_\varepsilon^2 + \sigma_{\delta y^*}^2)  \frac{1}{(\alpha_f-1)}
\end{array} \right. &\quad &\begin{array}{l}
\qif N_f < M\\
 \qif N_f > M.
\end{array} \label{eq:linreg_var}
\end{alignat}
\end{widetext}
In writing these expressions, we have taken the ridge-less limit, $\lambda \rightarrow 0$ (when a quantity is reported as zero, leading terms of order $\lambda^2$ are reported in Sec.~S1F of the Supplemental Material~\cite{supporting_info}).

In Fig.~\ref{fig:linreg}(a), we plot the expressions for the training and test error in Eqs.~\eqref{eq:linreg_train} and \eqref{eq:linreg_test} with comparisons to numerical results for a linear teacher model $\sigma_{\delta y^*}^2 = 0$.
We find that the model's behavior falls into two broad regimes, depending on whether  $\alpha_f >1$ or  $\alpha_f <1$
 (or equivalently, $\alpha_p>1$ or $\alpha_p<1$). 
In Fig.~\ref{fig:linreg}(a), we observe that below $\alpha_f=1$, the training error is finite, decreasing monotonically as $\alpha_f$ increases until
reaching zero at $\alpha_f=1$. 
Beyond this threshold, the addition of extra features/parameters has no effect and the training error remains pinned at zero. 
Thus, $\alpha_f=1$ corresponds to the interpolation threshold, separating the regions where the model has zero and nonzero training error, 
i.e., the under- and over-parameterized regimes. 
At the interpolation threshold, the test error diverges [Fig.~\ref{fig:linreg}(a)],
indicative of a phase transition based on the divergence of the corresponding susceptibilities in the cavity equations (see Sec.~\ref{sec:suscept} and Sec.~S1F of Supplemental Material~\cite{supporting_info}).

The bias and variance, reported in Eqs.~\eqref{eq:linreg_bias2} and \eqref{eq:linreg_var}, are plotted in Fig.~\ref{fig:linreg}(b).
When $\alpha_f \leq 1$, the bias is zero for a linear teacher model. 
This can be understood by noting that the teacher and student models match in this case and there are more
data points than parameters (i.e., we are working in a regime where intuitions for classical statistics apply). 
In this regime, the variance increases monotonically as $\alpha_f$ is increased, diverging at the interpolation threshold as the model succumbs to overfitting. 
However, when $\alpha_f > 1$,  the variance exhibits the opposite behavior,  decreasing monotonically as $\alpha_f$ is increased. 
The bias, on the other hand, \textit{increases} monotonically towards the limit $\sigma_\beta^2\sigma_X^2+\sigma_{\delta y^*}^2$ as $\alpha_f$ goes to infinity.  
Consequently, the test error in the over-parameterized regime is characterized by a surprising ``inverted  bias-variance trade-off'' where the bias increases with model complexity while the variance decreases. 

In the solutions for the training error, test error, and variance, we observe that the error due to the nonlinear label components (proportional to $\sigma_{\delta y^*}^2$) always appears as an additive component to the errors stemming from the label noise (proportional to $\sigma_\varepsilon^2$).
However, unlike the label noise, the nonlinear label variance $\sigma_{\delta y^*}^2$ also appears in the bias as an additional constant irreducible term that arises as a result of attempting to fit a nonlinear data distribution with a model that is linear in the fit parameters.

Finally, in Fig.~\ref{fig:linreg}(c), we report the minimum nonzero eigenvalue $\sigma_{\min}^2$ of the Hessian matrix $Z^TZ$,
with examples of the full eigenvalue spectrum shown in  {Figs.~\ref{fig:linreg}(i)-(iii)}.
Since $Z^TZ=X^TX$ for our model of linear regression, the eigenvalue spectrum is simply the Marchenko-Pastur distribution (see Sec.~S2A of Supplemental Material for derivation).
Importantly, we find the interpolation threshold  $\alpha_f = 1$ coincides with the point at which  $\sigma_{\min}^2$ goes to zero.
In the under-parameterized regime, there is a finite gap in the eigenvalue spectrum, with no small eigenvalues.
In the over-parameterized, there is also a finite gap, but instead between the bulk of the spectrum and a buildup of eigenvalues at exactly zero.
We discuss the implications of these findings later in Sec.~\ref{sec:variance}.

\begin{figure*}[t!]
\centering
\includegraphics[width=1.0\linewidth]{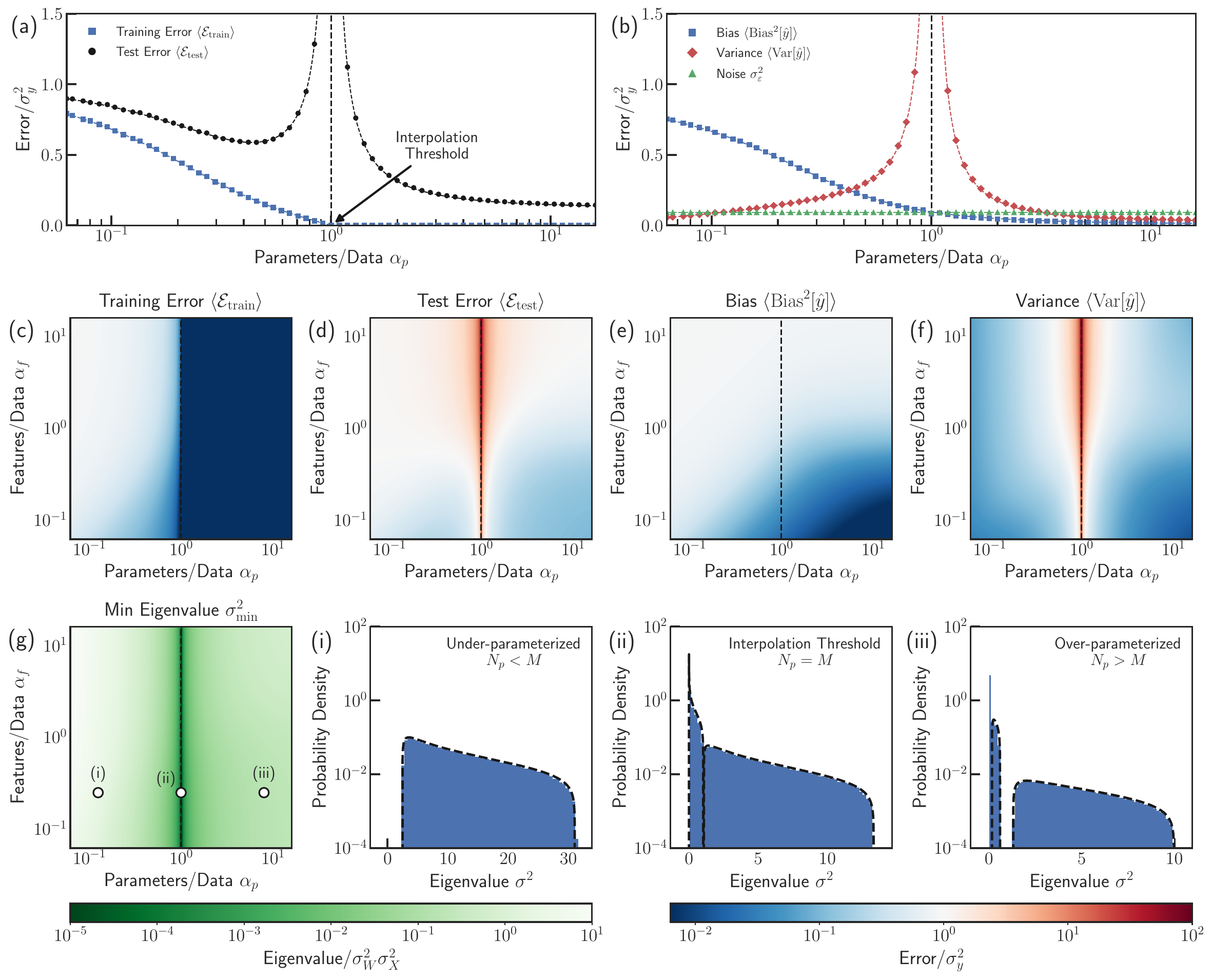} 
\caption{
{\bf Random Nonlinear Features Model (Two-layer Neural Network).}
Analytic solutions for the ensemble-averaged
{\bf(a)}  training error (blue squares) and test error (black circles), and
{\bf(b)} bias-variance decomposition of test error with contributions from the squared bias (blue squares), variance (red squares), and test set label noise (green triangles), plotted as a function of $\alpha_p=N_p/M$ for fixed $\alpha_f = N_f/M = 1/4$.
Analytic solutions are indicated as dashed lines with numerical results shown as points. 
Analytic solutions as a function of both $\alpha_p$ and $\alpha_f$ are also shown for the the ensemble-averaged {\bf(c)} training error, {\bf(d)} test error, {\bf(e)} squared bias, and {\bf(f)} variance.
In all panels, a black dashed line marks the boundary between the under- and over-parameterized regimes at $\alpha_p = 1$. 
{\bf(g)} Analytic solution for the minimum eigenvalue $\sigma_{\min}^2$ of the Hessian matrix $Z^TZ$. 
Examples of the eigenvalue distributions are shown  {\bf(i)} in the  under-parameterized regime with $\alpha_p = 1/8$,  {\bf(ii)} at the interpolation threshold, $\alpha_p = 1$, and {\bf(iii)} in the  over-parameterized regime with $\alpha_p = 8$, all for $\alpha_p = 1/4$.
Analytic solutions for the distributions are shown as blacked dashed curves with numerical results shown as blue histograms.
See Sec.~S4 of Supplemental Material~\cite{supporting_info} for additional simulation details.
}\label{fig:rnlfm}
\end{figure*}

\subsection{Random Nonlinear Features Model}\label{sec:results_rnlfm}

Unlike the solutions for linear regression, the analytic expressions for the random nonlinear features model are not so simple, so we defer these expressions to the Appendix.
In Fig.~\ref{fig:rnlfm}(a) and (b) we plot the training error, test error, bias, and variance as a function of $\alpha_p=N_p/M$ for fixed $\alpha_f<1$ (more data points $M$ than input features $N_f$),
while in Figs.~\ref{fig:rnlfm}(c)-(f), we plot all quantities as a function of both $\alpha_p$ and $\alpha_f$.
In all plots, we depict the special case of a linear teacher model $\sigma_{\delta y^*}^2 = 0$ and ReLU activation $\varphi(h) = \max(h, 0)$.
Analogously to linear regression,
the nonlinear model has two distinct regimes separated by the line $\alpha_p=1$. 
In Fig.~\ref{fig:rnlfm}(c), we find that the training error is finite when $\alpha_p < 1$ and goes to zero when $\alpha_p \geq 1$,
marking the boundary $\alpha_p=1$ as the interpolation threshold. 
Fig.~\ref{fig:rnlfm}(d) shows that the test error diverges at each point along this boundary and
no longer diverges when $\alpha_f=1$ as in the linear case [Fig.~\ref{fig:linreg}(a)].
However, we do still find that the divergence in the test error is associated with a phase transition indicated by diverging susceptibilities (see Sec.~\ref{sec:suscept} and Appendix).

In addition, the test error only displays a small qualitative difference between the regimes where $\alpha_f<1$ and $\alpha_f > 1$.
We find that the test error only shows a canonical  double-descent 
behavior when $\alpha_f < 1$ [Fig~\ref{fig:rnlfm}(d)].
As in linear regression, the variance [Fig.~\ref{fig:rnlfm}(f)], accounts for the divergence of the test error at the phase boundaries,
while the bias [Fig.~\ref{fig:rnlfm}(e)] remains finite, decreasing monotonically for all $\alpha_p$.
However, unlike linear regression, the bias of the nonlinear model never reaches zero, even for a linear teacher model. 
Furthermore, the closed-form solutions show that the nonlinear components of the labels $\sigma_{\delta y^*}^2$ contribute in the same way as in the two linear models, adding a small constant irreducible bias [Eq.~\eqref{eq:gen_bias}], along with an additive component to the label noise (see Appendix).

Finally, in Fig.~\ref{fig:rnlfm}(g), we report the minimum nonzero eigenvalue $\sigma_{\min}^2$ of the Hessian matrix $Z^TZ$ as a function of both $\alpha_p$ and $\alpha_f$ (see Sec.~S2B of Supplemental Material for derivation of analytic results).
We find that  $\sigma_{\min}^2$ approaches zero along the entire interpolation boundary $\alpha_p = 1$.
In {Figs.~\ref{fig:rnlfm}(i)-(iii)}, we show examples of the full eigenvalue spectrum for $\alpha_f < 1$ for the under- and over-parameterized regimes, along with the interpolation threshold.
We find that the spectrum in the under-parameterized regime displays a finite gap that goes to zero near the interpolation threshold.
In the over-parameterized regime, we find that although the gap between the buildup of eigenvalues at zero and the nonzero eigenvalues is much smaller, it is still finite.
Interestingly, we also find that additional gaps can appear in the eigenvalue distribution between sets of finite-valued eigenvalues,
which likely reflects the fact that ReLU activation functions result in a large fraction of zero-valued entries in $Z$.

\section{Understanding bias and variance in over-parameterized models}\label{sec:discuss}

Having presented our analytic results, we now discuss the implications of our calculations for understanding bias and variance in a more general setting. 
Our discussion emphasizes the qualitatively new phenomena that are present in over-parameterized models.

\subsection{Two sources of bias: Imperfect models and incomplete exploration of features}

Traditionally, bias is viewed as a symptom of a model making incorrect assumptions about the data distribution (a mismatch between the teacher and student models). 
However, our calculations show that this description of the origin of bias is incomplete. 
A striking feature of our results is that over-parameterized models can be biased even if our statistical models are expressive enough to fully capture all relationships underlying the data. 
In fact, linear regression shows us that one can have a nonzero bias even if the student and teacher models are identical [e.g., $f(h) = h$]. 
Even when the student and teacher model are the same, the bias is nonzero if there are more input features $N_f$ than data points $M$ [see $\alpha_f >1$ region of Fig.~\ref{fig:linreg}(b)].

To better understand this phenomenon, it is helpful to think of the input features as spanning an $N_f$-dimensional space.
The training data can be embedded in this $N_f$-dimensional input feature space by considering the eigenvectors, or principal components, 
of the empirical covariance matrix of input features $X^TX/M$, with $X$ defined as the $M \times N_f$  design matrix whose rows 
correspond to training data points and columns to input features (see Sec.~\ref{sec:setup}). 
When there are more data points than input features ($M > N_f$),
the training data will typically span the entire $N_f$-dimensional input feature space (i.e., the principal components of $X^TX$ generically span all of input feature space). 
In contrast, when there are fewer training data points than input features ($M < N_f$), the training data will typically span only a fraction of the entire input feature space
(i.e., the principal components will span a subspace of the full $N_f$-dimensional input feature space).
For this reason, when $M<N_f$ the model is ``blind'' to data that varies 
along these unsampled directions. 
Consequently, any predictions the model makes about these directions will reflect assumptions implicitly or explicitly built into the model rather than relationships learned from the training data set. 
This can result in a nonzero bias even when the teacher and student models are identical.

In the random nonlinear features model, the nonlinear activation function makes it impossible to perfectly represent the input features via the hidden features, even for a linear teacher model.
While increasing the number of hidden features does reduce bias, the nonlinear model is never able to perfectly capture the linear nature of the data distribution and is always biased.
Similarly, neither model is able to express the nonlinear components of the labels for a nonlinear teacher model, resulting in a constant, irreducible bias in both cases.

Finally, we wish to point out that, unlike some previous studies, we find that the bias never diverges~\cite{Adlam2020, Ba2020, Mei2019}, including at the interpolation transition,
nor does it reach a minimum at the transition, remaining constant into the over-parameterized regime in the ridge-less limit~\cite{Adlam2020, DAscoli2020, DAscoli2020a, Lin2020}.
Instead, we find that the bias remains finite and decreases monotonically, even in the absence of regularization.

\begin{figure*}[t]
\centering
\includegraphics[width=1.0\linewidth]{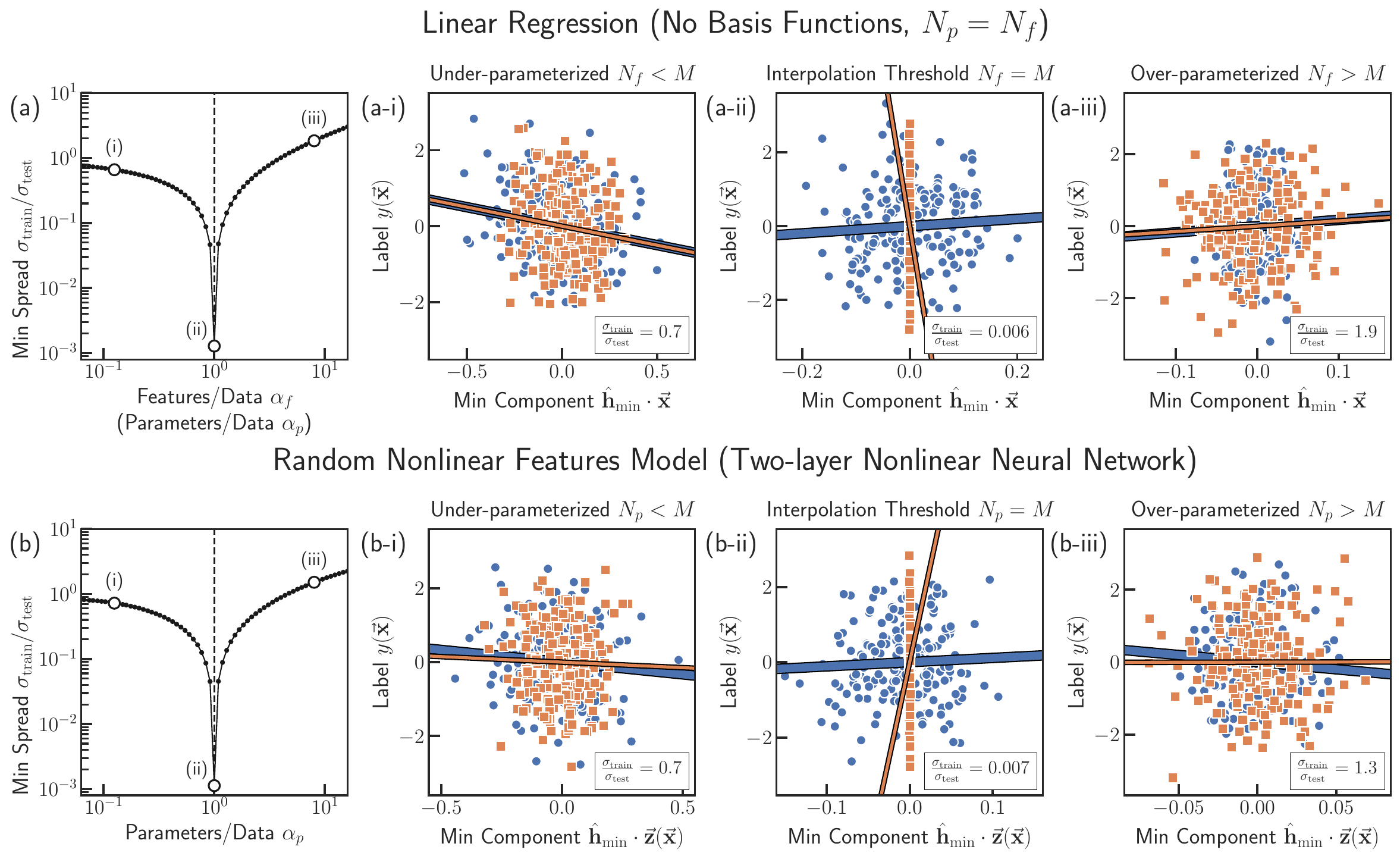} 
\caption{
{\bf Poorly sampled directions in space of features lead to overfitting.} Demonstrations of this phenomenon are shown for {\bf (a)} linear regression and {\bf(b)} the random nonlinear features model. Columns (i), (ii), and (iii) correspond to models which are under-parameterized, exactly at the interpolation threshold, or over-parameterized, respectively. 
In each example, the relationship between the labels and the projection of their associated input or hidden features onto the minimum principal component $\hbh_{\min}$ of $Z^TZ$ is depicted for a set of training data (orange squares) and a test set (blue circles).
Orange lines indicate the relationship learned by a model from the training set, 
while the expected relationship for an average test set is shown as a blue line.
In the left-most column, the spread (standard deviation) of an average training set along the $x$-axis, $\sigma_{\mathrm{train}}^2 = \sigma_{\min}^2/M$, is plotted relative to the spread that would be expected for an average test set, $\sigma_{\mathrm{test}}^2$, for simulated data as a function of $\alpha_p$. 
Smaller values are associated with lower prediction accuracy on out-of-sample data, coinciding with small eigenvalues in $Z^TZ$.
All results are shown for a linear teacher model.
See Supplemental Material~\cite{supporting_info} for analytic derivations of learned and expected relationships and spreads along minimum principal components (Sec.~S3), along with additional details of numerical simulations (Sec.~S4).
}\label{fig:narrow}
\end{figure*}

\subsection{Variance: Overfitting stems from poorly sampled direction in space of feature}\label{sec:variance}

Variance measures the tendency of a model to overfit, or attribute too much significance to,  
aspects of the training data that do not generalize to other data sets. 
Even when all data is drawn from the same distribution, 
the predictions of a trained model can vary depending on the details of each particular training set.
More specifically, a model may exhibit high variance when a direction in feature space is present in the training data, but not sampled well enough to reflect its 
true nature in the underlying data distribution.
When presented with new data that has a significant contribution along this under-sampled direction,
the model is forced to extrapolate (often incorrectly) based on the little information it can glean from the training set.

In linear regression, the empirical variance along each principal direction is explicitly measured by its associated eigenvalue in the empirical covariance matrix $X^TX/M$.
Generally, a model's variance will be dominated by the most poorly sampled principal direction, or minimum component $\hbh_{\min}$, corresponding to the smallest nonzero eigenvalue $\sigma_{\min}^2$ of $X^TX$.
The projection of an arbitrary data point $\vbx$ onto $\hbh_{\min}$ can be found by taking a dot product, $\hbh_{\min}\cdot \vbx$.
The observed variance of $\hbh_{\min}\cdot \vbx$ for a given training set is $\sigma^2_{\mathrm{train}} = \sigma_{\min}^2/M$.
For comparison, we define the true, or expected, variance of $\hbh_{\min}\cdot \vbx$ for an average test set as  $\sigma^2_{\mathrm{test}}$, representing data points drawn from the full data distribution (see Sec.~S3 of Supplemental Material~\cite{supporting_info}).

The first row of Fig.~\ref{fig:narrow} shows how observing a small variance along a particular direction sampled by the training data can lead to overfitting in linear regression.
In Fig.~\ref{fig:narrow}(a), we plot the average ratio $\sigma_{\mathrm{train}}/\sigma_{\mathrm{test}}$ as a function of $\alpha_f$ for simulated data.
In Figs.~\ref{fig:narrow}(a-i)-(a-iii), we then plot the labels $y$ versus $\hbh_{\min}\cdot \vbx$ for the training set (orange points) and an equally-sized test data set (blue points), representative of the full data distribution.
In each panel, the relationship between the labels and $\hbh_{\min}\cdot \vbx$ as predicted by the model based on the training set is depicted as an orange line.
For comparison, we also show the expected relationship for an average test set as a blue line, representing the true relationship underlying the data (see Sec.~S3 of Supplemental Material~\cite{supporting_info} for explicit formulas).

In Fig.~\ref{fig:narrow}(a-i), we see that when the model is under-parameterized ($N_f < M$), 
the spread of the training set along the minimum component is comparable to that of the test set ($\sigma_{\mathrm{train}} / \sigma_{\mathrm{test}}= 0.7$).
Because there are more data points than input features,
many of the data points are likely to contain significant contributions from each direction, including the minimum component,
corresponding to a finite gap in the corresponding eigenvalue distribution depicted in Fig.~\ref{fig:linreg}(c-i).
As a result, the training set will provide the model with an accurate representation of the data distribution along this direction in feature space.
In this case, we see that the model is able to closely approximate the true relationship in the data even in the presence of noise.

However, at the interpolation threshold when the number of input features equals the number of data points ($N_f = M$), 
Fig.~\ref{fig:narrow}(a-ii) shows that the spread of the training data points along the minimum component is very narrow compared to the test data ($\sigma_{\mathrm{train}} / \sigma_{\mathrm{test}} \approx 0.006$),
while in Fig.~\ref{fig:linreg}(c-ii), we observe that the gap in the eigenvalue distribution disappears.
In this case, the training set contains a very small, but insufficient, amount of information about the data distribution along this direction.
This poor sampling causes the model to overfit the noise of the training set, resulting in a slope that is much larger than that of the true relationship.
When presented with a new data point with a significant contribution along $\hbh_{\min}$, 
the model will be forced to extrapolate beyond the narrow range of $\hbh_{\min}\cdot \vbx$ observed in the training set.
This extrapolation will hamper the model's ability to generalize, leading to inaccurate predictions that are highly dependent on the precise details of the noise sampled by the training set.

Surprisingly, we find in Fig.~\ref{fig:narrow}(a-iii) that further increasing the number of features so that the model becomes over-parameterized ($N_f > M$) actually \textit{increases} the spread in the training data along the minimum component ($\sigma_{\mathrm{train}} / \sigma_{\mathrm{test}} \approx 1.9$), \textit{reducing} the effects of overfitting.
When there are more features than data points, 
each data point is likely to explore a never-before-seen combination of features.
Naively, one would expect this to leave many of the directions poorly sampled.
However, because the norm of each data point is approximately the same in the thermodynamic limit,
the fact that each data point is different means that all are likely to make independent contributions of different sizes to the sampled directions, including $\hbh_{\min}$.
Even if this means only a single data point contributes to a particular component, 
this contribution must be of significant size for the data point to be independent of the rest.
So while some directions are not represented in the training set at all,
the ones that are present are typically well-represented by at least one -- if not many -- data points, providing a sufficient amount of signal (or spread) to reveal relationships in the underlying distribution.
Consequently, the model is able to learn the true relationship between the labels and features,
just as in the under-parameterized case. 
We observe this phenomenon directly in the eigenvalue distribution in  Fig.~\ref{fig:linreg}(c-iii),
with a buildup of eigenvalues at exactly zero corresponding to unsampled directions accompanied by a finite gap separating these eigenvalues from the rest of the distribution.
This is the underlying reason that the variance decreases with model complexity beyond the interpolation threshold.
A similar observation was made in Ref.~\onlinecite{Advani2020} in the context of ridge regression using methods from Random Matrix Theory.

In the second row of Fig.~\ref{fig:narrow}, we demonstrate that the same patterns also lead to overfitting in the random nonlinear features model,
indicating that the intuition gained from linear regression translates directly to more complex settings.
In this case, the model can be interpreted as indirectly sampling the data distribution via the empirical covariance matrix of hidden features $Z^TZ/M$.
We calculate the minimum component $\hbh_{\min}$ as the principal component of $Z^TZ$ with the smallest eigenvalue.
In Figs.~\ref{fig:narrow}(b-i)-(b-iii), we plot the labels $y$ versus the projection of each data point's hidden features $\vbz$ onto this minimum component $\hbh_{\min}\cdot\vbz$, 
with the ratio $\sigma_{\mathrm{train}}/\sigma_{\mathrm{test}}$ shown in Fig.~\ref{fig:narrow}(b).
In contrast to linear regression, we see that overfitting results from poorly sampling --  or observing very limited spread along --
a direction in the space of hidden features rather than input features.
In the random nonlinear features model, overfitting is most pronounced when the number of hidden features matches the number of data points at the interpolation threshold ($N_p = M$),
where the gap in the eigenvalue distribution at zero disappears [Fig.~\ref{fig:rnlfm}(g-ii)].

\begin{figure*}[t!]
\centering
\includegraphics[width=1.0\linewidth]{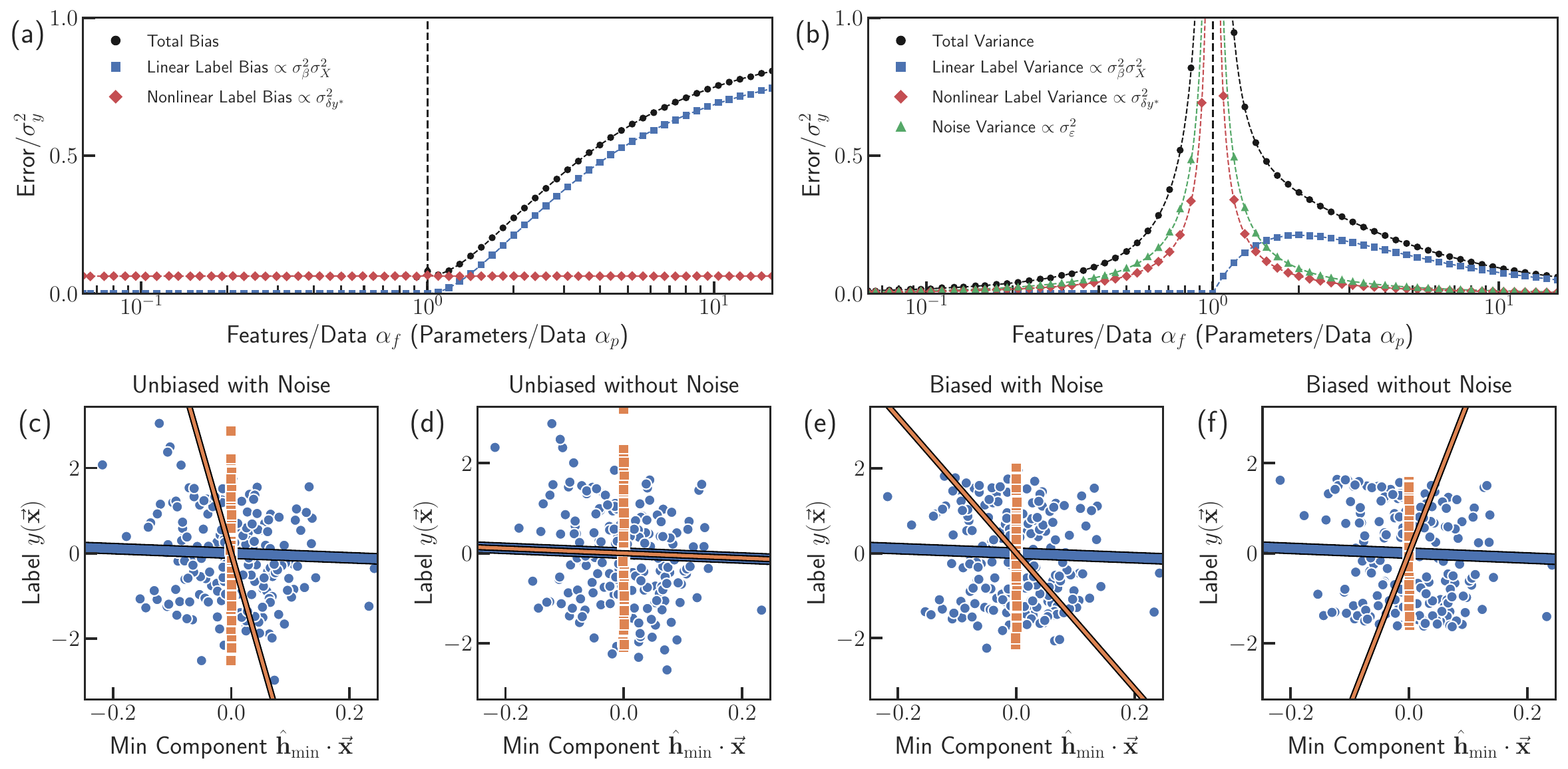} 
\caption{
{\bf Biased models can interpret signal as noise. }
{\bf(a)} The total bias (black circles) with contributions from the linear label components (blue squares) and nonlinear label components  (red diamonds), and the
 {\bf(b)} the total variance (black circles) with contributions from the linear label components (blue squares), nonlinear label components (red diamonds), and the label noise (green triangles)
are shown for linear regression with a nonlinear teacher model $f(h) = \tanh(h)$ [see Eq.~\eqref{eq:true_labels}].
Analytic solutions are indicated as dashed lines with numerical results shown as points.
Contributions from the linear label components, nonlinear label components, and label noise are found by identifying terms in the analytic solutions proportional to $\sigma_\beta^2\sigma_X^2$, $\sigma_{\delta y^*}^2$, and $\sigma_\varepsilon^2$, respectively.
 Each source of bias acts as effective noise, giving rise to a corresponding source of variance.
The effects of this phenomenon on the relationships learned by a linear regression model are depicted at the interpolation threshold for an unbiased model with linear data, $f(h) = h$, {\bf(c)} with noise and {\bf(d)} without noise, and for a biased model with nonlinear data, $f(h) = \tanh(h)$,  {\bf(e)} with noise and {\bf(f)} without noise. 
In each example, the relationship between the labels and the projection of their associated input features onto the minimum principal component $\hbh_{\min}$ of $X^TX$ is depicted for a set of training data (orange squares) and a test set (blue circles).
Orange lines indicate the relationship learned by a model from the training set, 
while the expected relationship for an average test set is shown as a blue line.
See Supplemental Material~\cite{supporting_info} for analytic derivations of learned and expected relationships (Sec.~S3), along with additional details of numerical simulations (Sec.~S4).
}\label{fig:narrow_sigvar}
\end{figure*}

\subsection{Biased models can interpret signal as noise}

Typically, variance is attributed to overfitting inconsistencies in the labels due to noise in the training set.
Indeed, we observe that the contribution to the variance due to noise is nonzero in each model.
Surprisingly, we also find that overfitting can occur in the absence of noise when a model is biased.
In each model, we observe a direct correspondence between each source of bias and a source of variance.
In other words, in the absence of noise, the variance is zero only when the bias is zero.

To illustrate this, in Figs.~\ref{fig:narrow_sigvar}(a) and (b), we plot the contributions to the bias and variance, respectively, from the different statistically independent components of the labels in Eq.~\eqref{eq:ydecomp} for our model of linear regression with a nonlinear teacher model of the form $f(h) = \tanh(h)$ [see Eq.~\eqref{eq:true_labels}]. In this case, note that our model can never fully represent the true data distribution and hence will always be biased. We find that both contributions to the bias from the linear (blue) and nonlinear (red) components of the training labels, proportional to $\sigma_\beta^2\sigma_X^2$ and $\sigma_{\delta y^*}^2$, respectively, in Eq.~\eqref{eq:linreg_bias2} (see Sec.~\ref{sec:res_gen}), have a corresponding contribution to the variance for all values of $\alpha_f$ in Eq.~\eqref{eq:linreg_var}.
This suggests the following interpretation:
a model with nonzero bias gives rise to variance by interpreting part of the training set's signal $y^*$ as noise.
In other words, a model which cannot fully express the relationships underlying the data distribution may inadvertently treat this unexpressed signal as noise.

We demonstrate this phenomena in Figs.~\ref{fig:narrow_sigvar}(c)-(f) using our model of linear regression trained at the interpolation threshold ($N_pf = M$),
where the only contribution to the bias stems from the nonlinear components of the training labels.
In each panel, we plot the labels as a function of the projection of each data point's input features onto the minimum principal component $\hbh_{\min}\cdot\vbx$ for a set of training data (orange) and a set of test data (blue).
We then compare the resulting model (orange line) to the expected relationship for an average test set (blue line), representing the underlying relationship in the data distribution.

To confirm that bias is necessary for this phenomenon, Figs.~\ref{fig:narrow_sigvar}(c) and (d) show simulations for a linear teacher model  $f(h) = h$ with and without label noise. In this case, the student and teacher models match and the model is unbiased.
As expected, we find that with label noise, the model overfits the training data and the resulting slope does not accurately reflect the true relationship underlying the data,
while without label noise, the model is  able to avoid overfitting and provides a good approximation of the true relationship

In Figs.~\ref{fig:narrow_sigvar}(e) and (f), we performed the same simulations for a non-linear teacher model $f(h) = \tanh(h)$ with and without label noise.
In this case, the student and teacher models do not match and the model is always biased.
 With label noise, the model overfits the training data.
However, even in the absence of label noise, the model \textit{still} overfits the training data.
Collectively, our simulations indicate that even if the label noise is zero, any finite amount of bias can result in overfitting, especially if the training set severely undersamples the data along a particular direction in feature space.

We find that this observation holds for both models for every observed source of bias, 
including contributions stemming from the linear and nonlinear components of the labels (proportional to $\sigma_\beta^2\sigma_X^2$ and $\sigma_{\delta y^*}^2$, respectively)
and the linear and nonlinear components of the hidden features in the test data (proportional to $\sigma_X^2$ and $\sigma_{\delta z}^2$, respectively, in Eq.~\eqref{eq:gen_bias} and \eqref{eq:gen_var}).
As a result, in the absence of label noise, the test error can only diverge at an interpolation threshold when a model is biased.
Finally, we note that this behavior also manifests in contributions to the training error in the under-parameterized regime for each model,
with each source of bias corresponding to an additional source of training error below the interpolation threshold.

\subsection{Interpolating is not the same as overfitting}

Our results make clear that interpolation (zero training error) occurs independently from overfitting (poor generalization) in over-parameterized models.
Interpolation occurs when the number of independent directions in the space of hidden features (or equivalently, input features in linear regression) sampled by the training set is sufficient to account for the variations in the labels.
In both models, the interpolation threshold is located where the number of principal components (measured via the rank of $Z^TZ$) matches the number of data points.
On the other hand, the test error diverges as a result of the variance diverging at the interpolation threshold. 
These larges variances result from poor sampling along directions in $Z^TZ$ (small eigenvalues), resulting in very little spread of data along these directions in the training set relative to the full distribution.

In under-parameterized models, interpolation and overfitting coincide.
Increasing the number of fit parameters results in a greater number of sampled directions in the space of features, 
but makes it more likely to poorly sample any particular direction, resulting in large variance.
The result is that the interpolation threshold always coincides with a divergence in the test error.
In contrast, interpolation and overfitting occur independently in over-parameterized models.
Once the interpolation threshold is reached, further increasing the number of fit parameters cannot improve the training error since it is already at a minimum.
However, increasing the model complexity can reduce the effects of overfitting and decrease the variance
by allowing for better sampling along the directions captured by the training set (Fig.~\ref{fig:narrow}). 
For this reason, increasing model complexity past the interpolation threshold can actually result in an \textit{increase} in model performance without succumbing to overfitting.

\subsection{Susceptibilities measure sensitivity to perturbations}\label{sec:suscept}

\begin{figure}[t!]
\centering
\includegraphics[width=1.0\linewidth]{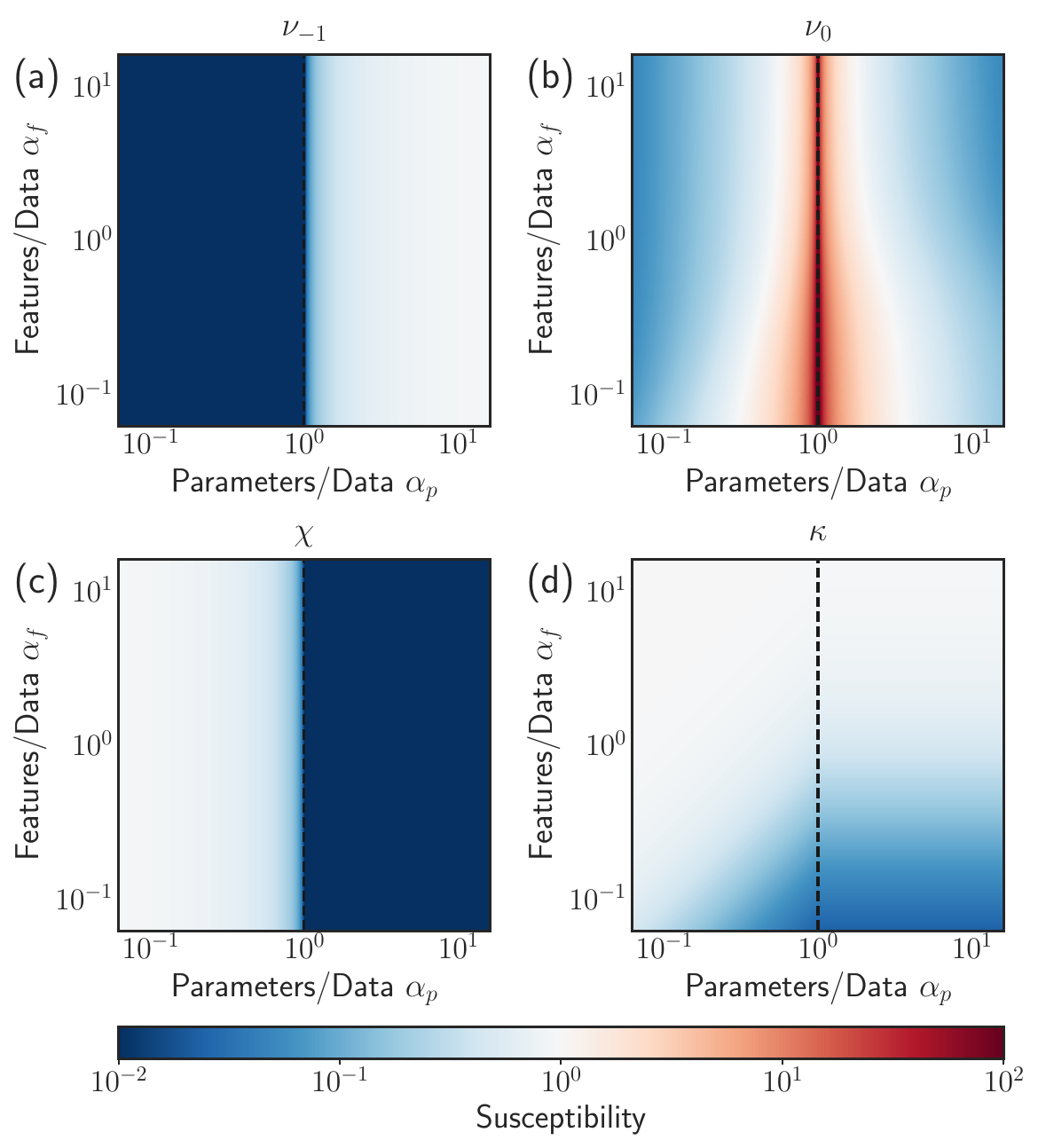} 
\caption{
{\bf Susceptibilities for Random Nonlinear Features Model.} 
Analytic solutions for three key susceptibilities as a function of $\alpha_p = N_p/M$ and $\alpha_f = N_f/M$.
{\bf(a)-(b)} The susceptibility $\nu$ measures the sensitivity of the fit parameters with respect to small perturbations in the gradient. 
In the small $\lambda$ limit, we make the approximation ${\nu \approx \lambda^{-1}\nu_{-1} + \nu_0}$.
{\bf(a)} The coefficient $\nu_{-1}$ characterizes over-parameterization, equal to the the fraction of fit parameters in excess of that needed to achieve zero training error.
{\bf(b)} The coefficient $\nu_0$ characterizes over-fitting, diverging at the interpolation threshold when $Z^TZ$ has a small eigenvalue.
{\bf(c)} The susceptibility $\chi$ measures the sensitivity of the residual label errors of the training set to small perturbations in the label noise. 
As a result,  $\chi$ characterizes interpolation, equal to the fraction of data points that would need to be removed from the training set to achieve zero training error.
{\bf(d)}  The susceptibility $\kappa$ measures the sensitivity of the residual parameter errors to small perturbations in the ground truth parameters.
We observe that $\kappa$ decreases as a model becomes less biased, 
indicating that the model is better able to express the relationships underlying the data.
In each panel, a black dashed line marks the boundary between the under- and over-parameterized regimes at $\alpha_p = 1$.
}\label{fig:sus}
\end{figure}

Here, we discuss the roles of the susceptibilities that naturally arise as part of our cavity calculations.
In many physical systems, susceptibilities are quantities of interest that measure the effects on a system due to small perturbations. 
In particular, the susceptibilities in our models each characterize a different type of perturbation and in doing so, a different aspect of the double-descent phenomenon.
Setting the gradient equation in Eq.~\eqref{eq:grad} equal to a small nonzero field $\vbeta$, such that $\pdv*{L}{\hbw} = \vbeta$, each of the key susceptibilities in our derivations can be expressed as the trace of a corresponding susceptibility matrix,
\begin{equation}
\begin{gathered}
\nu = \frac{1}{N_p}\Tr  \pdv{\hbw}{\vbeta}\qqc \chi = \frac{1}{M}\Tr  \pdv{\Delta \vby}{\vbeps}\\
\kappa = \frac{1}{N_f}\Tr\pdv{\Delta \vbbeta}{\vbbeta}.\label{eq:sus}
\end{gathered}
\end{equation}
In Fig.~\ref{fig:sus}, we plot each of these quantities as a function of $\alpha_p$ and $\alpha_f$  for the random nonlinear features model (see Appendix for analytic expressions).

The susceptibility $\nu$ measures perturbations to the fit parameters $\hbw$ due to small changes in the gradient $\vbeta$.
In the small $\lambda$ limit, we make the approximation $\nu \approx \lambda^{-1} \nu_{-1} + \nu_0$ and find that the coefficient of each term has a different interpretation.
The first coefficient $\nu_{-1}$, shown in Fig.~\ref{fig:sus}(a), characterizes over-parameterization, counting the fraction of fit parameters in excess of that needed to achieve zero training error.
Since these degrees of freedom are effectively unconstrained in the small $\lambda$ limit, this term diverges as $\lambda$ approaches zero.
The second coefficient $\nu_0$, shown in Fig.~\ref{fig:sus}(b), characterizes over-fitting and diverges at the interpolation threshold in concert with the variance when $Z^TZ$ has a small eigenvalue.
We note that $\nu$ is actually the trace of the inverse Hessian of the loss function in Eq.~\eqref{eq:loss}, or is equivalently the Green's function, and can be used to extract the eigenvalue spectrum of the Hessian matrix~\cite{Cui2020}.

The second susceptibility $\chi$, shown in ig.~\ref{fig:sus}(c), measures the sensitivity of the residual label errors of the training set $\Delta \vby$ to small changes in the label noise $\vbeps$. 
We observe that $\chi$ goes to zero at the interpolation threshold and remains zero in the interpolation regime.
Accordingly,  $\chi$ characterizes interpolation by measuring the fractions of data points that would need to be removed from the training set to achieve zero training error.

Finally, $\kappa$, shown in ig.~\ref{fig:sus}(d), measures the sensitivity of the residual parameter errors $\Delta \vbbeta$ to small changes in the underlying ground truth parameters $\vbbeta$.
We observe that $\kappa$ decreases as the model becomes less biased, indicating that the model is better able to express the relationships underlying the data
(the relationship of $\kappa$ to the bias is explored in more detail Ref.~\cite{Rocks2021}).

\subsection{Non-standard bias-variance decompositions lead to incorrect interpretations of double-descent}\label{sec:nonstandard}

The analytical results for bias and variance for the random nonlinear features model extend the classical understanding of generalization into a modern setting. 
While the model exhibits a classical bias-variance trade-off in the under-parameterized regime, 
in the over-parameterized regime the test error decreases monotonically due to a monotonic reduction in both bias and variance, 
even in the absence of regularization. 
In other words, the benefits of over-parameterization are two-fold:
it can reduce the likelihood of overfitting the training data, 
while simultaneously improving a model’s ability to capture trends hidden in the data.

The alternative and varying interpretations of the double-descent phenomenon found in previous studies are
a direct result of the use of non-standard bias-variance decompositions, 
highlighting the importance of using the historical definitions when using these quantities to interpret double-descent. 
Much of this confusion can be attributed to the precise definition of what we call the \textit{sampling average} in our definitions for bias and variance,
which captures the randomness associated with sampling the training data $\mathcal{D}$.
Previous studies have deviated from these standard definitions in two ways (see Sec.~S6 of the Supporting Information~\cite{supporting_info} for numerical comparisons of these alternatives with the standard definitions).

The first is the so-called fixed-design setting in which the design matrix $X$ is not included in the sampling average~\cite{Adlam2020, Ba2020, Derezinski2020, Hastie2019, Jacot2020, Li2020, Mei2019}.
By holding the design matrix fixed for the training set, but not the test set, an effective mismatch arises between their respective data distributions, introducing an additional source of bias.
As a result, one finds that the bias of the random nonlinear features model diverges at the transition, suggesting the model does not display a classical bias-variance trade-off in the under-parameterized regime, despite exhibiting a U-shaped test error~\cite{Adlam2020, Ba2020, Mei2019}. 

In the second non-standard formulation, the random initialization of the hidden layer is included as part of the sampling average~\cite{Adlam2020, DAscoli2020, DAscoli2020a, Jacot2020, Lin2020, Yang2020}. Consequently, the bias in this setting can be interpreted as measuring the bias of an ensemble model, ${\hat{y}_{\mathrm{ens}}(\vbx) = \E[W][\hat{y}(\vbx)]}$, composed of an average over all possible models with different matrices $W$, rather than the actual model under consideration.
In this setting, the bias misses a contribution that would normally be incurred due to the reality that the model only utilizes a single instance of the matrix $W$, 
rather than averaging over the entire ensemble.
In this setting, one finds that the bias of the random nonlinear features model decreases to a minimum at the interpolation threshold and then remains constant into the over-parameterized regime 
-- paradoxically suggesting that the ability of a model to express complex relationships stops increasing once one reaches the interpolation transition~\cite{Adlam2020, DAscoli2020, DAscoli2020a, Lin2020}. 

In contrast to these two scenarios, we find that the bias of the nonlinear random features model monotonically decreases with the number of parameters in both the under- \textit{and} over-parameterized regimes, suggesting that adding parameters while holding the number of input features fixed always increases the ability a model to capture trends in the data. 
This means that while there is a trade-off between bias and variance in the under-parameterized regime, 
this trade-off disappears in the over-parameterized regime where both bias and variance decrease as one adds fit parameters.

\section{Conclusions}\label{sec:conclude}

Understanding how the bias-variance trade-off manifests in over-parameterized models where the number of fit parameters
far exceeds the  number of data points is a fundamental problem in modern statistics and machine learning. 
Here, we have used the zero-temperature cavity method, a technique from statistical physics, to derive exact analytic expressions for the training error, test error, bias, and variance in the thermodynamic limit for two minimal model architectures: linear regression (no basis functions)  and the random nonlinear features model (a two-layer neural network with nonlinear activation functions where only the top layer is trained). 
These analytic expressions, when combined with numerical simulations, help explain one of the most puzzling features of modern ML methods: the ability to generalize well while simultaneously achieving zero error on the training data.

We observe this phenomenon of ``memorizing without overfitting'' in both models. 
Importantly, our results show that this ability to generalize is not unique to modern ML methods such as those employed in Deep Learning;
both models we consider here are convex. We also note that we do not employ commonly used methods such as stochastic gradient descent to train our models. 
Instead, we use a straightforward regularization procedure based on an $L_2$ penalty and even work in the limit where the strength of the regularization is sent to zero. 
This shows that the ability to generalize while achieving zero training error, sometimes referred to as interpolation, seems to be a generic property of even the simplest over-parameterized models such as linear regression and does not require any special training, regularization, or initialization methods.

Our results show that in stark contrast with the kinds of models considered in classical statistics, 
the variance of over-parameterized models reaches a maximum at the interpolation threshold (the model complexity at which one can achieve zero error on the training data set) and then surprisingly decreasing with model complexity beyond this threshold, giving rise to the double-descent phenomenon.
These large variances at the interpolation threshold are directly tied to the existence of small eigenvalues in the Hessian matrix,
which can be interpreted as a symptom of poor sampling of the data distribution by the training set when viewed by the model through the hidden features.
In addition, over-parameterized models can introduce new sources of bias.
Bias can arise not only from a mismatch between the model and the underlying data distribution, but also from training data sets that span only a subset of the data's feature space. 
Over-parameterized models with bias can also mistake signal for noise, resulting in a nonzero variance even in the absence of noise. 
This shows that the properties over-parameterized models are governed by a subtle interplay between model architecture and random sampling of the data distribution via the training data set.

We note that our models are limited in two significant ways: (i) we focus on the ``lazy regime'' in which the kernel remains fixed during optimization and (ii) we consider convex loss landscapes containing unique solutions.
In contrast, Deep Learning models in practical settings often exhibit highly non-convex loss landscapes and exist in the ``feature regime'' where their kernels evolve to better express the data.
Many questions remain regarding the relationship between these two properties: how do neural networks learn ``good'' sets of features via their kernels and how do such choices relate to different local minima in the overall loss landscape?
Recent work suggests that in this more complex setting, generalization error may be improved by looking for wider, more representative minima in the landscape~\cite{Chaudhari2019, Baldassi2020, Pittorino2020}.
Understanding bias, variance, and generalization in the context of non-convex fitting functions and the relationship of these quantities to the width and local entropy of minima represents an important future area of investigation.

One possible direction for exploring these ideas may be to exploit the relationship between wide neural nets and Gaussian processes~\cite{Jacot2018,Lee2019, Yaida2020} and explore how the spectrum changes with the properties of various minima. 
Alternatively, one could apply our analytical approach to study fixed kernel methods in non-convex settings.
For example, the perceptron exhibits a non-convex loss landscape by including negative constraint cutoffs and can be solved analytically by utilizing a Replica Symmetry Breaking ansatz~\cite{Franz2017}.
In principle, it should be possible to extend these calculations to compute the bias-variance decomposition, eigenvalue spectrum, and susceptibilities with and without basis functions.

Finally, our analysis suggests that our conclusions may be tested directly in practical settings in two ways. 
First, it would be instructive to compute the eigenvalue spectra of the Hessian of Deep Learning models.
It is known that the eigenvalue spectra of neural networks in the over-parameterized regime exhibit a gap with a large number of eigenvalues clustered around zero and the rest located in a nonzero bulk~\cite{Sagun2018}.
However, there has not been a comprehensive study of how the spectrum evolves as one advances through the interpolation threshold.
Second, it would be interesting to compute the relevant susceptibilities, such as those in Eq.~\eqref{eq:sus}.
While we do not expect the susceptibilities of Deep Learning models to quantitatively match those computed here, we do expect them to follow the qualitative behavior exhibited in Fig.~\ref{fig:sus}.
These susceptibilities could be computed by utilizing their matrix forms (e.g., $\nu$ is the trace of the inverse Hessian), or by calculating the linear response directly via efficient differentiation techniques originally developed for computing gradients for meta-learning~\cite{Finn2017}.
Examining such susceptibilities may also prove useful in understanding the nature of Deep Learning models in the feature regime and non-convex loss landscapes.

\section*{Acknowledgments}
We would like to thank Robert Marsland III for extremely useful discussions. 
This work was supported by NIH NIGMS grant 1R35GM119461 and a Simons Investigator in the Mathematical Modeling of Living Systems (MMLS) award to PM. 
The authors also acknowledge support from the Shared Computing Cluster administered by Boston University Research Computing Services.


\begin{thebibliography}{65}%
\makeatletter
\providecommand \@ifxundefined [1]{%
 \@ifx{#1\undefined}
}%
\providecommand \@ifnum [1]{%
 \ifnum #1\expandafter \@firstoftwo
 \else \expandafter \@secondoftwo
 \fi
}%
\providecommand \@ifx [1]{%
 \ifx #1\expandafter \@firstoftwo
 \else \expandafter \@secondoftwo
 \fi
}%
\providecommand \natexlab [1]{#1}%
\providecommand \enquote  [1]{``#1''}%
\providecommand \bibnamefont  [1]{#1}%
\providecommand \bibfnamefont [1]{#1}%
\providecommand \citenamefont [1]{#1}%
\providecommand \href@noop [0]{\@secondoftwo}%
\providecommand \href [0]{\begingroup \@sanitize@url \@href}%
\providecommand \@href[1]{\@@startlink{#1}\@@href}%
\providecommand \@@href[1]{\endgroup#1\@@endlink}%
\providecommand \@sanitize@url [0]{\catcode `\\12\catcode `\$12\catcode
  `\&12\catcode `\#12\catcode `\^12\catcode `\_12\catcode `\%12\relax}%
\providecommand \@@startlink[1]{}%
\providecommand \@@endlink[0]{}%
\providecommand \url  [0]{\begingroup\@sanitize@url \@url }%
\providecommand \@url [1]{\endgroup\@href {#1}{\urlprefix }}%
\providecommand \urlprefix  [0]{URL }%
\providecommand \Eprint [0]{\href }%
\providecommand \doibase [0]{http://dx.doi.org/}%
\providecommand \selectlanguage [0]{\@gobble}%
\providecommand \bibinfo  [0]{\@secondoftwo}%
\providecommand \bibfield  [0]{\@secondoftwo}%
\providecommand \translation [1]{[#1]}%
\providecommand \BibitemOpen [0]{}%
\providecommand \bibitemStop [0]{}%
\providecommand \bibitemNoStop [0]{.\EOS\space}%
\providecommand \EOS [0]{\spacefactor3000\relax}%
\providecommand \BibitemShut  [1]{\csname bibitem#1\endcsname}%
\let\auto@bib@innerbib\@empty
\bibitem [{\citenamefont {Lecun}\ \emph {et~al.}(2015)\citenamefont {Lecun},
  \citenamefont {Bengio},\ and\ \citenamefont {Hinton}}]{Lecun2015}%
  \BibitemOpen
  \bibfield  {author} {\bibinfo {author} {\bibfnamefont {Yann}\ \bibnamefont
  {Lecun}}, \bibinfo {author} {\bibfnamefont {Yoshua}\ \bibnamefont {Bengio}},
  \ and\ \bibinfo {author} {\bibfnamefont {Geoffrey}\ \bibnamefont {Hinton}},\
  }\bibfield  {title} {\enquote {\bibinfo {title} {{Deep learning}},}\ }\href
  {\doibase 10.1038/nature14539} {\bibfield  {journal} {\bibinfo  {journal}
  {Nature}\ }\textbf {\bibinfo {volume} {521}},\ \bibinfo {pages} {436--444}
  (\bibinfo {year} {2015})}\BibitemShut {NoStop}%
\bibitem [{\citenamefont {Canziani}\ \emph {et~al.}(2017)\citenamefont
  {Canziani}, \citenamefont {Paszke},\ and\ \citenamefont
  {Culurciello}}]{Canziani2016}%
  \BibitemOpen
  \bibfield  {author} {\bibinfo {author} {\bibfnamefont {Alfredo}\ \bibnamefont
  {Canziani}}, \bibinfo {author} {\bibfnamefont {Adam}\ \bibnamefont {Paszke}},
  \ and\ \bibinfo {author} {\bibfnamefont {Eugenio}\ \bibnamefont
  {Culurciello}},\ }\bibfield  {title} {\enquote {\bibinfo {title} {{An
  Analysis of Deep Neural Network Models for Practical Applications}},}\ }\href
  {http://arxiv.org/abs/1605.07678} {\  (\bibinfo {year} {2017})},\ \Eprint
  {http://arxiv.org/abs/1605.07678} {arXiv:1605.07678} \BibitemShut {NoStop}%
\bibitem [{\citenamefont {Zhang}\ \emph {et~al.}(2017)\citenamefont {Zhang},
  \citenamefont {Bengio}, \citenamefont {Hardt}, \citenamefont {Recht},\ and\
  \citenamefont {Vinyals}}]{Zhang2017}%
  \BibitemOpen
  \bibfield  {author} {\bibinfo {author} {\bibfnamefont {Chiyuan}\ \bibnamefont
  {Zhang}}, \bibinfo {author} {\bibfnamefont {Samy}\ \bibnamefont {Bengio}},
  \bibinfo {author} {\bibfnamefont {Moritz}\ \bibnamefont {Hardt}}, \bibinfo
  {author} {\bibfnamefont {Benjamin}\ \bibnamefont {Recht}}, \ and\ \bibinfo
  {author} {\bibfnamefont {Oriol}\ \bibnamefont {Vinyals}},\ }\bibfield
  {title} {\enquote {\bibinfo {title} {{Understanding Deep Learning Requires
  Re-thinking Generalization}},}\ }\href
  {https://openreview.net/forum?id=Sy8gdB9xx&;amp;noteId=Sy8gdB9xx} {\bibfield
  {journal} {\bibinfo  {journal} {International Conference on Learning
  Representations (ICLR)}\ } (\bibinfo {year} {2017})}\BibitemShut {NoStop}%
\bibitem [{\citenamefont {Mehta}\ \emph
  {et~al.}(2019{\natexlab{a}})\citenamefont {Mehta}, \citenamefont {Bukov},
  \citenamefont {Wang}, \citenamefont {Day}, \citenamefont {Richardson},
  \citenamefont {Fisher},\ and\ \citenamefont {Schwab}}]{Mehta2019}%
  \BibitemOpen
  \bibfield  {author} {\bibinfo {author} {\bibfnamefont {Pankaj}\ \bibnamefont
  {Mehta}}, \bibinfo {author} {\bibfnamefont {Marin}\ \bibnamefont {Bukov}},
  \bibinfo {author} {\bibfnamefont {Ching~Hao}\ \bibnamefont {Wang}}, \bibinfo
  {author} {\bibfnamefont {Alexandre~G.R.}\ \bibnamefont {Day}}, \bibinfo
  {author} {\bibfnamefont {Clint}\ \bibnamefont {Richardson}}, \bibinfo
  {author} {\bibfnamefont {Charles~K.}\ \bibnamefont {Fisher}}, \ and\ \bibinfo
  {author} {\bibfnamefont {David~J.}\ \bibnamefont {Schwab}},\ }\bibfield
  {title} {\enquote {\bibinfo {title} {{A high-bias, low-variance introduction
  to Machine Learning for physicists}},}\ }\href {\doibase
  10.1016/j.physrep.2019.03.001} {\bibfield  {journal} {\bibinfo  {journal}
  {Physics Reports}\ }\textbf {\bibinfo {volume} {810}},\ \bibinfo {pages}
  {1--124} (\bibinfo {year} {2019}{\natexlab{a}})}\BibitemShut {NoStop}%
\bibitem [{sup()}]{supporting_info}%
  \BibitemOpen
  \bibfield  {title} {\enquote {\bibinfo {title} {See supplemental material at
  [url] for complete analytic derivations and additional numerical results.}}\
  }\href@noop {} {\ }\BibitemShut {NoStop}%
\bibitem [{\citenamefont {Belkin}\ \emph {et~al.}(2019)\citenamefont {Belkin},
  \citenamefont {Hsu}, \citenamefont {Ma},\ and\ \citenamefont
  {Mandal}}]{Belkin2019}%
  \BibitemOpen
  \bibfield  {author} {\bibinfo {author} {\bibfnamefont {Mikhail}\ \bibnamefont
  {Belkin}}, \bibinfo {author} {\bibfnamefont {Daniel}\ \bibnamefont {Hsu}},
  \bibinfo {author} {\bibfnamefont {Siyuan}\ \bibnamefont {Ma}}, \ and\
  \bibinfo {author} {\bibfnamefont {Soumik}\ \bibnamefont {Mandal}},\
  }\bibfield  {title} {\enquote {\bibinfo {title} {{Reconciling modern
  machine-learning practice and the classical bias–variance trade-off}},}\
  }\href {\doibase 10.1073/pnas.1903070116} {\bibfield  {journal} {\bibinfo
  {journal} {Proceedings of the National Academy of Sciences}\ }\textbf
  {\bibinfo {volume} {116}},\ \bibinfo {pages} {15849--15854} (\bibinfo {year}
  {2019})}\BibitemShut {NoStop}%
\bibitem [{\citenamefont {Geiger}\ \emph {et~al.}(2019)\citenamefont {Geiger},
  \citenamefont {Spigler}, \citenamefont {D'Ascoli}, \citenamefont {Sagun},
  \citenamefont {Baity-Jesi}, \citenamefont {Biroli},\ and\ \citenamefont
  {Wyart}}]{Geiger2019}%
  \BibitemOpen
  \bibfield  {author} {\bibinfo {author} {\bibfnamefont {Mario}\ \bibnamefont
  {Geiger}}, \bibinfo {author} {\bibfnamefont {Stefano}\ \bibnamefont
  {Spigler}}, \bibinfo {author} {\bibfnamefont {St{\'{e}}phane}\ \bibnamefont
  {D'Ascoli}}, \bibinfo {author} {\bibfnamefont {Levent}\ \bibnamefont
  {Sagun}}, \bibinfo {author} {\bibfnamefont {Marco}\ \bibnamefont
  {Baity-Jesi}}, \bibinfo {author} {\bibfnamefont {Giulio}\ \bibnamefont
  {Biroli}}, \ and\ \bibinfo {author} {\bibfnamefont {Matthieu}\ \bibnamefont
  {Wyart}},\ }\bibfield  {title} {\enquote {\bibinfo {title} {{Jamming
  transition as a paradigm to understand the loss landscape of deep neural
  networks}},}\ }\href {\doibase 10.1103/PhysRevE.100.012115} {\bibfield
  {journal} {\bibinfo  {journal} {Physical Review E}\ }\textbf {\bibinfo
  {volume} {100}},\ \bibinfo {pages} {012115} (\bibinfo {year}
  {2019})}\BibitemShut {NoStop}%
\bibitem [{\citenamefont {Spigler}\ \emph {et~al.}(2019)\citenamefont
  {Spigler}, \citenamefont {Geiger}, \citenamefont {D'Ascoli}, \citenamefont
  {Sagun}, \citenamefont {Biroli},\ and\ \citenamefont {Wyart}}]{Spigler2019}%
  \BibitemOpen
  \bibfield  {author} {\bibinfo {author} {\bibfnamefont {S.}~\bibnamefont
  {Spigler}}, \bibinfo {author} {\bibfnamefont {M.}~\bibnamefont {Geiger}},
  \bibinfo {author} {\bibfnamefont {S}~\bibnamefont {D'Ascoli}}, \bibinfo
  {author} {\bibfnamefont {L.}~\bibnamefont {Sagun}}, \bibinfo {author}
  {\bibfnamefont {G.}~\bibnamefont {Biroli}}, \ and\ \bibinfo {author}
  {\bibfnamefont {M.}~\bibnamefont {Wyart}},\ }\bibfield  {title} {\enquote
  {\bibinfo {title} {{A jamming transition from under- to over-parametrization
  affects generalization in deep learning}},}\ }\href {\doibase
  10.1088/1751-8121/ab4c8b} {\bibfield  {journal} {\bibinfo  {journal} {Journal
  of Physics A: Mathematical and Theoretical}\ }\textbf {\bibinfo {volume}
  {52}},\ \bibinfo {pages} {474001} (\bibinfo {year} {2019})}\BibitemShut
  {NoStop}%
\bibitem [{\citenamefont {Mailman}\ and\ \citenamefont
  {Chakraborty}(2011)}]{Geiger2020}%
  \BibitemOpen
  \bibfield  {author} {\bibinfo {author} {\bibfnamefont {M}~\bibnamefont
  {Mailman}}\ and\ \bibinfo {author} {\bibfnamefont {B}~\bibnamefont
  {Chakraborty}},\ }\bibfield  {title} {\enquote {\bibinfo {title} {{A
  signature of a thermodynamic phase transition in jammed granular packings:
  growing correlations in force space}},}\ }\href {\doibase
  10.1088/1742-5468/2011/07/L07002} {\bibfield  {journal} {\bibinfo  {journal}
  {Journal of Statistical Mechanics: Theory and Experiment}\ }\textbf {\bibinfo
  {volume} {2011}},\ \bibinfo {pages} {L07002} (\bibinfo {year}
  {2011})}\BibitemShut {NoStop}%
\bibitem [{\citenamefont {Bahri}\ \emph {et~al.}(2020)\citenamefont {Bahri},
  \citenamefont {Kadmon}, \citenamefont {Pennington}, \citenamefont
  {Schoenholz}, \citenamefont {Sohl-Dickstein},\ and\ \citenamefont
  {Ganguli}}]{Bahri2020}%
  \BibitemOpen
  \bibfield  {author} {\bibinfo {author} {\bibfnamefont {Yasaman}\ \bibnamefont
  {Bahri}}, \bibinfo {author} {\bibfnamefont {Jonathan}\ \bibnamefont
  {Kadmon}}, \bibinfo {author} {\bibfnamefont {Jeffrey}\ \bibnamefont
  {Pennington}}, \bibinfo {author} {\bibfnamefont {Sam~S.}\ \bibnamefont
  {Schoenholz}}, \bibinfo {author} {\bibfnamefont {Jascha}\ \bibnamefont
  {Sohl-Dickstein}}, \ and\ \bibinfo {author} {\bibfnamefont {Surya}\
  \bibnamefont {Ganguli}},\ }\bibfield  {title} {\enquote {\bibinfo {title}
  {{Statistical Mechanics of Deep Learning}},}\ }\href {\doibase
  10.1146/annurev-conmatphys-031119-050745} {\bibfield  {journal} {\bibinfo
  {journal} {Annual Review of Condensed Matter Physics}\ }\textbf {\bibinfo
  {volume} {11}},\ \bibinfo {pages} {501--528} (\bibinfo {year}
  {2020})}\BibitemShut {NoStop}%
\bibitem [{\citenamefont {{Kobak [@hippopedoid]}}(2020)}]{Kobak2020}%
  \BibitemOpen
  \bibfield  {author} {\bibinfo {author} {\bibfnamefont {Dmitry}\ \bibnamefont
  {{Kobak [@hippopedoid]}}},\ }\href
  {https://twitter.com/hippopedoid/status/1243229021921579010} {\enquote
  {\bibinfo {title} {{Twitter Thread:
  twitter.com/hippopedoid/status/1243229021921579010}},}\ } (\bibinfo {year}
  {2020})\BibitemShut {NoStop}%
\bibitem [{\citenamefont {Loog}\ \emph {et~al.}(2020)\citenamefont {Loog},
  \citenamefont {Viering}, \citenamefont {Mey}, \citenamefont {Krijthe},\ and\
  \citenamefont {Tax}}]{Loog2020}%
  \BibitemOpen
  \bibfield  {author} {\bibinfo {author} {\bibfnamefont {Marco}\ \bibnamefont
  {Loog}}, \bibinfo {author} {\bibfnamefont {Tom}\ \bibnamefont {Viering}},
  \bibinfo {author} {\bibfnamefont {Alexander}\ \bibnamefont {Mey}}, \bibinfo
  {author} {\bibfnamefont {Jesse~H.}\ \bibnamefont {Krijthe}}, \ and\ \bibinfo
  {author} {\bibfnamefont {David M.~J.}\ \bibnamefont {Tax}},\ }\bibfield
  {title} {\enquote {\bibinfo {title} {{A brief prehistory of double
  descent}},}\ }\href {\doibase 10.1073/pnas.2001875117} {\bibfield  {journal}
  {\bibinfo  {journal} {Proceedings of the National Academy of Sciences}\
  }\textbf {\bibinfo {volume} {117}},\ \bibinfo {pages} {10625--10626}
  (\bibinfo {year} {2020})}\BibitemShut {NoStop}%
\bibitem [{\citenamefont {Nakkiran}\ \emph {et~al.}(2020)\citenamefont
  {Nakkiran}, \citenamefont {Kaplun}, \citenamefont {Bansal}, \citenamefont
  {Yang}, \citenamefont {Barak},\ and\ \citenamefont
  {Sutskever}}]{Nakkiran2020}%
  \BibitemOpen
  \bibfield  {author} {\bibinfo {author} {\bibfnamefont {Preetum}\ \bibnamefont
  {Nakkiran}}, \bibinfo {author} {\bibfnamefont {Gal}\ \bibnamefont {Kaplun}},
  \bibinfo {author} {\bibfnamefont {Yamini}\ \bibnamefont {Bansal}}, \bibinfo
  {author} {\bibfnamefont {Tristan}\ \bibnamefont {Yang}}, \bibinfo {author}
  {\bibfnamefont {Boaz}\ \bibnamefont {Barak}}, \ and\ \bibinfo {author}
  {\bibfnamefont {Ilya}\ \bibnamefont {Sutskever}},\ }\bibfield  {title}
  {\enquote {\bibinfo {title} {{Deep Double Descent: Where Bigger Models and
  More Data Hurt}},}\ }\href {https://openreview.net/forum?id=B1g5sA4twr}
  {\bibfield  {journal} {\bibinfo  {journal} {International Conference on
  Learning Representations (ICLR)}\ } (\bibinfo {year} {2020})}\BibitemShut
  {NoStop}%
\bibitem [{\citenamefont {Jacot}\ \emph {et~al.}(2018)\citenamefont {Jacot},
  \citenamefont {Gabriel},\ and\ \citenamefont {Hongler}}]{Jacot2018}%
  \BibitemOpen
  \bibfield  {author} {\bibinfo {author} {\bibfnamefont {Arthur}\ \bibnamefont
  {Jacot}}, \bibinfo {author} {\bibfnamefont {Franck}\ \bibnamefont {Gabriel}},
  \ and\ \bibinfo {author} {\bibfnamefont {Cl{\'{e}}ment}\ \bibnamefont
  {Hongler}},\ }\bibfield  {title} {\enquote {\bibinfo {title} {{Neural tangent
  kernel: Convergence and generalization in neural networks}},}\ }\href
  {https://proceedings.neurips.cc/paper/2018/file/5a4be1fa34e62bb8a6ec6b91d2462f5a-Paper.pdf}
  {\bibfield  {journal} {\bibinfo  {journal} {Advances in Neural Information
  Processing Systems (NeurIPS)}\ }\textbf {\bibinfo {volume} {31}} (\bibinfo
  {year} {2018})}\BibitemShut {NoStop}%
\bibitem [{\citenamefont {Lee}\ \emph {et~al.}(2019)\citenamefont {Lee},
  \citenamefont {Xiao}, \citenamefont {Schoenholz}, \citenamefont {Novak},
  \citenamefont {Sohl-Dickstein},\ and\ \citenamefont {Pennington}}]{Lee2019}%
  \BibitemOpen
  \bibfield  {author} {\bibinfo {author} {\bibfnamefont {Jaehoon}\ \bibnamefont
  {Lee}}, \bibinfo {author} {\bibfnamefont {Lechao}\ \bibnamefont {Xiao}},
  \bibinfo {author} {\bibfnamefont {Samuel~S.}\ \bibnamefont {Schoenholz}},
  \bibinfo {author} {\bibfnamefont {Yasaman Bahri~Roman}\ \bibnamefont
  {Novak}}, \bibinfo {author} {\bibfnamefont {Jascha}\ \bibnamefont
  {Sohl-Dickstein}}, \ and\ \bibinfo {author} {\bibfnamefont {Jeffrey}\
  \bibnamefont {Pennington}},\ }\bibfield  {title} {\enquote {\bibinfo {title}
  {{Wide neural networks of any depth evolve as linear models under gradient
  descent}},}\ }\href
  {https://proceedings.neurips.cc/paper/2019/file/0d1a9651497a38d8b1c3871c84528bd4-Paper.pdf}
  {\bibfield  {journal} {\bibinfo  {journal} {Advances in Neural Information
  Processing Systems (NeurIPS)}\ }\textbf {\bibinfo {volume} {32}} (\bibinfo
  {year} {2019})}\BibitemShut {NoStop}%
\bibitem [{\citenamefont {Geiger}\ \emph {et~al.}(2020)\citenamefont {Geiger},
  \citenamefont {Spigler}, \citenamefont {Jacot},\ and\ \citenamefont
  {Wyart}}]{Geiger2020b}%
  \BibitemOpen
  \bibfield  {author} {\bibinfo {author} {\bibfnamefont {Mario}\ \bibnamefont
  {Geiger}}, \bibinfo {author} {\bibfnamefont {Stefano}\ \bibnamefont
  {Spigler}}, \bibinfo {author} {\bibfnamefont {Arthur}\ \bibnamefont {Jacot}},
  \ and\ \bibinfo {author} {\bibfnamefont {Matthieu}\ \bibnamefont {Wyart}},\
  }\bibfield  {title} {\enquote {\bibinfo {title} {{Disentangling feature and
  lazy training in deep neural networks}},}\ }\href {\doibase
  10.1088/1742-5468/abc4de} {\bibfield  {journal} {\bibinfo  {journal} {Journal
  of Statistical Mechanics: Theory and Experiment}\ }\textbf {\bibinfo {volume}
  {2020}},\ \bibinfo {pages} {113301} (\bibinfo {year} {2020})}\BibitemShut
  {NoStop}%
\bibitem [{\citenamefont {Adlam}\ and\ \citenamefont
  {Pennington}(2020)}]{Adlam2020}%
  \BibitemOpen
  \bibfield  {author} {\bibinfo {author} {\bibfnamefont {Ben}\ \bibnamefont
  {Adlam}}\ and\ \bibinfo {author} {\bibfnamefont {Jeffrey}\ \bibnamefont
  {Pennington}},\ }\bibfield  {title} {\enquote {\bibinfo {title}
  {{Understanding double descent requires a fine-grained bias-variance
  decomposition}},}\ }\href
  {https://proceedings.neurips.cc/paper/2020/file/7d420e2b2939762031eed0447a9be19f-Paper.pdf}
  {\bibfield  {journal} {\bibinfo  {journal} {Advances in Neural Information
  Processing Systems (NeurIPS)}\ }\textbf {\bibinfo {volume} {33}},\ \bibinfo
  {pages} {11022--11032} (\bibinfo {year} {2020})}\BibitemShut {NoStop}%
\bibitem [{\citenamefont {Advani}\ \emph {et~al.}(2020)\citenamefont {Advani},
  \citenamefont {Saxe},\ and\ \citenamefont {Sompolinsky}}]{Advani2020}%
  \BibitemOpen
  \bibfield  {author} {\bibinfo {author} {\bibfnamefont {Madhu~S.}\
  \bibnamefont {Advani}}, \bibinfo {author} {\bibfnamefont {Andrew~M.}\
  \bibnamefont {Saxe}}, \ and\ \bibinfo {author} {\bibfnamefont {Haim}\
  \bibnamefont {Sompolinsky}},\ }\bibfield  {title} {\enquote {\bibinfo {title}
  {{High-dimensional dynamics of generalization error in neural networks}},}\
  }\href {\doibase 10.1016/j.neunet.2020.08.022} {\bibfield  {journal}
  {\bibinfo  {journal} {Neural Networks}\ }\textbf {\bibinfo {volume} {132}},\
  \bibinfo {pages} {428--446} (\bibinfo {year} {2020})}\BibitemShut {NoStop}%
\bibitem [{\citenamefont {Ba}\ \emph {et~al.}(2020)\citenamefont {Ba},
  \citenamefont {Erdogdu}, \citenamefont {Suzuki}, \citenamefont {Wu},\ and\
  \citenamefont {Zhang}}]{Ba2020}%
  \BibitemOpen
  \bibfield  {author} {\bibinfo {author} {\bibfnamefont {Jimmy}\ \bibnamefont
  {Ba}}, \bibinfo {author} {\bibfnamefont {Murat}\ \bibnamefont {Erdogdu}},
  \bibinfo {author} {\bibfnamefont {Taiji}\ \bibnamefont {Suzuki}}, \bibinfo
  {author} {\bibfnamefont {Denny}\ \bibnamefont {Wu}}, \ and\ \bibinfo {author}
  {\bibfnamefont {Tianzong}\ \bibnamefont {Zhang}},\ }\bibfield  {title}
  {\enquote {\bibinfo {title} {{Generalization of Two-layer Neural Networks: An
  Asymptotic Viewpoint}},}\ }\href {https://openreview.net/forum?id=H1gBsgBYwH}
  {\bibfield  {journal} {\bibinfo  {journal} {International Conference on
  Learning Representations (ICLR)}\ } (\bibinfo {year} {2020})}\BibitemShut
  {NoStop}%
\bibitem [{\citenamefont {Barbier}\ \emph {et~al.}(2019)\citenamefont
  {Barbier}, \citenamefont {Krzakala}, \citenamefont {Macris}, \citenamefont
  {Miolane},\ and\ \citenamefont {Zdeborov{\'{a}}}}]{Barbier2019}%
  \BibitemOpen
  \bibfield  {author} {\bibinfo {author} {\bibfnamefont {Jean}\ \bibnamefont
  {Barbier}}, \bibinfo {author} {\bibfnamefont {Florent}\ \bibnamefont
  {Krzakala}}, \bibinfo {author} {\bibfnamefont {Nicolas}\ \bibnamefont
  {Macris}}, \bibinfo {author} {\bibfnamefont {L{\'{e}}o}\ \bibnamefont
  {Miolane}}, \ and\ \bibinfo {author} {\bibfnamefont {Lenka}\ \bibnamefont
  {Zdeborov{\'{a}}}},\ }\bibfield  {title} {\enquote {\bibinfo {title}
  {{Optimal errors and phase transitions in high-dimensional generalized linear
  models}},}\ }\href {\doibase 10.1073/pnas.1802705116} {\bibfield  {journal}
  {\bibinfo  {journal} {Proceedings of the National Academy of Sciences}\
  }\textbf {\bibinfo {volume} {116}},\ \bibinfo {pages} {5451--5460} (\bibinfo
  {year} {2019})}\BibitemShut {NoStop}%
\bibitem [{\citenamefont {Bartlett}\ \emph {et~al.}(2020)\citenamefont
  {Bartlett}, \citenamefont {Long}, \citenamefont {Lugosi},\ and\ \citenamefont
  {Tsigler}}]{Bartlett2020}%
  \BibitemOpen
  \bibfield  {author} {\bibinfo {author} {\bibfnamefont {Peter~L.}\
  \bibnamefont {Bartlett}}, \bibinfo {author} {\bibfnamefont {Philip~M.}\
  \bibnamefont {Long}}, \bibinfo {author} {\bibfnamefont {G{\'{a}}bor}\
  \bibnamefont {Lugosi}}, \ and\ \bibinfo {author} {\bibfnamefont {Alexander}\
  \bibnamefont {Tsigler}},\ }\bibfield  {title} {\enquote {\bibinfo {title}
  {{Benign overfitting in linear regression}},}\ }\href {\doibase
  10.1073/pnas.1907378117} {\bibfield  {journal} {\bibinfo  {journal}
  {Proceedings of the National Academy of Sciences}\ }\textbf {\bibinfo
  {volume} {117}},\ \bibinfo {pages} {30063--30070} (\bibinfo {year}
  {2020})}\BibitemShut {NoStop}%
\bibitem [{\citenamefont {Belkin}\ \emph {et~al.}(2020)\citenamefont {Belkin},
  \citenamefont {Hsu},\ and\ \citenamefont {Xu}}]{Belkin2020}%
  \BibitemOpen
  \bibfield  {author} {\bibinfo {author} {\bibfnamefont {Mikhail}\ \bibnamefont
  {Belkin}}, \bibinfo {author} {\bibfnamefont {Daniel}\ \bibnamefont {Hsu}}, \
  and\ \bibinfo {author} {\bibfnamefont {Ji}~\bibnamefont {Xu}},\ }\bibfield
  {title} {\enquote {\bibinfo {title} {{Two Models of Double Descent for Weak
  Features}},}\ }\href {\doibase 10.1137/20M1336072} {\bibfield  {journal}
  {\bibinfo  {journal} {SIAM Journal on Mathematics of Data Science}\ }\textbf
  {\bibinfo {volume} {2}},\ \bibinfo {pages} {1167--1180} (\bibinfo {year}
  {2020})}\BibitemShut {NoStop}%
\bibitem [{\citenamefont {Bibas}\ \emph {et~al.}(2019)\citenamefont {Bibas},
  \citenamefont {Fogel},\ and\ \citenamefont {Feder}}]{Bibas2019}%
  \BibitemOpen
  \bibfield  {author} {\bibinfo {author} {\bibfnamefont {Koby}\ \bibnamefont
  {Bibas}}, \bibinfo {author} {\bibfnamefont {Yaniv}\ \bibnamefont {Fogel}}, \
  and\ \bibinfo {author} {\bibfnamefont {Meir}\ \bibnamefont {Feder}},\
  }\bibfield  {title} {\enquote {\bibinfo {title} {{A New Look at an Old
  Problem: A Universal Learning Approach to Linear Regression}},}\ }\href
  {\doibase 10.1109/ISIT.2019.8849398} {\bibfield  {journal} {\bibinfo
  {journal} {IEEE International Symposium on Information Theory (ISIT)}\ ,\
  \bibinfo {pages} {2304--2308}} (\bibinfo {year} {2019})}\BibitemShut
  {NoStop}%
\bibitem [{\citenamefont {Deng}\ \emph {et~al.}(2020)\citenamefont {Deng},
  \citenamefont {Kammoun},\ and\ \citenamefont {Thrampoulidis}}]{Deng2019}%
  \BibitemOpen
  \bibfield  {author} {\bibinfo {author} {\bibfnamefont {Zeyu}\ \bibnamefont
  {Deng}}, \bibinfo {author} {\bibfnamefont {Abla}\ \bibnamefont {Kammoun}}, \
  and\ \bibinfo {author} {\bibfnamefont {Christos}\ \bibnamefont
  {Thrampoulidis}},\ }\bibfield  {title} {\enquote {\bibinfo {title} {{A Model
  of Double Descent for High-Dimensional Logistic Regression}},}\ }\href
  {\doibase 10.1109/ICASSP40776.2020.9053524} {\bibfield  {journal} {\bibinfo
  {journal} {IEEE International Conference on Acoustics, Speech and Signal
  Processing (ICASSP)}\ ,\ \bibinfo {pages} {4267--4271}} (\bibinfo {year}
  {2020})}\BibitemShut {NoStop}%
\bibitem [{\citenamefont {D'Ascoli}\ \emph
  {et~al.}(2020{\natexlab{a}})\citenamefont {D'Ascoli}, \citenamefont {Sagun},\
  and\ \citenamefont {Biroli}}]{DAscoli2020}%
  \BibitemOpen
  \bibfield  {author} {\bibinfo {author} {\bibfnamefont {St{\'{e}}phane}\
  \bibnamefont {D'Ascoli}}, \bibinfo {author} {\bibfnamefont {Levent}\
  \bibnamefont {Sagun}}, \ and\ \bibinfo {author} {\bibfnamefont {Giulio}\
  \bibnamefont {Biroli}},\ }\bibfield  {title} {\enquote {\bibinfo {title}
  {{Triple descent and the two kinds of overfitting: Where \& why do they
  appear?}}}\ }\href
  {https://proceedings.neurips.cc/paper/2020/file/1fd09c5f59a8ff35d499c0ee25a1d47e-Paper.pdf}
  {\bibfield  {journal} {\bibinfo  {journal} {Advances in Neural Information
  Processing Systems (NeurIPS)}\ }\textbf {\bibinfo {volume} {33}} (\bibinfo
  {year} {2020}{\natexlab{a}})}\BibitemShut {NoStop}%
\bibitem [{\citenamefont {D'Ascoli}\ \emph
  {et~al.}(2020{\natexlab{b}})\citenamefont {D'Ascoli}, \citenamefont
  {Refinetti}, \citenamefont {Biroli},\ and\ \citenamefont
  {Krzakala}}]{DAscoli2020a}%
  \BibitemOpen
  \bibfield  {author} {\bibinfo {author} {\bibfnamefont {St{\'{e}}phane}\
  \bibnamefont {D'Ascoli}}, \bibinfo {author} {\bibfnamefont {Maria}\
  \bibnamefont {Refinetti}}, \bibinfo {author} {\bibfnamefont {Giulio}\
  \bibnamefont {Biroli}}, \ and\ \bibinfo {author} {\bibfnamefont {Florent}\
  \bibnamefont {Krzakala}},\ }\bibfield  {title} {\enquote {\bibinfo {title}
  {{Double Trouble in Double Descent: Bias and Variance(s) in the Lazy
  Regime}},}\ }\href
  {http://proceedings.mlr.press/v119/d-ascoli20a/d-ascoli20a.pdf} {\bibfield
  {journal} {\bibinfo  {journal} {Proceedings of the 37th International
  Conference on Machine Learning (ICML)}\ }\bibinfo {series} {PMLR},\ \textbf
  {\bibinfo {volume} {119}},\ \bibinfo {pages} {2280--2290} (\bibinfo {year}
  {2020}{\natexlab{b}})}\BibitemShut {NoStop}%
\bibitem [{\citenamefont {Derezi{\'{n}}ski}\ \emph {et~al.}(2020)\citenamefont
  {Derezi{\'{n}}ski}, \citenamefont {Liang},\ and\ \citenamefont
  {Mahoney}}]{Derezinski2020}%
  \BibitemOpen
  \bibfield  {author} {\bibinfo {author} {\bibfnamefont {Micha{\l}}\
  \bibnamefont {Derezi{\'{n}}ski}}, \bibinfo {author} {\bibfnamefont {Feynman}\
  \bibnamefont {Liang}}, \ and\ \bibinfo {author} {\bibfnamefont {Michael~W.}\
  \bibnamefont {Mahoney}},\ }\bibfield  {title} {\enquote {\bibinfo {title}
  {{Exact expressions for double descent and implicit regularization via
  surrogate random design}},}\ }\href
  {https://proceedings.neurips.cc/paper/2020/file/37740d59bb0eb7b4493725b2e0e5289b-Paper.pdf}
  {\bibfield  {journal} {\bibinfo  {journal} {Advances in Neural Information
  Processing Systems (NeurIPS)}\ }\textbf {\bibinfo {volume} {33}},\ \bibinfo
  {pages} {5152--5164} (\bibinfo {year} {2020})}\BibitemShut {NoStop}%
\bibitem [{\citenamefont {Dhifallah}\ and\ \citenamefont
  {Lu}(2020)}]{Dhifallah2020}%
  \BibitemOpen
  \bibfield  {author} {\bibinfo {author} {\bibfnamefont {Oussama}\ \bibnamefont
  {Dhifallah}}\ and\ \bibinfo {author} {\bibfnamefont {Yue~M.}\ \bibnamefont
  {Lu}},\ }\bibfield  {title} {\enquote {\bibinfo {title} {{A Precise
  Performance Analysis of Learning with Random Features}},}\ }\href
  {http://arxiv.org/abs/2008.11904} {\  (\bibinfo {year} {2020})},\ \Eprint
  {http://arxiv.org/abs/2008.11904} {arXiv:2008.11904} \BibitemShut {NoStop}%
\bibitem [{\citenamefont {Gerace}\ \emph {et~al.}(2020)\citenamefont {Gerace},
  \citenamefont {Loureiro}, \citenamefont {Krzakala}, \citenamefont
  {M{\'{e}}zard},\ and\ \citenamefont {Zdeborov{\'{a}}}}]{Gerace2020}%
  \BibitemOpen
  \bibfield  {author} {\bibinfo {author} {\bibfnamefont {Federica}\
  \bibnamefont {Gerace}}, \bibinfo {author} {\bibfnamefont {Bruno}\
  \bibnamefont {Loureiro}}, \bibinfo {author} {\bibfnamefont {Florent}\
  \bibnamefont {Krzakala}}, \bibinfo {author} {\bibfnamefont {Marc}\
  \bibnamefont {M{\'{e}}zard}}, \ and\ \bibinfo {author} {\bibfnamefont
  {Lenka}\ \bibnamefont {Zdeborov{\'{a}}}},\ }\bibfield  {title} {\enquote
  {\bibinfo {title} {{Generalisation error in learning with random features and
  the hidden manifold model}},}\ }\href
  {http://proceedings.mlr.press/v119/gerace20a/gerace20a.pdf} {\bibfield
  {journal} {\bibinfo  {journal} {Proceedings of the 37th International
  Conference on Machine Learning (ICML)}\ }\bibinfo {series} {PMLR},\ \textbf
  {\bibinfo {volume} {119}},\ \bibinfo {pages} {3452--3462} (\bibinfo {year}
  {2020})}\BibitemShut {NoStop}%
\bibitem [{\citenamefont {Hastie}\ \emph {et~al.}(2019)\citenamefont {Hastie},
  \citenamefont {Montanari}, \citenamefont {Rosset},\ and\ \citenamefont
  {Tibshirani}}]{Hastie2019}%
  \BibitemOpen
  \bibfield  {author} {\bibinfo {author} {\bibfnamefont {Trevor}\ \bibnamefont
  {Hastie}}, \bibinfo {author} {\bibfnamefont {Andrea}\ \bibnamefont
  {Montanari}}, \bibinfo {author} {\bibfnamefont {Saharon}\ \bibnamefont
  {Rosset}}, \ and\ \bibinfo {author} {\bibfnamefont {Ryan~J.}\ \bibnamefont
  {Tibshirani}},\ }\bibfield  {title} {\enquote {\bibinfo {title} {{Surprises
  in High-Dimensional Ridgeless Least Squares Interpolation}},}\ }\href
  {http://arxiv.org/abs/1903.08560} {\  (\bibinfo {year} {2019})},\ \Eprint
  {http://arxiv.org/abs/1903.08560} {arXiv:1903.08560} \BibitemShut {NoStop}%
\bibitem [{\citenamefont {Jacot}\ \emph {et~al.}(2020)\citenamefont {Jacot},
  \citenamefont {Şimşek}, \citenamefont {Spadaro}, \citenamefont {Hongler},\
  and\ \citenamefont {Gabriel}}]{Jacot2020}%
  \BibitemOpen
  \bibfield  {author} {\bibinfo {author} {\bibfnamefont {Arthur}\ \bibnamefont
  {Jacot}}, \bibinfo {author} {\bibfnamefont {Berfin}\ \bibnamefont
  {Şimşek}}, \bibinfo {author} {\bibfnamefont {Francesco}\ \bibnamefont
  {Spadaro}}, \bibinfo {author} {\bibfnamefont {Cl{\'{e}}ment}\ \bibnamefont
  {Hongler}}, \ and\ \bibinfo {author} {\bibfnamefont {Franck}\ \bibnamefont
  {Gabriel}},\ }\bibfield  {title} {\enquote {\bibinfo {title} {{Implicit
  regularization of random feature models}},}\ }\href
  {http://proceedings.mlr.press/v119/jacot20a.html} {\bibfield  {journal}
  {\bibinfo  {journal} {Proceedings of the 37th International Conference on
  Machine Learning (ICML)}\ }\bibinfo {series} {PMLR},\ \textbf {\bibinfo
  {volume} {119}},\ \bibinfo {pages} {4631--4640} (\bibinfo {year}
  {2020})}\BibitemShut {NoStop}%
\bibitem [{\citenamefont {Kini}\ and\ \citenamefont
  {Thrampoulidis}(2020)}]{Kini2020}%
  \BibitemOpen
  \bibfield  {author} {\bibinfo {author} {\bibfnamefont {Ganesh~Ramachandra}\
  \bibnamefont {Kini}}\ and\ \bibinfo {author} {\bibfnamefont {Christos}\
  \bibnamefont {Thrampoulidis}},\ }\bibfield  {title} {\enquote {\bibinfo
  {title} {{Analytic Study of Double Descent in Binary Classification: The
  Impact of Loss}},}\ }\href {\doibase 10.1109/ISIT44484.2020.9174344}
  {\bibfield  {journal} {\bibinfo  {journal} {IEEE International Symposium on
  Information Theory (ISIT)}\ ,\ \bibinfo {pages} {2527--2532}} (\bibinfo
  {year} {2020})}\BibitemShut {NoStop}%
\bibitem [{\citenamefont {Lampinen}\ and\ \citenamefont
  {Ganguli}(2019)}]{Lampinen2019}%
  \BibitemOpen
  \bibfield  {author} {\bibinfo {author} {\bibfnamefont {Andrew~K.}\
  \bibnamefont {Lampinen}}\ and\ \bibinfo {author} {\bibfnamefont {Surya}\
  \bibnamefont {Ganguli}},\ }\bibfield  {title} {\enquote {\bibinfo {title}
  {{An analytic theory of generalization dynamics and transfer learning in deep
  linear networks}},}\ }\href {https://openreview.net/forum?id=ryfMLoCqtQ}
  {\bibfield  {journal} {\bibinfo  {journal} {International Conference on
  Learning Representations (ICLR)}\ } (\bibinfo {year} {2019})}\BibitemShut
  {NoStop}%
\bibitem [{\citenamefont {Li}\ \emph {et~al.}(2020)\citenamefont {Li},
  \citenamefont {Su},\ and\ \citenamefont {Sejdinovic}}]{Li2020}%
  \BibitemOpen
  \bibfield  {author} {\bibinfo {author} {\bibfnamefont {Zhu}\ \bibnamefont
  {Li}}, \bibinfo {author} {\bibfnamefont {Weijie~J.}\ \bibnamefont {Su}}, \
  and\ \bibinfo {author} {\bibfnamefont {Dino}\ \bibnamefont {Sejdinovic}},\
  }\bibfield  {title} {\enquote {\bibinfo {title} {{Benign overfitting and
  noisy features}},}\ }\href@noop {} {\  (\bibinfo {year} {2020})},\ \Eprint
  {http://arxiv.org/abs/2008.02901} {arXiv:2008.02901} \BibitemShut {NoStop}%
\bibitem [{\citenamefont {Liang}\ and\ \citenamefont
  {Rakhlin}(2020)}]{Liang2020}%
  \BibitemOpen
  \bibfield  {author} {\bibinfo {author} {\bibfnamefont {Tengyuan}\
  \bibnamefont {Liang}}\ and\ \bibinfo {author} {\bibfnamefont {Alexander}\
  \bibnamefont {Rakhlin}},\ }\bibfield  {title} {\enquote {\bibinfo {title}
  {{Just interpolate: Kernel “Ridgeless” regression can generalize}},}\
  }\href {\doibase 10.1214/19-AOS1849} {\bibfield  {journal} {\bibinfo
  {journal} {Annals of Statistics}\ }\textbf {\bibinfo {volume} {48}},\
  \bibinfo {pages} {1329--1347} (\bibinfo {year} {2020})}\BibitemShut {NoStop}%
\bibitem [{\citenamefont {Liang}\ \emph {et~al.}(2020)\citenamefont {Liang},
  \citenamefont {Rakhlin},\ and\ \citenamefont {Zhai}}]{Liang2020b}%
  \BibitemOpen
  \bibfield  {author} {\bibinfo {author} {\bibfnamefont {Tengyuan}\
  \bibnamefont {Liang}}, \bibinfo {author} {\bibfnamefont {Alexander}\
  \bibnamefont {Rakhlin}}, \ and\ \bibinfo {author} {\bibfnamefont {Xiyu}\
  \bibnamefont {Zhai}},\ }\bibfield  {title} {\enquote {\bibinfo {title} {{On
  the Multiple Descent of Minimum-Norm Interpolants and Restricted Lower
  Isometry of Kernels}},}\ }\href
  {http://proceedings.mlr.press/v125/liang20a.html} {\bibfield  {journal}
  {\bibinfo  {journal} {Proceedings of Thirty Third Conference on Learning
  Theory}\ }\bibinfo {series} {PMLR},\ \textbf {\bibinfo {volume} {125}},\
  \bibinfo {pages} {2683--2711} (\bibinfo {year} {2020})}\BibitemShut {NoStop}%
\bibitem [{\citenamefont {Liao}\ \emph {et~al.}(2020)\citenamefont {Liao},
  \citenamefont {Couillet},\ and\ \citenamefont {Mahoney}}]{Liao2020}%
  \BibitemOpen
  \bibfield  {author} {\bibinfo {author} {\bibfnamefont {Zhenyu}\ \bibnamefont
  {Liao}}, \bibinfo {author} {\bibfnamefont {Romain}\ \bibnamefont {Couillet}},
  \ and\ \bibinfo {author} {\bibfnamefont {Michael~W}\ \bibnamefont
  {Mahoney}},\ }\bibfield  {title} {\enquote {\bibinfo {title} {{A random
  matrix analysis of random Fourier features: beyond the Gaussian kernel, a
  precise phase transition, and the corresponding double descent}},}\ }\href
  {https://proceedings.neurips.cc/paper/2020/file/a03fa30821986dff10fc66647c84c9c3-Paper.pdf}
  {\bibfield  {journal} {\bibinfo  {journal} {Advances in Neural Information
  Processing Systems (NeurIPS)}\ }\textbf {\bibinfo {volume} {33}} (\bibinfo
  {year} {2020})}\BibitemShut {NoStop}%
\bibitem [{\citenamefont {Lin}\ and\ \citenamefont {Dobriban}(2021)}]{Lin2020}%
  \BibitemOpen
  \bibfield  {author} {\bibinfo {author} {\bibfnamefont {Licong}\ \bibnamefont
  {Lin}}\ and\ \bibinfo {author} {\bibfnamefont {Edgar}\ \bibnamefont
  {Dobriban}},\ }\bibfield  {title} {\enquote {\bibinfo {title} {{What causes
  the test error? Going beyond bias-variance via ANOVA}},}\ }\href
  {http://jmlr.org/papers/v22/20-1211.html} {\bibfield  {journal} {\bibinfo
  {journal} {Journal of Machine Learning Research}\ }\textbf {\bibinfo {volume}
  {22}},\ \bibinfo {pages} {1--82} (\bibinfo {year} {2021})}\BibitemShut
  {NoStop}%
\bibitem [{\citenamefont {Mitra}(2019)}]{Mitra2019}%
  \BibitemOpen
  \bibfield  {author} {\bibinfo {author} {\bibfnamefont {Partha~P}\
  \bibnamefont {Mitra}},\ }\bibfield  {title} {\enquote {\bibinfo {title}
  {{Understanding overfitting peaks in generalization error: Analytical risk
  curves for $l_2$ and $l_1$ penalized interpolation}},}\ }\href
  {http://arxiv.org/abs/1906.03667} {\  (\bibinfo {year} {2019})},\ \Eprint
  {http://arxiv.org/abs/1906.03667} {arXiv:1906.03667} \BibitemShut {NoStop}%
\bibitem [{\citenamefont {Mei}\ and\ \citenamefont
  {Montanari}(2021)}]{Mei2019}%
  \BibitemOpen
  \bibfield  {author} {\bibinfo {author} {\bibfnamefont {Song}\ \bibnamefont
  {Mei}}\ and\ \bibinfo {author} {\bibfnamefont {Andrea}\ \bibnamefont
  {Montanari}},\ }\bibfield  {title} {\enquote {\bibinfo {title} {{The
  Generalization Error of Random Features Regression: Precise Asymptotics and
  the Double Descent Curve}},}\ }\href {\doibase 10.1002/cpa.22008} {\bibfield
  {journal} {\bibinfo  {journal} {Communications on Pure and Applied
  Mathematics}\ } (\bibinfo {year} {2021}),\ 10.1002/cpa.22008}\BibitemShut
  {NoStop}%
\bibitem [{\citenamefont {Muthukumar}\ \emph {et~al.}(2019)\citenamefont
  {Muthukumar}, \citenamefont {Vodrahalli},\ and\ \citenamefont
  {Sahai}}]{Muthukumar2019}%
  \BibitemOpen
  \bibfield  {author} {\bibinfo {author} {\bibfnamefont {Vidya}\ \bibnamefont
  {Muthukumar}}, \bibinfo {author} {\bibfnamefont {Kailas}\ \bibnamefont
  {Vodrahalli}}, \ and\ \bibinfo {author} {\bibfnamefont {Anant}\ \bibnamefont
  {Sahai}},\ }\bibfield  {title} {\enquote {\bibinfo {title} {{Harmless
  interpolation of noisy data in regression}},}\ }\href {\doibase
  10.1109/ISIT.2019.8849614} {\bibfield  {journal} {\bibinfo  {journal} {IEEE
  International Symposium on Information Theory (ISIT)}\ ,\ \bibinfo {pages}
  {2299--2303}} (\bibinfo {year} {2019})}\BibitemShut {NoStop}%
\bibitem [{\citenamefont {Nakkiran}(2019)}]{Nakkiran2019}%
  \BibitemOpen
  \bibfield  {author} {\bibinfo {author} {\bibfnamefont {Preetum}\ \bibnamefont
  {Nakkiran}},\ }\bibfield  {title} {\enquote {\bibinfo {title} {{More Data Can
  Hurt for Linear Regression: Sample-wise Double Descent}},}\ }\href
  {http://arxiv.org/abs/1912.07242} {\  (\bibinfo {year} {2019})},\ \Eprint
  {http://arxiv.org/abs/1912.07242} {arXiv:1912.07242} \BibitemShut {NoStop}%
\bibitem [{\citenamefont {Xu}\ and\ \citenamefont {Hsu}(2019)}]{Xu2019}%
  \BibitemOpen
  \bibfield  {author} {\bibinfo {author} {\bibfnamefont {Ji}~\bibnamefont
  {Xu}}\ and\ \bibinfo {author} {\bibfnamefont {Daniel}\ \bibnamefont {Hsu}},\
  }\bibfield  {title} {\enquote {\bibinfo {title} {{On the number of variables
  to use in principal component regression}},}\ }\href
  {https://proceedings.neurips.cc/paper/2019/file/e465ae46b07058f4ab5e96b98f101756-Paper.pdf}
  {\bibfield  {journal} {\bibinfo  {journal} {Advances in Neural Information
  Processing Systems (NeurIPS)}\ }\textbf {\bibinfo {volume} {32}} (\bibinfo
  {year} {2019})}\BibitemShut {NoStop}%
\bibitem [{\citenamefont {Yang}\ \emph {et~al.}(2020)\citenamefont {Yang},
  \citenamefont {Yu}, \citenamefont {You}, \citenamefont {Steinhardt},\ and\
  \citenamefont {Ma}}]{Yang2020}%
  \BibitemOpen
  \bibfield  {author} {\bibinfo {author} {\bibfnamefont {Zitong}\ \bibnamefont
  {Yang}}, \bibinfo {author} {\bibfnamefont {Yaodong}\ \bibnamefont {Yu}},
  \bibinfo {author} {\bibfnamefont {Chong}\ \bibnamefont {You}}, \bibinfo
  {author} {\bibfnamefont {Jacob}\ \bibnamefont {Steinhardt}}, \ and\ \bibinfo
  {author} {\bibfnamefont {Yi}~\bibnamefont {Ma}},\ }\bibfield  {title}
  {\enquote {\bibinfo {title} {{Rethinking Bias-Variance Trade-off for
  Generalization of Neural Networks}},}\ }\href
  {http://proceedings.mlr.press/v119/yang20j.html} {\bibfield  {journal}
  {\bibinfo  {journal} {Proceedings of the 37th International Conference on
  Machine Learning (ICML)}\ }\bibinfo {series} {PMLR},\ \textbf {\bibinfo
  {volume} {119}},\ \bibinfo {pages} {10767--10777} (\bibinfo {year}
  {2020})}\BibitemShut {NoStop}%
\bibitem [{\citenamefont {Dauphin}\ \emph {et~al.}(2014)\citenamefont
  {Dauphin}, \citenamefont {Pascanu}, \citenamefont {Gulcehre}, \citenamefont
  {Cho}, \citenamefont {Ganguli},\ and\ \citenamefont {Bengio}}]{Dauphin2014}%
  \BibitemOpen
  \bibfield  {author} {\bibinfo {author} {\bibfnamefont {Yann~N.}\ \bibnamefont
  {Dauphin}}, \bibinfo {author} {\bibfnamefont {Razvan}\ \bibnamefont
  {Pascanu}}, \bibinfo {author} {\bibfnamefont {Caglar}\ \bibnamefont
  {Gulcehre}}, \bibinfo {author} {\bibfnamefont {Kyunghyun}\ \bibnamefont
  {Cho}}, \bibinfo {author} {\bibfnamefont {Surya}\ \bibnamefont {Ganguli}}, \
  and\ \bibinfo {author} {\bibfnamefont {Yoshua}\ \bibnamefont {Bengio}},\
  }\bibfield  {title} {\enquote {\bibinfo {title} {{Identifying and attacking
  the saddle point problem in high-dimensional non-convex optimization}},}\
  }\href
  {https://proceedings.neurips.cc/paper/2014/file/17e23e50bedc63b4095e3d8204ce063b-Paper.pdf}
  {\bibfield  {journal} {\bibinfo  {journal} {Advances in Neural Information
  Processing Systems (NeurIPS)}\ }\textbf {\bibinfo {volume} {27}} (\bibinfo
  {year} {2014})}\BibitemShut {NoStop}%
\bibitem [{\citenamefont {Engel}\ \emph {et~al.}(2001)\citenamefont {Engel},
  \citenamefont {den Broeck},\ and\ \citenamefont
  {Broeck}}]{engel2001statistical}%
  \BibitemOpen
  \bibfield  {author} {\bibinfo {author} {\bibfnamefont {A}~\bibnamefont
  {Engel}}, \bibinfo {author} {\bibfnamefont {C}~\bibnamefont {den Broeck}}, \
  and\ \bibinfo {author} {\bibfnamefont {C}~\bibnamefont {Broeck}},\ }\href
  {https://books.google.com/books?id=qVo4IT9ByfQC} {\emph {\bibinfo {title}
  {{Statistical Mechanics of Learning}}}},\ Statistical Mechanics of Learning\
  (\bibinfo  {publisher} {Cambridge University Press},\ \bibinfo {year}
  {2001})\BibitemShut {NoStop}%
\bibitem [{\citenamefont {M{\'{e}}zard}\ and\ \citenamefont
  {Parisi}(2003)}]{Mezard2003}%
  \BibitemOpen
  \bibfield  {author} {\bibinfo {author} {\bibfnamefont {Marc}\ \bibnamefont
  {M{\'{e}}zard}}\ and\ \bibinfo {author} {\bibfnamefont {Giorgio}\
  \bibnamefont {Parisi}},\ }\bibfield  {title} {\enquote {\bibinfo {title}
  {{The Cavity Method at Zero Temperature}},}\ }\href {\doibase
  10.1023/A:1022221005097} {\bibfield  {journal} {\bibinfo  {journal} {Journal
  of Statistical Physics}\ }\textbf {\bibinfo {volume} {111}},\ \bibinfo
  {pages} {1--34} (\bibinfo {year} {2003})}\BibitemShut {NoStop}%
\bibitem [{\citenamefont {Ramezanali}\ \emph {et~al.}(2015)\citenamefont
  {Ramezanali}, \citenamefont {Mitra},\ and\ \citenamefont
  {Sengupta}}]{Ramezanali2015}%
  \BibitemOpen
  \bibfield  {author} {\bibinfo {author} {\bibfnamefont {Mohammad}\
  \bibnamefont {Ramezanali}}, \bibinfo {author} {\bibfnamefont {Partha~P.}\
  \bibnamefont {Mitra}}, \ and\ \bibinfo {author} {\bibfnamefont {Anirvan~M.}\
  \bibnamefont {Sengupta}},\ }\bibfield  {title} {\enquote {\bibinfo {title}
  {{The cavity method for analysis of large-scale penalized regression}},}\
  }\href {http://arxiv.org/abs/1501.03194} {\  (\bibinfo {year} {2015})},\
  \Eprint {http://arxiv.org/abs/1501.03194} {arXiv:1501.03194} \BibitemShut
  {NoStop}%
\bibitem [{\citenamefont {Mehta}\ \emph
  {et~al.}(2019{\natexlab{b}})\citenamefont {Mehta}, \citenamefont {Cui},
  \citenamefont {Wang},\ and\ \citenamefont {Marsland}}]{Mehta2019a}%
  \BibitemOpen
  \bibfield  {author} {\bibinfo {author} {\bibfnamefont {Pankaj}\ \bibnamefont
  {Mehta}}, \bibinfo {author} {\bibfnamefont {Wenping}\ \bibnamefont {Cui}},
  \bibinfo {author} {\bibfnamefont {Ching-Hao}\ \bibnamefont {Wang}}, \ and\
  \bibinfo {author} {\bibfnamefont {Robert}\ \bibnamefont {Marsland}},\
  }\bibfield  {title} {\enquote {\bibinfo {title} {{Constrained optimization as
  ecological dynamics with applications to random quadratic programming in high
  dimensions}},}\ }\href {\doibase 10.1103/PhysRevE.99.052111} {\bibfield
  {journal} {\bibinfo  {journal} {Physical Review E}\ }\textbf {\bibinfo
  {volume} {99}},\ \bibinfo {pages} {052111} (\bibinfo {year}
  {2019}{\natexlab{b}})}\BibitemShut {NoStop}%
\bibitem [{\citenamefont {Advani}\ \emph {et~al.}(2018)\citenamefont {Advani},
  \citenamefont {Bunin},\ and\ \citenamefont {Mehta}}]{Advani2018}%
  \BibitemOpen
  \bibfield  {author} {\bibinfo {author} {\bibfnamefont {Madhu}\ \bibnamefont
  {Advani}}, \bibinfo {author} {\bibfnamefont {Guy}\ \bibnamefont {Bunin}}, \
  and\ \bibinfo {author} {\bibfnamefont {Pankaj}\ \bibnamefont {Mehta}},\
  }\bibfield  {title} {\enquote {\bibinfo {title} {{Statistical physics of
  community ecology: a cavity solution to MacArthur's consumer resource
  model}},}\ }\href {\doibase 10.1088/1742-5468/aab04e} {\bibfield  {journal}
  {\bibinfo  {journal} {Journal of Statistical Mechanics: Theory and
  Experiment}\ }\textbf {\bibinfo {volume} {2018}},\ \bibinfo {pages} {033406}
  (\bibinfo {year} {2018})}\BibitemShut {NoStop}%
\bibitem [{\citenamefont {Mar{\v{c}}enko}\ and\ \citenamefont
  {Pastur}(1967)}]{Marchenko1967}%
  \BibitemOpen
  \bibfield  {author} {\bibinfo {author} {\bibfnamefont
  {Vladimir~Alexandrovich}\ \bibnamefont {Mar{\v{c}}enko}}\ and\ \bibinfo
  {author} {\bibfnamefont {Leonid~Andreevich}\ \bibnamefont {Pastur}},\
  }\bibfield  {title} {\enquote {\bibinfo {title} {{Distribution of eigenvalues
  for some sets of random matrices}},}\ }\href {\doibase
  10.1070/SM1967v001n04ABEH001994} {\bibfield  {journal} {\bibinfo  {journal}
  {Mathematics of the USSR-Sbornik}\ }\textbf {\bibinfo {volume} {1}},\
  \bibinfo {pages} {457--483} (\bibinfo {year} {1967})}\BibitemShut {NoStop}%
\bibitem [{\citenamefont {Pennington}\ and\ \citenamefont
  {Worah}(2019)}]{Pennington2019}%
  \BibitemOpen
  \bibfield  {author} {\bibinfo {author} {\bibfnamefont {Jeffrey}\ \bibnamefont
  {Pennington}}\ and\ \bibinfo {author} {\bibfnamefont {Pratik}\ \bibnamefont
  {Worah}},\ }\bibfield  {title} {\enquote {\bibinfo {title} {{Nonlinear random
  matrix theory for deep learning}},}\ }\href {\doibase
  10.1088/1742-5468/ab3bc3} {\bibfield  {journal} {\bibinfo  {journal} {Journal
  of Statistical Mechanics: Theory and Experiment}\ }\textbf {\bibinfo {volume}
  {2019}},\ \bibinfo {pages} {124005} (\bibinfo {year} {2019})}\BibitemShut
  {NoStop}%
\bibitem [{\citenamefont {M{\'{e}}zard}(2017)}]{Mezard2017}%
  \BibitemOpen
  \bibfield  {author} {\bibinfo {author} {\bibfnamefont {Marc}\ \bibnamefont
  {M{\'{e}}zard}},\ }\bibfield  {title} {\enquote {\bibinfo {title}
  {{Mean-field message-passing equations in the Hopfield model and its
  generalizations}},}\ }\href {\doibase 10.1103/PhysRevE.95.022117} {\bibfield
  {journal} {\bibinfo  {journal} {Physical Review E}\ }\textbf {\bibinfo
  {volume} {95}},\ \bibinfo {pages} {022117} (\bibinfo {year}
  {2017})}\BibitemShut {NoStop}%
\bibitem [{\citenamefont {Ramezanali}\ \emph {et~al.}(2019)\citenamefont
  {Ramezanali}, \citenamefont {Mitra},\ and\ \citenamefont
  {Sengupta}}]{Ramezanali2019}%
  \BibitemOpen
  \bibfield  {author} {\bibinfo {author} {\bibfnamefont {Mohammad}\
  \bibnamefont {Ramezanali}}, \bibinfo {author} {\bibfnamefont {Partha~P.}\
  \bibnamefont {Mitra}}, \ and\ \bibinfo {author} {\bibfnamefont {Anirvan~M.}\
  \bibnamefont {Sengupta}},\ }\bibfield  {title} {\enquote {\bibinfo {title}
  {{Critical Behavior and Universality Classes for an Algorithmic Phase
  Transition in Sparse Reconstruction}},}\ }\href {\doibase
  10.1007/s10955-019-02292-6} {\bibfield  {journal} {\bibinfo  {journal}
  {Journal of Statistical Physics}\ }\textbf {\bibinfo {volume} {175}},\
  \bibinfo {pages} {764--788} (\bibinfo {year} {2019})}\BibitemShut {NoStop}%
\bibitem [{\citenamefont {Cui}\ \emph {et~al.}(2020)\citenamefont {Cui},
  \citenamefont {Rocks},\ and\ \citenamefont {Mehta}}]{Cui2020}%
  \BibitemOpen
  \bibfield  {author} {\bibinfo {author} {\bibfnamefont {Wenping}\ \bibnamefont
  {Cui}}, \bibinfo {author} {\bibfnamefont {Jason~W.}\ \bibnamefont {Rocks}}, \
  and\ \bibinfo {author} {\bibfnamefont {Pankaj}\ \bibnamefont {Mehta}},\
  }\bibfield  {title} {\enquote {\bibinfo {title} {{The perturbative resolvent
  method: Spectral densities of random matrix ensembles via perturbation
  theory}},}\ }\href@noop {} {\  (\bibinfo {year} {2020})},\ \Eprint
  {http://arxiv.org/abs/2012.00663} {arXiv:2012.00663} \BibitemShut {NoStop}%
\bibitem [{\citenamefont {Bishop}(2006)}]{Bishop2006}%
  \BibitemOpen
  \bibfield  {author} {\bibinfo {author} {\bibfnamefont {Christopher~M.}\
  \bibnamefont {Bishop}},\ }\href@noop {} {\emph {\bibinfo {title} {{Pattern
  Recognition and Machine Learning}}}}\ (\bibinfo  {publisher} {Springer},\
  \bibinfo {year} {2006})\BibitemShut {NoStop}%
\bibitem [{\citenamefont {Goldt}\ \emph {et~al.}(2020)\citenamefont {Goldt},
  \citenamefont {M{\'{e}}zard}, \citenamefont {Krzakala},\ and\ \citenamefont
  {Zdeborov{\'{a}}}}]{Goldt2019}%
  \BibitemOpen
  \bibfield  {author} {\bibinfo {author} {\bibfnamefont {Sebastian}\
  \bibnamefont {Goldt}}, \bibinfo {author} {\bibfnamefont {Marc}\ \bibnamefont
  {M{\'{e}}zard}}, \bibinfo {author} {\bibfnamefont {Florent}\ \bibnamefont
  {Krzakala}}, \ and\ \bibinfo {author} {\bibfnamefont {Lenka}\ \bibnamefont
  {Zdeborov{\'{a}}}},\ }\bibfield  {title} {\enquote {\bibinfo {title}
  {{Modeling the Influence of Data Structure on Learning in Neural Networks:
  The Hidden Manifold Model}},}\ }\href {\doibase 10.1103/PhysRevX.10.041044}
  {\bibfield  {journal} {\bibinfo  {journal} {Physical Review X}\ }\textbf
  {\bibinfo {volume} {10}},\ \bibinfo {pages} {041044} (\bibinfo {year}
  {2020})}\BibitemShut {NoStop}%
\bibitem [{\citenamefont {Rocks}\ and\ \citenamefont
  {Mehta}(2021)}]{Rocks2021}%
  \BibitemOpen
  \bibfield  {author} {\bibinfo {author} {\bibfnamefont {Jason~W.}\
  \bibnamefont {Rocks}}\ and\ \bibinfo {author} {\bibfnamefont {Pankaj}\
  \bibnamefont {Mehta}},\ }\bibfield  {title} {\enquote {\bibinfo {title} {{The
  Geometry of Over-parameterized Regression and Adversarial Perturbations}},}\
  }\href@noop {} {\  (\bibinfo {year} {2021})},\ \Eprint
  {http://arxiv.org/abs/2103.14108} {arXiv:2103.14108} \BibitemShut {NoStop}%
\bibitem [{\citenamefont {Chaudhari}\ \emph {et~al.}(2019)\citenamefont
  {Chaudhari}, \citenamefont {Choromanska}, \citenamefont {Soatto},
  \citenamefont {LeCun}, \citenamefont {Baldassi}, \citenamefont {Borgs},
  \citenamefont {Chayes}, \citenamefont {Sagun},\ and\ \citenamefont
  {Zecchina}}]{Chaudhari2019}%
  \BibitemOpen
  \bibfield  {author} {\bibinfo {author} {\bibfnamefont {Pratik}\ \bibnamefont
  {Chaudhari}}, \bibinfo {author} {\bibfnamefont {Anna}\ \bibnamefont
  {Choromanska}}, \bibinfo {author} {\bibfnamefont {Stefano}\ \bibnamefont
  {Soatto}}, \bibinfo {author} {\bibfnamefont {Yann}\ \bibnamefont {LeCun}},
  \bibinfo {author} {\bibfnamefont {Carlo}\ \bibnamefont {Baldassi}}, \bibinfo
  {author} {\bibfnamefont {Christian}\ \bibnamefont {Borgs}}, \bibinfo {author}
  {\bibfnamefont {Jennifer}\ \bibnamefont {Chayes}}, \bibinfo {author}
  {\bibfnamefont {Levent}\ \bibnamefont {Sagun}}, \ and\ \bibinfo {author}
  {\bibfnamefont {Riccardo}\ \bibnamefont {Zecchina}},\ }\bibfield  {title}
  {\enquote {\bibinfo {title} {{Entropy-SGD: biasing gradient descent into wide
  valleys}},}\ }\href {\doibase 10.1088/1742-5468/ab39d9} {\bibfield  {journal}
  {\bibinfo  {journal} {Journal of Statistical Mechanics: Theory and
  Experiment}\ }\textbf {\bibinfo {volume} {2019}},\ \bibinfo {pages} {124018}
  (\bibinfo {year} {2019})}\BibitemShut {NoStop}%
\bibitem [{\citenamefont {Baldassi}\ \emph {et~al.}(2020)\citenamefont
  {Baldassi}, \citenamefont {Malatesta}, \citenamefont {Negri},\ and\
  \citenamefont {Zecchina}}]{Baldassi2020}%
  \BibitemOpen
  \bibfield  {author} {\bibinfo {author} {\bibfnamefont {Carlo}\ \bibnamefont
  {Baldassi}}, \bibinfo {author} {\bibfnamefont {Enrico~M.}\ \bibnamefont
  {Malatesta}}, \bibinfo {author} {\bibfnamefont {Matteo}\ \bibnamefont
  {Negri}}, \ and\ \bibinfo {author} {\bibfnamefont {Riccardo}\ \bibnamefont
  {Zecchina}},\ }\bibfield  {title} {\enquote {\bibinfo {title} {{Wide flat
  minima and optimal generalization in classifying high-dimensional Gaussian
  mixtures}},}\ }\href {\doibase 10.1088/1742-5468/abcd31} {\bibfield
  {journal} {\bibinfo  {journal} {Journal of Statistical Mechanics: Theory and
  Experiment}\ }\textbf {\bibinfo {volume} {2020}},\ \bibinfo {pages} {124012}
  (\bibinfo {year} {2020})}\BibitemShut {NoStop}%
\bibitem [{\citenamefont {Pittorino}\ \emph {et~al.}(2021)\citenamefont
  {Pittorino}, \citenamefont {Lucibello}, \citenamefont {Feinauer},
  \citenamefont {Malatesta}, \citenamefont {Perugini}, \citenamefont
  {Baldassi}, \citenamefont {Negri}, \citenamefont {Demyanenko},\ and\
  \citenamefont {Zecchina}}]{Pittorino2020}%
  \BibitemOpen
  \bibfield  {author} {\bibinfo {author} {\bibfnamefont {Fabrizio}\
  \bibnamefont {Pittorino}}, \bibinfo {author} {\bibfnamefont {Carlo}\
  \bibnamefont {Lucibello}}, \bibinfo {author} {\bibfnamefont {Christoph}\
  \bibnamefont {Feinauer}}, \bibinfo {author} {\bibfnamefont {Enrico~M.}\
  \bibnamefont {Malatesta}}, \bibinfo {author} {\bibfnamefont {Gabriele}\
  \bibnamefont {Perugini}}, \bibinfo {author} {\bibfnamefont {Carlo}\
  \bibnamefont {Baldassi}}, \bibinfo {author} {\bibfnamefont {Matteo}\
  \bibnamefont {Negri}}, \bibinfo {author} {\bibfnamefont {Elizaveta}\
  \bibnamefont {Demyanenko}}, \ and\ \bibinfo {author} {\bibfnamefont
  {Riccardo}\ \bibnamefont {Zecchina}},\ }\bibfield  {title} {\enquote
  {\bibinfo {title} {{Entropic gradient descent algorithms and wide flat
  minima}},}\ }\href {https://openreview.net/forum?id=xjXg0bnoDmS} {\bibfield
  {journal} {\bibinfo  {journal} {International Conference on Learning
  Representations (ICLR)}\ } (\bibinfo {year} {2021})}\BibitemShut {NoStop}%
\bibitem [{\citenamefont {Yaida}(2020)}]{Yaida2020}%
  \BibitemOpen
  \bibfield  {author} {\bibinfo {author} {\bibfnamefont {Sho}\ \bibnamefont
  {Yaida}},\ }\bibfield  {title} {\enquote {\bibinfo {title} {{Non-Gaussian
  processes and neural networks at finite widths}},}\ }\href
  {http://proceedings.mlr.press/v107/yaida20a.html} {\bibfield  {journal}
  {\bibinfo  {journal} {Proceedings of The First Mathematical and Scientific
  Machine Learning Conference}\ }\bibinfo {series} {PMLR},\ \textbf {\bibinfo
  {volume} {107}},\ \bibinfo {pages} {165--192} (\bibinfo {year}
  {2020})}\BibitemShut {NoStop}%
\bibitem [{\citenamefont {Franz}\ \emph {et~al.}(2017)\citenamefont {Franz},
  \citenamefont {Parisi}, \citenamefont {Sevelev}, \citenamefont {Urbani},\
  and\ \citenamefont {Zamponi}}]{Franz2017}%
  \BibitemOpen
  \bibfield  {author} {\bibinfo {author} {\bibfnamefont {Silvio}\ \bibnamefont
  {Franz}}, \bibinfo {author} {\bibfnamefont {Giorgio}\ \bibnamefont {Parisi}},
  \bibinfo {author} {\bibfnamefont {Maksim}\ \bibnamefont {Sevelev}}, \bibinfo
  {author} {\bibfnamefont {Pierfrancesco}\ \bibnamefont {Urbani}}, \ and\
  \bibinfo {author} {\bibfnamefont {Francesco}\ \bibnamefont {Zamponi}},\
  }\bibfield  {title} {\enquote {\bibinfo {title} {{Universality of the
  SAT-UNSAT (jamming) threshold in non-convex continuous constraint
  satisfaction problems}},}\ }\href {\doibase 10.21468/SciPostPhys.2.3.019}
  {\bibfield  {journal} {\bibinfo  {journal} {SciPost Physics}\ }\textbf
  {\bibinfo {volume} {2}},\ \bibinfo {pages} {019} (\bibinfo {year}
  {2017})}\BibitemShut {NoStop}%
\bibitem [{\citenamefont {Sagun}\ \emph {et~al.}(2018)\citenamefont {Sagun},
  \citenamefont {Evci}, \citenamefont {G{\"{u}}ney}, \citenamefont {Dauphin},\
  and\ \citenamefont {Bottou}}]{Sagun2018}%
  \BibitemOpen
  \bibfield  {author} {\bibinfo {author} {\bibfnamefont {Levent}\ \bibnamefont
  {Sagun}}, \bibinfo {author} {\bibfnamefont {Utku}\ \bibnamefont {Evci}},
  \bibinfo {author} {\bibfnamefont {V.~U˘gur}\ \bibnamefont {G{\"{u}}ney}},
  \bibinfo {author} {\bibfnamefont {Yann}\ \bibnamefont {Dauphin}}, \ and\
  \bibinfo {author} {\bibfnamefont {L{\'{e}}on}\ \bibnamefont {Bottou}},\
  }\bibfield  {title} {\enquote {\bibinfo {title} {{Empirical analysis of the
  hessian of over-parametrized neural networks}},}\ }\href@noop {} {\
  (\bibinfo {year} {2018})},\ \Eprint {http://arxiv.org/abs/1706.04454}
  {arXiv:1706.04454} \BibitemShut {NoStop}%
\bibitem [{\citenamefont {Finn}\ \emph {et~al.}(2017)\citenamefont {Finn},
  \citenamefont {Abbeel},\ and\ \citenamefont {Levine}}]{Finn2017}%
  \BibitemOpen
  \bibfield  {author} {\bibinfo {author} {\bibfnamefont {Chelsea}\ \bibnamefont
  {Finn}}, \bibinfo {author} {\bibfnamefont {Pieter}\ \bibnamefont {Abbeel}}, \
  and\ \bibinfo {author} {\bibfnamefont {Sergey}\ \bibnamefont {Levine}},\
  }\bibfield  {title} {\enquote {\bibinfo {title} {{Model-agnostic
  meta-learning for fast adaptation of deep networks}},}\ }\href
  {https://proceedings.mlr.press/v70/finn17a.html} {\bibfield  {journal}
  {\bibinfo  {journal} {Proceedings of the 34th International Conference on
  Machine Learning (ICML)}\ }\bibinfo {series} {PMLR},\ \textbf {\bibinfo
  {volume} {70}},\ \bibinfo {pages} {1126----1135} (\bibinfo {year}
  {2017})}\BibitemShut {NoStop}%
\end{thebibliography}


%

\onecolumngrid

\appendix

\section*{Appendix: Analytic Expressions for Random Nonlinear Features Model}\label{sec:exprnfm}

For the random nonlinear features model, the solutions for the five averages used to express the solutions in Eqs.~\eqref{eq:gen_train}-\eqref{eq:gen_var} are in turn related to a set of five scalar susceptibilities that are a natural result of the cavity
method, $\nu$, $\chi$, $\kappa$, $\omega$, and $\phi$. 
Each of these susceptibilities is defined as the ensemble average of the trace of a different susceptibility matrix which measures the responses of quantities such as the residual label error, residual parameter error, fit parameter values, etc., to small perturbations (see Sec.~\ref{sec:suscept}).
Collectively, the ensemble-averaged quantities satisfy the equations
\begin{align}
\mqty(\expval*{\hat{w}^2}\\  \expval*{\hat{u}^2} \\ \expval*{\Delta y^2} \\ \expval*{\Delta \beta^2}) &=
\mqty(1 & -\sigma_W^2  \frac{\alpha_f}{\alpha_p}\nu^2 &  -\sigma_{\delta z}^2 \alpha_p^{-1}\nu^2 & 0\\
- \sigma_W^2 \omega^2   & 1  & -  \sigma_X^2\alpha_f^{-1}\kappa^2 & 0\\
-  \sigma_{\delta z}^2 \chi^2  & 0 & 1 &  -\sigma_X^2\chi^2 \\
-  \sigma_W^2\kappa^2 & 0 &  -   \sigma_X^2 \alpha_f^{-1}\phi^2 & 1)^{-1}
\mqty(0 \\    \sigma_\beta^2 \omega^2\\ (\sigma_\varepsilon^2 + \sigma_{\delta y^*}^2)\chi^2 \\  \sigma_\beta^2\kappa^2)\label{eq:rnlfm_means}
\end{align}
\vspace{-\baselineskip}
\begin{align}
\expval*{\hat{w}_1\hat{w}_2} &= \frac{\sigma_\beta^2}{\sigma_W^2} \frac{ \sigma_W^4\frac{\alpha_f}{\alpha_p} \omega^2\nu^2}{\qty(1- \sigma_W^4\frac{\alpha_f}{\alpha_p} \omega^2\nu^2)}, &
\expval*{\Delta \beta_1\Delta \beta_2}&= \sigma_\beta^2 \frac{ \kappa^2}{\qty(1- \sigma_W^4\frac{\alpha_f}{\alpha_p} \omega^2\nu^2)}\label{eq:rnlfm_ovleraps}.
\end{align}
The quantity $\expval*{\hat{u}^2}$ is the mean squared average of the length-$N_f$ vector quantity $\hbu= X^T\Delta \vby$ obtained as a byproduct of the cavity derivation.
The solutions for linear regression can be formulated similarly in terms of a pair of scalar susceptibilities that are analogous to $\chi$ and $\nu$ (see Sec.~S1F of Supplemental Material~\cite{supporting_info}).

In each model, a subset of the scalar susceptibilities diverges wherever two different sets of solutions meet, indicating the existence of a second-order phase transition.
For the random nonlinear features model, these susceptibilities are (to leading order in small $\lambda$)
\begin{alignat}{2}
\chi &= \left\{
\begin{array}{c}
1-\alpha_p\\
 \frac{\lambda}{2 \sigma_{\delta z}^2 }\frac{\alpha_p}{ (\alpha_p-1)} \qty[ 1- \qty(1+\Delta \varphi)\alpha_f + \sqrt{\qty[1-\qty(1+ \Delta \varphi)\alpha_f]^2   + 4  \Delta\varphi  \alpha_f }]
\end{array}
\right. &\quad & \begin{array}{l}
\qif N_p < M\\
\qif N_p > M
\end{array}\label{eq:rnlfm_chi}\\
\nu &= \left\{
\begin{array}{c}
\frac{1}{2 \sigma_{\delta z}^2}\frac{1}{(1-\alpha_p)}\qty[\alpha_p -  \qty(1+ \Delta \varphi) \alpha_f + \sqrt{  \qty[\alpha_p -   \qty(1+ \Delta \varphi)\alpha_f]^2   + 4  \Delta \varphi \alpha_f\alpha_p }] \\
\frac{1}{\lambda} \frac{(\alpha_p-1)}{\alpha_p} + + \frac{1}{2\sigma_{\delta z}^2  }\frac{1}{ (\alpha_p-1)} \qty[ 1- \qty(1+\Delta \varphi)\alpha_f + \sqrt{\qty[1-\qty(1+ \Delta \varphi)\alpha_f]^2   + 4  \Delta\varphi  \alpha_f }] 
\end{array}
\right.  &\quad & \begin{array}{l}
\qif N_p < M\\
\qif N_p > M
\end{array}\label{eq:rnlfm_nu}
\end{alignat}
\vspace{-\baselineskip}
\begin{align}
\kappa &= \frac{1}{1 +   \sigma_X^2 \sigma_W^2 \alpha_f^{-1} \chi\nu}, &
\omega &=  \sigma_X^2\alpha_f^{-1}\chi\kappa, &
\phi  &=   -\sigma_W^2 \nu \kappa.\label{eq:rnlfm_kappa}
\end{align}
We plot these analytic forms for $\chi$, $\nu$ and $\kappa$ in Fig.~\ref{fig:sus}.

\end{document}


\title{Supplemental Material:\\ Memorizing without overfitting:\\  Bias, variance, and interpolation in over-parameterized models}
\author{Jason W. Rocks}
\affiliation{Department of Physics, Boston University, Boston, Massachusetts 02215, USA}

\author{Pankaj Mehta}
\affiliation{Department of Physics, Boston University, Boston, Massachusetts 02215, USA}
\affiliation{Faculty of Computing and Data Sciences, Boston University, Boston, Massachusetts 02215, USA}
\maketitle

\tableofcontents

\section{Cavity Derivations}

In this section, we provide detailed derivations of all closed-form solutions for both models. 
We begin by setting up the theoretical framework and providing some useful approximations before deriving solutions for our two models.
These calculations follow the general procedure laid out in Ref.~\onlinecite{Mehta2019a}.

\subsection{Notation Conventions} 

\begin{itemize}
\item We define $M$ as the number of points in the training data set, $N_f$ as the number of input features, and $N_p$ as the number of fit parameters/hidden features. We define the ratios $\alpha_f = N_f/M$ and $\alpha_p = N_p/M$.
\item Unless otherwise specified, the type of symbol used for an index label (e.g., $\Delta y_a$) or as a summation index (e.g., $\sum_a$) implies its range. The symbols $a$, $b$, or $c$ imply ranges over the training data points from $1$ to $M$, 
the symbols $j$, $k$, or $l$ imply ranges over the input features from $1$ to $N_f$, and the symbols $J$, $K$, or $L$ imply ranges over the fit parameters/hidden features from $1$ to $N_p$.
\item The notation $\E[x]\qty[\cdot]$, $\Var[x]\qty[\cdot]$ and $\Cov[x]\qty[\cdot, \cdot]$ represent the mean, variance, and covariance, respectively, with respect to one or more random variables $x$. 
A lack of subscript implies averages taken with respect to the total ensemble distribution, i.e., taken over all possible sources of randomness.
A subscript $0$ implies averages taken with respect to random variables containing one or more $0$-valued indices (e.g., $X_{a0}$, $X_{0j}$, $W_{0J}$, or $W_{j0}$).
\item We use the notation $\order{\cdot}$ to represent standard ``Big-O'' notation, indicating an upper bound on the limiting scaling behavior of a quantity with respect to the argument.
\end{itemize}

\subsection{Theoretical Setup}

For completeness, we begin by reproducing some of our theoretical setup from the main text.
We consider data points $(y, \vbx)$, each consisting of a label $y$ and a vector $\vbx$ of $N_f$ input features.
The labels are related to the input features via the teacher model
\begin{equation}
y(\vbx) = y^*(\vbx; \vbbeta) + \varepsilon,
\end{equation}
where $\varepsilon$ is the label noise and $y^*(\vbx; \vbbeta)$ are the true labels which depend on a vector $\vbbeta$ of ``ground truth'' parameters.
We consider features and label noise that are independently and identically distributed, drawn from a normal distributions with zero mean and variances $\sigma_X^2/N_f$ and $\sigma_\varepsilon^2$, respectively, so that 
\begin{alignat}{2}
\E\qty[x_{a, j}] &=0 \qqc &\Cov\qty[x_{a, j}, x_{b, k}] &= \frac{\sigma_X^2}{N_f}\delta_{ab}\delta_{jk}\\
\E\qty[\varepsilon_a] &=0\qqc  &\Cov\qty[\varepsilon_a, \varepsilon_b] &= \sigma_\varepsilon^2\delta_{ab}
\end{alignat}
for two data points $\vbx_a$ and $\vbx_b$ with label noise $\varepsilon_a$ and $\varepsilon_b$.

We also restrict ourselves to a tacher model of the form
\begin{equation}
y^*(\vbx;\vbbeta) =  \frac{\sigma_\beta\sigma_X}{\expval*{f'}} f\qty(\frac{\vbx\cdot\vbbeta}{\sigma_X\sigma_\beta} )
\end{equation}
where the function $f$ is an arbitrary nonlinear function and ${\expval*{f'} = \frac{1}{\sqrt{2\pi}}\int_{-\infty}^\infty \dd he^{-\frac{h^2}{2}}f'(h)}$ with prime notation used to indicate a derivative. 
We assume the ground truth parameters are independent of all other random parameters and are also normally distributed with zero mean and variance $\sigma_\beta^2$,
\begin{equation}
\E\qty[\beta_k] =0\qqc  \Cov\qty[\beta_j, \beta_k] = \sigma_\beta^2\delta_{jk}.
\end{equation}

We consider a training set of $M$ data points, ${\mathcal{D}=\{(y_b, \vbx_b)\}_{b=1}^M}$.
We organize each input feature vector into the rows of an observation matrix $X$ of size $M\times N_f$.

We consider a linear student model,
\begin{equation}
\hat{y}(\vbx) = \vbz(\vbx)\cdot \hbw,
\end{equation}
where $\hbw$ is a vector of $N_p$ fit parameters.
The values of the fit parameters are determined by minimizing the loss function
\begin{equation}
L (\hbw) = \frac{1}{2}\sum_b\Delta y_b^2 + \frac{\lambda}{2}\sum_K\hat{w}_K^2,\label{eq:SIloss}
\end{equation}
where we have defined the residual label error as $\Delta y_a = y_a - \hat{y}_a$.
Taking the gradient of the loss with respect to the fit parameters and setting it to zero results in a system of $N_p$ equations for the $N_p$ fit parameter,
\begin{equation}
0 = \pdv{L (\hbw)}{\hat{w}_J} = -\sum_b\Delta y_b Z_{bJ} + \lambda \hat{w}_J\label{eq:SIgrad}.
\end{equation}
Note that the regularization term ensures that this system of equations always has a unique solution.

\subsection{Central Limit Approximation}

Frequently in these derivations, we encounter large sums of statistically independent random variables.
In the thermodynamic limit, we utilize the central limit theorem to approximate these sums as a single random variable defined by only a mean and a variance. 
Here, we derive expressions for the approximate forms of three different types of sums that will be needed.
In the following derivations $N$ and $N'$ are considered to be thermodynamically large variables.

First, we define a length-$N$ vector $\vba$ of random variables $a_j$ that are normally distributed with zero mean and variance $\sigma_a^2/N$,
\begin{equation}
\E\qty[a_j] =0 \qqc \Cov\qty[a_j,a_k] = \frac{\sigma_a^2}{N}\delta_{jk}.
\end{equation} 
The first sum we approximate is the dot product $\vbc\cdot\vba$ where $\vbc$ is a length-$N$ vector with elements $c_j$ that are independent of $\vba$. In the thermodynamic limit, this sum approximates to
\begin{equation}
\sum_k c_k a_k \approx  \sigma z\qqc \sigma^2 = \frac{\sigma_a^2}{N} \sum_kc_k^2,\label{eq:clt_vec}
\end{equation}
where $z$ is a normally distributed variable with zero mean and unit variance.
To derive this we simply evaluate the mean and variance of this sum with respect to $\vba$,
\begin{align}
\E[\vba]\qty[\sum_k c_k a_k] &= \sum_k x_k \E[\vba]\qty[a_k ] = 0\\
\Var[\vba]\qty[\sum_k c_k a_k] &= \sum_k c_k^2\Var[\vba]\qty[a_k ] = \frac{\sigma_a^2}{N} \sum_kc_k^2.
\end{align}

The second sum we consider is the product $\vba^TA\vba$,
where $A$ is an $N\times N$ matrix  whose elements are independent of $\vba$,
\begin{equation}
\sum_{jk}A_{jk}a_ja_k \approx  \frac{\sigma_a^2}{N}\sum_k A_{kk}.\label{eq:clt_square}
\end{equation}
To derive this, we evaluate the mean of this sum with respect to $\vba$ to be
\begin{equation}
\E[\vba]\qty[\sum_{jk}A_{jk}a_ja_k] =\sum_{jk}A_{jk}\E[\vba]\qty[a_ja_k] = \frac{\sigma_a^2}{N}\sum_k A_{kk}.
\end{equation}
To calculate the variance, we use Wick's theorem to derive the fourth moment of the elements of $\vba$,
\begin{equation}
\E\qty[a_ja_ka_la_m] =  \sigma_a^4\qty(\delta_{jk}\delta_{lm} + \delta_{jl}\delta_{km} +\delta_{jm}\delta_{kl} ).
\end{equation}
Applying this identity, we find the variance to be
\begin{equation}
\begin{aligned}
\Var[\vba]\qty[\sum_{jk}A_{jk}a_ja_k] &= \sum_{jklm}A_{jk}A_{lm}\Cov[\vba]\qty[a_ja_k, a_la_m]\\
&= \sum_{jklm}A_{jk}A_{lm}\qty(\E[\vba]\qty[a_ja_k a_la_m] - \E[\vba]\qty[a_ja_k]\E[\vba]\qty[ a_la_m])\\
&= 2\frac{\sigma_a^4}{N^2}\sum_{jk} A_{jk}^2\\
&= 2\frac{\sigma_a^4}{N^2}\sum_i\sigma_i^2\\
& \approx  \order{\frac{1}{N}}.
\end{aligned}
\end{equation}
In the second-to-last line, we have rewritten the trace over $A^TA$ in terms of the eigenvalues $\sigma_i$ of $A$.
If each eigenvalue is $\order{1}$, then the total variance is $\order{1/N}$ and can be neglected.

For the third sum, we define an additional vector $\vbb$ of length $N'$ whose elements $b_J$ are independent of $\vba$ with zero mean and variance $\sigma_b^2/N'$,
\begin{equation}
\E\qty[b_J] =0 \qqc \Cov\qty[b_J,b_K] = \frac{\sigma_b^2}{N'}\delta_{JK}.
\end{equation}
The third sum we approximate is the product $\vba^TB\vbb$, where $B$ is a $N\times N'$ rectangular matrix whose elements are independent of both $\vba$ and $\vbb$,
\begin{align}
\sum_{jK}B_{jK}a_jb_K &\approx 0.\label{eq:clt_rect}
\end{align}
To derive this, we take the mean with respect to both $\vba$ and $\vbb$,
\begin{equation}
\E[\vba, \vbb]\qty[\sum_{jK}B_{jK}a_jb_K] =\sum_{jK}B_{jK}\E[\vba]\qty[a_j]\E[\vbb]\qty[b_K] = 0,
\end{equation}
and also evaluate the variance to be
\begin{equation}
\begin{aligned}
\Var[\vba, \vbb]\qty[\sum_{jK}B_{jK}a_jb_K] &= \sum_{jKlM}B_{jK}B_{lM}\Cov[\vba, \vbb]\qty[a_jb_K, a_lb_M]\\
&= \sum_{jKlM}B_{jK}B_{lM}\qty(\E[\vba]\qty[a_j a_l]\E[\vbb]\qty[b_Kb_M] - \E[\vba]\qty[a_j]\E[\vbb]\qty[b_K]\E[\vba]\qty[a_l]\E[\vbb]\qty[b_M])\\
&= \frac{\sigma_a^2\sigma_b^2}{NN'}\sum_{jK} B_{jK}^2\\
&= \frac{\sigma_a^2\sigma_b^2}{NN'}\sum_i\sigma_i^2\\
&\approx \order{\frac{1}{N}}.
\end{aligned}
\end{equation}
Analogous to the variance of the previous sum, in the second-to-last line we have decomposed $B$ in terms of its singular values $\sigma_i$. If each singular value of  $\order{1}$, then the total variance is $\order{1/N}$ and can be neglected.
Since the mean is also zero, we find that all sums of this form can be ignored in the thermodynamic limit.

\subsection{Nonlinear Function Statistics}

Here, we show how the labels and hidden features can each be decomposed into linear and nonlinear components that are statistically independent of one another.
We also derive the statistical properties of the resulting nonlinear components.

\subsubsection{Integral Identities}\label{sec:integrals}

First, we derive some useful integral identities for the expectation values of the nonlinear functions encountered in this work.
In this section, we consider input features that are correlated, but collectively follow a multivariate normal distribution with mean zero and covariance matrix $\Sigma_{\vbx}$,
\begin{equation}
\E\qty[\vbx] = 0 \qqc \Cov[][\vbx, \vbx^T] = \Sigma_{\vbx},
\end{equation}
where the covariance is normalized so that  $\Tr \Sigma_{\vbx} = \sigma_X^2$.
Throughout the rest this work, we usually consider independent input features where  $\Sigma_{\vbx} = \frac{\sigma_X^2}{N_f} I_{N_f}$.
We also define a pair of random vectors $\vba$ and $\vbb$, each of length $N_f$, whose elements are independent of $\vbx$ with mean and variances
\begin{equation}
\begin{gathered}
\E\qty[a_j] =0 \qqc \Cov\qty[a_j,a_k] = \sigma_a^2\delta_{jk}\\
\E\qty[b_j] =0 \qqc \Cov\qty[b_j,b_k] = \sigma_b^2\delta_{jk}\\
\Cov\qty[a_j, b_k] = 0.
\end{gathered}
\end{equation}
Defining $g(h)$ as an arbitrary function and taking the thermodynamic limit, we will utilize the following three approximate identities:
\begin{align}
\E[\vbx]\qty[g\qty(\frac{\vbx\cdot\vba}{\sigma_X\sigma_a})] & \approx \expval*{g}\label{eq:integral1}\\
\E[\vbx]\qty[\vbx g\qty(\frac{\vbx\cdot\vba}{\sigma_X\sigma_a})] &\approx\frac{\Sigma_{\vbx}\vba}{\sigma_X\sigma_a}\expval*{g'}\label{eq:integral2}\\
\E[\vbx]\qty[g\qty(\frac{\vbx\cdot\vba}{\sigma_X\sigma_a})g\qty(\frac{\vbx\cdot\vbb}{\sigma_X\sigma_b})]  & \approx \expval*{g}^2 \approx \E[\vbx]\qty[g\qty(\frac{\vbx\cdot\vba}{\sigma_X\sigma_a})]\E[\vbx]\qty[g\qty(\frac{\vbx\cdot\vbb}{\sigma_X\sigma_b})],\label{eq:integral3}
\end{align}
where we have defined the integrals
\begin{equation}
\expval*{g} =  \frac{1}{\sqrt{2\pi}} \int \dd h e^{-\frac{h^2}{2}} g(h)\qqc \expval*{g'} = \frac{1}{\sqrt{2\pi}} \int \dd h e^{-\frac{h^2}{2}} g'(h).
\end{equation}

Each of the above averages is evaluated with respect to the distribution of input features, where we define the differential over all elements of a vector of input features $\vbx$ as,
\begin{equation}
\mathcal{D}\vbx = \frac{\dd \vbx}{\sqrt{(2\pi )^{N_f}\det \Sigma_{\vbx}}} e^{-\frac{1}{2}\vbx^T\Sigma_{\vbx}^{-1}\vbx}.
\end{equation}

Next, we derive the identity in Eq.~\eqref{eq:integral1},
\begin{equation}
\begin{aligned}
\E[\vbx]\qty[g\qty(\frac{\vbx\cdot\vba}{\sigma_X\sigma_a})] &= \int \mathcal{D}\vbx g\qty(\frac{\vbx\cdot\vba}{\sigma_X\sigma_a})\\
&= \int \mathcal{D}\vbx \dd h g\qty(h)\delta\qty(h-\frac{\vbx\cdot\vba}{\sigma_X\sigma_a})\\
&=  \int \mathcal{D}\vbx \dd h \frac{\dd \tilde{h}}{2\pi}  g(h)e^{i\tilde{h}\qty(h- \frac{\vbx\cdot\vba}{\sigma_X\sigma_a})}\\
&=  \int \dd h \frac{\dd \tilde{h}}{2\pi} g(h)e^{i\tilde{h}h} \int\frac{\dd \vbx}{\sqrt{(2\pi )^{N_f}\det \Sigma_{\vbx}}} e^{-\frac{1}{2}\vbx^T\Sigma_{\vbx}^{-1}\vbx -i\tilde{h}\frac{\vbx\cdot\vba}{\sigma_X\sigma_a} }\\
&=  \int \dd h \frac{\dd \tilde{h}}{2\pi}g(h)e^{i\tilde{h}h-\frac{\tilde{h}^2}{2}\frac{\vba^T\Sigma_{\vbx}\vba}{\sigma_X^2\sigma_a^2}}.
\end{aligned}
\end{equation}
At this point, we approximate the sum in the exponential using the central limit theorem. With proper rescaling, we apply Eq.~\eqref{eq:clt_square} to find
\begin{equation}
\frac{\vba^T\Sigma_{\vbx}\vba}{\sigma_X^2\sigma_a^2} = \frac{1}{\sigma_X^2\sigma_a^2} \sum_{jk}\qty(N_f\Sigma_{\vbx, jk})\qty(\frac{a_j}{\sqrt{N_f}})\qty(\frac{a_k}{\sqrt{N_f}}) \approx 1,
\end{equation}
where we have identified $N_f\Sigma_{\vbx}$ and $a_j/\sqrt{N_f}$ with $A_{jk}$ and $a_j$, respectively, in Eq.~\eqref{eq:clt_vec}.
Using this approximation, we proceed to find
\begin{equation}
\begin{aligned}
\E[\vbx]\qty[g\qty(\frac{\vbx\cdot\vba}{\sigma_X\sigma_a})] &\approx  \int \dd h \frac{\dd \tilde{h}}{2\pi}g(h)e^{i\tilde{h}h-\frac{\tilde{h}^2}{2}}\\
&= \frac{1}{\sqrt{2\pi}} \int \dd h e^{-\frac{h^2}{2}} g(h)\\
& = \expval*{g}.
\end{aligned}
\end{equation}

To derive the remaining two identities in Eqs.~\eqref{eq:integral2} and \eqref{eq:integral3}, we follow analogous derivations.
In particular, we note that Eq.~\eqref{eq:integral3} implies that the two functions $g\qty(\frac{\vbx\cdot\vba}{\sigma_X\sigma_a})$ and $g\qty(\frac{\vbx\cdot\vbb}{\sigma_X\sigma_b})$ are statistically independent from one another in the thermodynamic limit.

\subsubsection{Label Decomposition}

By defining the ground truth parameters as shown below, we are able to decompose the labels into linear and nonlinear components,
\begin{equation}
y(\vbx) = \vbx\cdot\vbbeta + \delta y^*_{\mathrm{NL}}(\vbx) + \varepsilon  \qqc \vbbeta \equiv \Sigma_{\vbx}^{-1}\Cov[\vbx][\vbx, y^*(\vbx)],\label{eq:SIydecomp}
\end{equation}
where $ \delta y^*_{\mathrm{NL}}(\vbx)\equiv y^*(\vbx) -  \vbx\cdot\vbbeta$ and the covariance matrix of the input features $\Sigma_{\vbx} = \Cov[\vbx][\vbx, \vbx^T]$ is assumed to be invertible.

We prove that the linear and nonlinear terms are statistically independent  with respect the input features $\vbx$ as follows:
\begin{equation}
\begin{aligned}
\Cov[\vbx][\vbx\cdot\vbbeta, \delta y^*_{\mathrm{NL}}(\vbx)] &= \Cov[\vbx][\vbx\cdot\vbbeta, y(\vbx) - \vbx\cdot\vbbeta]\\
&= \vbbeta\cdot\Cov[\vbx][\vbx, y(\vbx) ] - \vbbeta\cdot\Cov[\vbx][\vbx, \vbx^T ]\vbbeta\\
&= \vbbeta\cdot\Sigma_{\vbx}\vbbeta - \vbbeta\cdot\Sigma_{\vbx}\vbbeta\\
&= 0.
\end{aligned}
\end{equation}

Furthermore, we can show that the ground truth parameters as defined in Eq.~\eqref{eq:SIydecomp} coincide with those of the teacher model,
\begin{equation}
y^*(\vbx) = \frac{\sigma_\beta\sigma_X}{\expval*{f'}}f\qty(\frac{\vbx\cdot\vbbeta}{\sigma_X\sigma_\beta}).
\end{equation}
To do this, we evaluate the covariance in Eq.~\eqref{eq:SIydecomp} and use the identity in Eq.~\eqref{eq:integral2} to find
\begin{equation}
\begin{aligned}
\vbbeta &= \Sigma_{\vbx}^{-1}\Cov[\vbx][\vbx, y^*(\vbx)]\\
&= \Sigma_{\vbx}^{-1} \frac{\sigma_\beta\sigma_X}{\expval*{f'}} \E[\vbx]\qty[\vbx f\qty(\frac{\vbx\cdot\vbbeta}{\sigma_X\sigma_\beta})]\\
&\approx \Sigma_{\vbx}^{-1} \frac{\sigma_\beta\sigma_X}{\expval*{f'}}\frac{\Sigma_{\vbx}\vbbeta}{\sigma_X\sigma_\beta}\expval*{f'}\\
&=\vbbeta.
\end{aligned}
\end{equation}
So we see that the definitions are consistent with one another.

Next, we calculate the variance of the nonlinear components of the labels $\delta y^*_{\mathrm{NL}}(\vbx)$. To do this, we first calculate the mean of the squared true labels using the identity in Eq.~\eqref{eq:integral1},
\begin{equation}
\begin{aligned}
\E[\vbx]\qty[(y^*(\vbx))^2] &= \frac{\sigma_\beta^2\sigma_X^2}{\expval*{f'}^2}\E[\vbx]\qty[ f\qty(\frac{\vbx\cdot\vbbeta}{\sigma_X\sigma_\beta})^2] \approx \sigma_\beta^2\sigma_X^2 \frac{\expval*{f^2}}{\expval*{f'}^2}.
\end{aligned}
\end{equation}
Using this result and the fact that the linear and nonlinear components of the labels are independent, we find the variance of the nonlinear components in the thermodynamic limit to be
\begin{equation}
\begin{aligned}
\Var[\vbx]\qty[\delta y^*_{\mathrm{NL}}(\vbx)] &= \sigma_\beta^2\sigma_X^2  \frac{\expval*{f^2} -\expval*{f'}^2 }{\expval*{f'}^2}.
\end{aligned}
\end{equation}
Furthermore, it is clear that the nonlinear components for two independent data points $\vbx_a$ and $\vbx_b$ will also be independent.

Since there are no other random variables present in the above variance, we summarize the statistical properties of the nonlinear components of the labels for independent data points $\vbx_a$ and $\vbx_b$ with full ensemble averages, giving us
\begin{equation}
\E\qty[\delta y^*_{\mathrm{NL}}(\vbx_a)] = 0 \qqc \Cov\qty[\delta y^*_{\mathrm{NL}}(\vbx_a), \delta y^*_{\mathrm{NL}}(\vbx_b)] = \sigma_{\delta y^*}^2\delta_{ab},
\end{equation}
where we have defined the variance $\sigma_{\delta y^*}^2$ of the nonlinear components as
\begin{equation}
\begin{gathered}
\sigma_{\delta y^*}^2 = \sigma_\beta^2\sigma_X^2 \Delta f\qqc \Delta f = \frac{\expval*{f^2} -\expval*{f'}^2 }{\expval*{f'}^2}\\
\expval*{f^2} = \frac{1}{\sqrt{2\pi}}\int\limits_{-\infty}^\infty dh e^{-\frac{h^2}{2}}f^2(h)\qqc \expval*{f'} = \frac{1}{\sqrt{2\pi}}\int\limits_{-\infty}^\infty dh e^{-\frac{h^2}{2}}f'(h).
\end{gathered}
\end{equation}

\subsubsection{Hidden Feature Decomposition}
Similar to the decomposition of the labels, we decompose the hidden features into linear and nonlinear components that are statistically independent with respect to the input features by defining $W$ as follows:
\begin{equation}
\vbz(\vbx) = \frac{\mu_Z}{\sqrt{N_p}}\vec{\mathbf{1}}  + W^T\vbx + \delta \vbz_{\mathrm{NL}}(\vbx) \qqc W \equiv \Sigma_{\vbx}^{-1}\Cov[\vbx][\vbx, \vbz^T(\vbx)],\label{eq:SIzdecomp}
\end{equation}
where we have defined the nonlinear component as ${\delta \vbz_{\mathrm{NL}}(\vbx)\equiv \vbz(\vbx)-  \frac{\mu_Z}{\sqrt{N_p}}\vec{\mathbf{1}}  - W^T\vbx}$. We have also defined the mean as $\mu_z/\sqrt{N_p}\vec{\mathbf{1}}$ where $\vec{\mathbf{1}}$ is a length-$N_p$ vector of ones. 

We prove that the linear and nonlinear terms are statistically independent with respect to the input features $\vbx$ as follows:
\begin{equation}
\begin{aligned}
\Cov[\vbx][W^T\vbx, \delta\vbz_{\mathrm{NL}}(\vbx)^T] &= \Cov[\vbx][W^T\vbx, \vbz^T(\vbx) - \frac{\mu_Z}{\sqrt{N_f}}\vec{\mathbf{1}}^T - \vbx^TW]\\
&= W^T\Cov[\vbx][\vbx, \vbz^T(\vbx) ] -  \frac{\mu_Z}{\sqrt{N_p}}W^T\Cov[\vbx][\vbx, \vec{\mathbf{1}}^T] - W^T\Cov[\vbx][\vbx, \vbx^T ]W\\
&= W^T\Sigma_{\vbx}W - W^T\Sigma_{\vbx}W\\
&= 0.
\end{aligned}
\end{equation}

We also show that $W$ as defined in Eq.~\eqref{eq:SIzdecomp} coincides with that in the definition of the hidden features,
\begin{equation}
\vbz(\vbx) = \frac{1}{\expval*{\varphi'}} \frac{\sigma_W\sigma_X}{\sqrt{N_p}}\varphi\qty( \frac{\sqrt{N_p}}{\sigma_W\sigma_X} W^T\vbx).
\end{equation}
As in the previous section, we evaluate the covariance in Eq.~\eqref{eq:SIzdecomp} and use the identity in Eq.~\eqref{eq:integral2} to find
\begin{equation}
\begin{aligned}
W &= \Sigma_{\vbx}^{-1}\Cov[\vbx][\vbx, \vbz^T(\vbx)]\\
&= \Sigma_{\vbx}^{-1} \frac{1}{\expval*{\varphi'}} \frac{\sigma_W\sigma_X}{\sqrt{N_p}} \E[\vbx]\qty[\vbx \varphi\qty(\frac{\sqrt{N_p}}{\sigma_W\sigma_X}\vbx^T W)]\\
&\approx  \Sigma_{\vbx}^{-1}\frac{1}{\expval*{\varphi'}} \frac{\sigma_W\sigma_X}{\sqrt{N_p}}  \frac{\sqrt{N_p}}{\sigma_W\sigma_X} \Sigma_{\vbx}W\expval*{\varphi'}\\
&= W.
\end{aligned}
\end{equation}
So we see that the definitions are consistent with one another.

Next, we calculate the covariance of the nonlinear components of the hidden features $\delta \vbz_{\mathrm{NL}}(\vbx)$ with respect to the full ensemble distribution.
To do this, we first calculate the mean of each hidden feature using the identity in Eq.~\eqref{eq:integral1},
\begin{equation}
\begin{aligned}
\E[\vbx]\qty[z_J(\vbx)] &= \frac{1}{\expval*{\varphi'}}\frac{\sigma_W\sigma_X}{\sqrt{N_p}}\E[\vbx]\qty[\varphi\qty(\frac{\sqrt{N_p}}{\sigma_W\sigma_X}\sum_k W_{kJ}x_k)] \approx  \frac{\sigma_W\sigma_X}{\sqrt{N_p}} \frac{\expval*{\varphi}}{\expval*{\varphi'}},
\end{aligned}
\end{equation}
from which we see that
\begin{equation}
\mu_z =  \sigma_W\sigma_X  \frac{\expval*{\varphi}}{\expval*{\varphi'}}.
\end{equation}
Similarly, we calculate the mean of the square of each hidden feature,
\begin{equation}
\begin{aligned}
\E[\vbx]\qty[z_J^2(\vbx)] &= \frac{1}{\expval*{\varphi'}^2}\frac{\sigma_W^2\sigma_X^2}{N_p}\E[\vbx]\qty[\varphi^2\qty(\frac{\sqrt{N_p}}{\sigma_W\sigma_X}\sum_k W_{kJ}x_k)] \approx  \frac{\sigma_W^2\sigma_X^2}{N_p} \frac{\expval*{\varphi^2}}{\expval*{\varphi'}^2}.
\end{aligned}
\end{equation}
Using these two results and the independence of the linear and nonlinear components of the hidden features, we calculate the variance of the nonlinear component of each hidden feature to be
\begin{equation}
\begin{aligned}
\Var[\vbx]\qty[\delta z_{\mathrm{NL}, J}(\vbx)] &= \frac{\sigma_W^2\sigma_X^2}{N_p} \frac{\expval*{\varphi^2} - \expval*{\varphi}^2 - \expval*{\varphi'}^2}{\expval*{\varphi'}^2}.
\end{aligned}
\end{equation}

Finally, we calculate the mean of the product of two different hidden features $J\neq K$ for the same input features using the identity in Eq.~\eqref{eq:integral3},
\begin{equation}
\begin{aligned}
\E[\vbx]\qty[z_J(\vbx)z_K(\vbx)]\approx \E[\vbx]\qty[z_J(\vbx)]\E[\vbx]\qty[z_K(\vbx)].
\end{aligned}
\end{equation}
We observe that different hidden features are independent in the thermodynamic limit.

Since there are no other random variables present in any of the formulas,
we summarize the statistical properties of the nonlinear components of the hidden features for independent data points $\vbx_a$ and $\vbx_b$ with full ensemble averages, giving us
\begin{equation}
\E\qty[\delta z_{\mathrm{NL}, J}(\vbx_a)] = 0 \qqc \Cov\qty[\delta z_{\mathrm{NL}, J}(\vbx_a), \delta z_{\mathrm{NL}, K}(\vbx_b)] = \frac{\sigma_{\delta z}^2}{N_P}\delta_{ab}\delta_{JK},
\end{equation}
where we have defined the variance $\sigma_{\delta z}^2$ of the nonlinear components as
\begin{equation}
\begin{gathered}
 \sigma_{\delta z}^2 =\sigma_W^2\sigma_X^2\Delta \varphi \qqc \Delta \varphi =  \frac{\expval*{\varphi^2} - \expval*{\varphi}^2 - \expval*{\varphi'}^2}{\expval*{\varphi'}^2}\\
  \expval*{\varphi} = \frac{1}{\sqrt{2\pi}}\int\limits_{-\infty}^\infty dh e^{-\frac{h^2}{2}}\varphi(h)\qqc\expval*{\varphi^2} = \frac{1}{\sqrt{2\pi}}\int\limits_{-\infty}^\infty dh e^{-\frac{h^2}{2}}\varphi^2(h)\qqc \expval*{\varphi'} = \frac{1}{\sqrt{2\pi}}\int\limits_{-\infty}^\infty dh e^{-\frac{h^2}{2}}\varphi'(h).\label{eq:dznl}
  \end{gathered}
\end{equation}

\subsection{General Solutions}\label{sec:gen_sol}

Next, we derive the forms of the general solutions reported in Eq.~\eqref{eq:gen_train}-\eqref{eq:gen_var} in the results section of the main text.
First, we consider the training error. Recall this error takes the form
\begin{equation}
\mathcal{E}_{\mathrm{train}} = \frac{1}{M}\sum_b \qty(\Delta y_b)^2.
\end{equation}
Taking the ensemble average, we express the training error as
\begin{equation}
\etrain = \expval*{\Delta y^2},\label{eq:SIgen_train}
\end{equation}
where we have defined the mean of the squared label errors as
\begin{equation}
\expval*{\Delta y^2} = \E\qty[\frac{1}{M}\sum_b \Delta y_b^2].
\end{equation}

Next, we consider the test error, bias and variance.
To evaluate these quantities, we first  decompose the predicted label for an arbitrary test data point $(y, \vbx)$ using the hidden feature decomposition in Eq.~\eqref{eq:SIzdecomp},
\begin{equation}
\hat{y}(\vbx) = \vbx\cdot \hbbeta + \frac{\mu_Z}{\sqrt{N_p}}\sum_J \hat{w}_J + \delta \vbz_{\mathrm{NL}}(\vbx)\cdot \hbw,\label{eq:yhatdecomp}
\end{equation}
where we have defined the estimated ground truth parameters $\hbbeta \equiv W\hbw$. It is interesting to note that we can identify a  definition for these parameters analagous to those of the ground truth parameters $\vbbeta$ in Eq.~\eqref{eq:SIydecomp},
\begin{equation}
\hbbeta \equiv \Sigma_{\vbx}^{-1}\Cov[\vbx]\qty[\vbx, \hat{y}(\vbx)].
\end{equation}

Because we consider data with labels that have zero mean, the label predictions should also have zero mean with respect to $\vbx$. 
Therefore, if the first and last terms in Eq.~\eqref{eq:yhatdecomp} each have zero mean with respect to $\vbx$, the second term should evaluate to zero to ensure $\hat{y}(\vbx)$ overall has zero mean.
For linear regression, this is clearly the case since $\mu_Z = 0$, but we will later prove that $\sum_J \hat{w}_J = 0$ for the random nonlinear features model.
For now, we will neglect this term for the remainder of this section.

We evaluate the test error on a data set, $\mathcal{D}'=\{(y_b', \vbx_b')\}_{b=1}^{M'}$, sampled independently from the same distribution as the training set. 
Recall that the test error is defined as
\begin{equation}
\mathcal{E}_{\mathrm{test}} = \frac{1}{M'}\sum_b \qty(\Delta y'_b)^2,
\end{equation}
where the sum ranges from $1$ to $M'$.
We note that the residual label error of each test data point is described by the same distribution in the ensemble.
Therefore, once we have taken the average over the test data, the test error can be expressed as an average over a single arbitrary test data point $(y, \vbx)$.
Using this fact and applying the label decomposition in Eq.~\eqref{eq:SIydecomp} and the predicted label decomposition in Eq.~\eqref{eq:yhatdecomp}, we find
\begin{equation}
\begin{aligned}
\E[\mathcal{D}']\qty[\mathcal{E}_{\mathrm{test}}] &= \E[(y, \vbx)]\qty[\qty(y(\vbx) - \hat{y}(\vbx))^2]\\
&= \E[(y, \vbx)]\qty[\qty(\vbx\cdot \Delta \vbbeta - \delta \vbz_{\mathrm{NL}}(\vbx)\cdot \hbw + \delta y^*_{\mathrm{NL}}(\vbx) + \varepsilon )^2]\\
&= \frac{\sigma_X^2 }{N_f}\sum_k \Delta \beta_k^2  + \frac{\sigma_{\delta z}^2}{N_p}\sum_K \hat{w}_K^2 + \sigma_{\delta y^*}^2 + \sigma_\varepsilon^2.
\end{aligned}
\end{equation}
Next, we apply the remainder of the ensemble average to find
\begin{equation}
\etest = \sigma_X^2 \expval*{\Delta\beta^2}+ \sigma_{\delta z}^2\expval*{\hat{w}^2} + \sigma_{\delta y^*}^2 + \sigma_\varepsilon^2,\label{eq:SIgen_test}
\end{equation}
where we have defined the average of the squared parameter errors and squared fit parameters, respectively, as
\begin{align}
\expval*{\Delta\beta^2} = \E\qty[\frac{1}{N_f}\sum_k \Delta \beta_k^2]\qqc\expval*{\hat{w}^2} = \E\qty[\frac{1}{N_p}\sum_K \hat{w}_K^2].
\end{align}

Next, recall that the squared bias is defined as
\begin{equation}
\begin{aligned}
\Bias^2[\hat{y}(\vbx)] = (\E[\mathcal{D}]\qty[\hat{y}(\vbx)] - y^*(\vbx))^2.
\end{aligned}
\end{equation}
Note that averaging over $\mathcal{D}$ implies averaging over only the features $X$ and noise $\vbeps$ of the training set.
In order to compute this average correctly, we make use of the following trick: we reinterpret the squared average over $\mathcal{D}$ as two separate averages over uncorrelated training data sets.
Now, instead of a single regression problem trained on a single data set $\mathcal{D}$, we consider two separate regression problems each trained independently on different training sets, $\mathcal{D}_1$ and $\mathcal{D}_2$, drawn from the same distribution with the same ground truth parameters $\vbbeta$.
These regression problems will also share all other random variables including the test data point $(y, \vbx)$, $W$, etc.
This allows us to express the squared bias as
\begin{equation}
\Bias^2[\hat{y}(\vbx)] = \E[\mathcal{D}_1, \mathcal{D}_2]\qty[\qty(y(\vbx) - \hat{y}_1(\vbx))\qty(y(\vbx) - \hat{y}_2(\vbx))],\label{eq:biastwodatasets}
\end{equation}
where we use subscripts $1$ and $2$ to denote quantities that result from training on data sets  $\mathcal{D}_1$ and $\mathcal{D}_2$, respectively. From here, we average over the test data point to find
\begin{equation}
\begin{aligned}
\E[\vbx]\qty[\Bias^2[\hat{y}(\vbx)]]  &= \E[\vbx, \mathcal{D}_1, \mathcal{D}_2]\qty[\qty(\hat{y}_1(\vbx) - y^*(\vbx))\qty(\hat{y}_2(\vbx) - y^*(\vbx))]\\
&= \E[\mathcal{D}_1, \mathcal{D}_2]\qty[\frac{\sigma_X^2 }{N_f}\sum_k \Delta \beta_{1, k}\Delta \beta_{2, k}] +\E[\mathcal{D}_1, \mathcal{D}_2]\qty[ \frac{\sigma_{\delta z}^2 }{N_p}\sum_K \hat{w}_{1, K}\hat{w}_{2, K}] + \sigma_{\delta y^*}^2.
\end{aligned}
\end{equation}
Next, we average over the remainder of the ensemble variables, giving us
\begin{equation}
\expval*{\Bias^2[\hat{y}(\vbx)]} = \sigma_X^2 \expval*{\Delta\beta_1\Delta\beta_2} + \sigma_{\delta z}^2 \expval*{\hat{w}_1\hat{w}_2}+ \sigma_{\delta y^*}^2,\label{eq:SIgen_bias}
\end{equation}
where we have defined the quantities,
\begin{align}
\expval*{\Delta\beta_1\Delta\beta_2} = \E\qty[\frac{1}{N_f}\sum_k \Delta \beta_{1, k}\Delta\beta_{2,k}]\qqc
\expval*{\hat{w}_1\hat{w}_2} = \E\qty[\frac{1}{N_p}\sum_K \hat{w}_{1, K}\hat{w}_{2, K}].
\end{align}

To derive the expression for the ensemble-averaged variance, we simply make use of the bias-variance decomposition in Eq.~\eqref{eq:bvdecomp}, giving us
\begin{equation}
\begin{aligned}
\expval*{\Var[][\hat{y}(\vbx)]}  &= \expval*{\mathcal{E}_{\mathrm{test}}} - \expval*{\Bias^2[\hat{y}(\vbx)]} - \mathrm{Noise}\\
&= \sigma_X^2  \qty(\expval*{\Delta\beta^2}- \expval*{\Delta\beta_1\Delta\beta_2}) + \sigma_{\delta z}^2\qty(\expval*{\hat{w}^2} - \expval*{\hat{w}_1\hat{w}_2}).\label{eq:SIgen_var}
\end{aligned}
\end{equation}

Based on these expressions, we find that the training error, test error, bias, and variance depend on five key ensemble-averaged quantities: $\expval*{\Delta y^2}$, $\expval*{\Delta \beta^2}$, $\expval*{\hat{w}^2}$, $\expval*{\Delta\beta_1\Delta\beta_2}$, and $\expval*{\hat{w}_1\hat{w}_2}$.

\subsection{Linear Regression (No Basis Functions)}

In linear regression without basis functions, the hidden features are the same as the input features,
\begin{equation}
\vbz(\vbx) = \vbx.
\end{equation}
Using this definition for the hidden features, we decompose the equation for the gradient in Eq.~\eqref{eq:SIgrad} into three sets of equations for the fit parameters, residual label errors, and residual parameters errors,
\begin{equation}
\begin{aligned}
\lambda \hat{w}_j &= \sum_b \Delta y_b  X_{bj}  + \eta_j\\
\Delta y_a &= \sum_k \Delta \beta_k X_{ak} + \delta y^*_{\mathrm{NL}}(\vbx_a) + \varepsilon_a + \xi_a\\
\Delta \beta_j &= \beta_j - \hat{w}_j,
\end{aligned}
\end{equation}
where we have also utilized Eq.~\eqref{eq:SIydecomp} to decompose the training labels into linear and nonlinear components.
We have also added small auxiliary fields $\eta_j$ and $\xi_a$ to the two equations containing sums.
We will use these extra fields to define perturbations about the solutions to these equations with the intent of setting the fields to zero by the end of the derivation.

\subsubsection{Cavity Expansion}

Next, we add an additional variable of each type, resulting in a total of $M+1$ data points and $N_f+1$ features.
We specify each new variable using an index value of $0$, giving us the new unknown quantities $\hat{w}_0$, $\Delta y_0$, and $\Delta \beta_0$. 
These new variables result in the addition of an extra term in each sum, giving us the equations
\begin{equation}
\begin{aligned}
\lambda \hat{w}_j &= \sum_b \Delta y_b X_{bj} + \eta_j + \Delta y_0 X_{0j} \\
\Delta y_a &= \sum_k \Delta \beta_k X_{ak}  + \delta y^*_{\mathrm{NL}}(\vbx_a) + \varepsilon_a + \xi_a + \Delta\beta_0 X_{a0} .\label{eq:linreg_plus0}
\end{aligned}
\end{equation}
Each new variable is also described by a new equation,
\begin{equation}
\begin{aligned}
\lambda \hat{w}_0 &=\sum_b  \Delta y_b X_{b 0} + \eta_0 +  \Delta y_0 X_{0 0}\\
\Delta y_0 &=  \sum_k \Delta \beta_k X_{0 k}  + \delta y^*_{\mathrm{NL}}(\vbx_0) + \varepsilon_0 + \xi_0 +  \Delta \beta_0 X_{0 0} \\
\Delta \beta_0 &= \beta_0 - \hat{w}_0.\label{eq:linreg_zeroeq}
\end{aligned}
\end{equation}
As a reminder, sums always start at an index value of $1$. Therefore, we explicitly specify any terms with an index value of $0$.

Now we take the thermodynamic limit in which $M$ and $N_f$ tend towards infinity, but their ratio $\alpha_f$ remains fixed. 
In this limit, we can interpret the extra terms in Eq.~\eqref{eq:linreg_plus0} as small perturbations to the auxiliary fields since they each contain an element of $X$ which has mean zero and an infinitesimal variance of $\order{1/N_f}$,
\begin{equation}
\delta\eta_j = \Delta y_0X_{0j}\qqc \delta\xi_a = \Delta\beta_0X_{a0}  .
\end{equation}
This allows us to expand each variable about its solution in the absence of the $0$-indexed quantities, corresponding to the solution for $M$ data points and $N_f$ features,
\begin{equation}
\begin{aligned}
\hat{w}_j &\approx \hat{w}_{j\setminus 0} + \sum_k \nu^{\hat{w}}_{jk}\delta \eta_k + \sum_b \chi^{\hat{w}}_{jb}\delta \xi_b \\
\Delta y_a &\approx \Delta y_{a\setminus 0} + \sum_k \nu^{\Delta y}_{ak}\delta \eta_k + \sum_b \chi^{\Delta y}_{ab}\delta \xi_b \\
\Delta \beta_j &\approx \Delta \beta_{j\setminus 0} + \sum_k \nu^{\Delta \beta}_{jk}\delta \eta_k + \sum_b \chi^{\Delta \beta}_{jb}\delta \xi_b .\label{eq:linreg_cavexp}
\end{aligned}
\end{equation}
We use subscripts with $\setminus 0$ to refer to the unperturbed solutions for each unknown quantity; that is, the solutions in the absence of the 0-indexed variables.
We also define the susceptibility matrices as the following derivatives with respect to the auxiliary fields:
\begin{equation}
\begin{alignedat}{2}
\nu^{\hat{w}}_{jk} &= \pdv{\hat{w}_j}{\eta_k}\qqc & \chi^{\hat{w}}_{jb} &= \pdv{\hat{w}_j}{\xi_b},\\
\nu^{\Delta y}_{ak} &= \pdv{\Delta y_a}{\eta_k}\qqc & \chi^{\Delta y}_{ab} &= \pdv{\Delta y_a}{\xi_b},\\
\nu^{\Delta \beta}_{jk} &= \pdv{\Delta \beta_j}{\eta_k}\qqc & \chi^{\Delta \beta}_{jb} &= \pdv{\Delta \beta_j}{\xi_b}.
\end{alignedat}
\end{equation}
It is useful to note that the susceptibilities for the residual parameter errors are related to those for the fit parameters via a negative sign,
\begin{equation}
 \nu^{\Delta \beta}_{jk} = - \nu^{\hat{w}}_{jk}\qqc\chi^{\Delta \beta}_{jb}  = -\chi^{\hat{w}}_{jb}.
\end{equation}
Therefore, we replace all susceptibilities for the residual parameter errors with their fit parameter counterparts.
Substituting the expansions in Eq.~\eqref{eq:linreg_cavexp} into the equations for the 0-indexed variables, Eq.~\eqref{eq:linreg_zeroeq}, we arrive at the following equations:
\begin{equation}
\begin{aligned}
\lambda \hat{w}_0 &=\sum_a\qty( \Delta y_{a\setminus 0} + \sum_k \nu^{\Delta y}_{ak}\delta \eta_k + \sum_b \chi^{\Delta y}_{ab}\delta \xi_b) X_{a 0} + \eta_0 + X_{0 0} \Delta y_0\\
\Delta y_0 &=  \sum_j\qty(\Delta \beta_{j\setminus 0} - \sum_k \nu^{\hat{w}}_{jk}\delta \eta_k - \sum_b \chi^{\hat{w}}_{jb}\delta \xi_b) X_{0 j}  + \delta y^*_{\mathrm{NL}}(\vbx_0) + \varepsilon_0 + \xi_0 +  \Delta \beta_0X_{0 0}.\label{eq:linreg_zero_cavexp}
\end{aligned}
\end{equation}
Our next step is to simplify these equations by approximating the sums over large numbers of random variables.

\subsubsection{Central Limit Approximations}

Each of the sums in Eq.~\eqref{eq:linreg_zero_cavexp} contains a thermodynamically large number of statistically uncorrelated terms. 
This means that each sum satisfies the conditions necessary to apply the central limit theorem,
allowing us to express each in terms of a single normally-distributed random variable described by just its mean and its variance.
First, we approximate the two sums that contain one of the unperturbed unknown quantities, $\Delta y_{a\setminus 0}$ or $\Delta\beta_{j\setminus 0}$. 
In both sums, the unperturbed quantities are statistically independent of any elements of $X$ with a $0$-index such as $X_{a0}$ or $X_{0j}$.
Using this independence, we apply the identity in Eq.~\eqref{eq:clt_vec} to find
\begin{equation}
\begin{alignedat}{3}
\sum_b  \Delta y_{b \setminus 0} X_{b 0} &\approx \sigma_{\hat{w}}z_{\hat{w}} \qqc&
 \sigma_{\hat{w}}^2 &= \sigma_X^2\alpha_f^{-1}\expval*{\Delta y^2} \qqc & \expval*{\Delta y^2} &=   \frac{1}{M}\sum_b\Delta y_{b\setminus 0}^2\\
 \sum_k \Delta \beta_{k\setminus 0}X_{0 k} &\approx \sigma_{\Delta y}z_{\Delta y} \qqc
& \sigma_{\Delta y}^2 &=  \sigma_X^2\expval*{\Delta \beta^2} \qqc &\expval*{\Delta \beta^2} &= \frac{1}{N_f}\sum_k\Delta \beta_{k \setminus 0}^2,
\end{alignedat}
\end{equation}
where $ \sigma_{ \hat{w} }^2$ and $\sigma_{\Delta y}^2$ are the total variances of the two sums and  $z_{\hat{w}}$ are $z_{\Delta y}$ are random variables with zero mean and unit variance. 
It is straightforward to show that $z_{ \hat{w} }$ and $z_{\Delta y}$ are statistically independent since the two sums are independent with respect to the zero-indexed elements of $X$.

Note that we have used the same notation, $\expval*{\Delta y^2}$ and $\expval*{\Delta \beta^2}$, for the two averages that defined previously in Sec.~\ref{sec:gen_sol} even though they each lack an ensemble average. In doing so, we have made the ansatz that these sums will converge to their ensemble averages in the thermodynamic limit.
This assumption is typical of the cavity method.

Next, we approximate the sums that include either of the square susceptibility matrices, $\chi_{ab}^{\Delta y}$ or $\nu^{\hat{w}}_{jk}$. 
Similar to the unperturbed unknown quantities, we use the property that the susceptibilities are statistically independent of the elements of $X$ with $0$-valued indices.
Applying the identity in Eq.~\eqref{eq:clt_square}, we find
\begin{equation}
\begin{alignedat}{2}
\sum_{ab} \chi^{\Delta y}_{ab}X_{a 0}X_{b 0} &\approx   \sigma_X^2\alpha_f^{-1}\chi \qqc &\chi &=  \frac{1}{M}\sum_b\chi^{\Delta y}_{bb}\\
\sum_{jk} \nu^{\hat{w}}_{jk}X_{0 j}X_{0 k} &\approx \sigma_X^2\nu \qqc& \nu& = \frac{1}{N_f}\sum_k  \nu^{\hat{w}}_{kk},
\end{alignedat}
\end{equation}
where $\chi$ and $\nu$ can be interpreted as a pair of scalar susceptibilities.

The remainder of the sums contain rectangular susceptibility matrices which follow the form in Eq.~\eqref{eq:clt_rect}. Therefore, each of these sums is expected to be small in the thermodynamic limit and can be neglected.

\subsubsection{Self-consistency Equations}

Applying the approximations from the previous section to Eq.~\eqref{eq:linreg_zero_cavexp}, we obtain the following set of self-consistency equations for the $0$-indexed variables, $\hat{w}_0$, $\Delta y_0$, and $\Delta \beta_0$:
\begin{equation}
\begin{aligned}
\lambda \hat{w}_0 &\approx \sigma_{\hat{w} }z_{\hat{w}} + \Delta \beta_0 \sigma_X^2\alpha_f^{-1}\chi + \eta_0\\
\Delta y_0 &\approx \sigma_{\Delta y}z_{\Delta y} - \Delta y_0 \sigma_X^2\nu +  \delta y^*_{\mathrm{NL}}(\vbx_0)+ \varepsilon_0 + \xi_0 \\
\Delta \beta_0 &\approx \beta_0 - \hat{w}_0.
\end{aligned}
\end{equation}
In these equations, we have also dropped terms proportional to $X_{00}$ since this quantity has zero mean and a variance which goes to zero in the thermodynamic limit. 
Next, we solve these three equations for the $0$-indexed variables, giving us
\begin{equation}
\begin{aligned}
\hat{w}_0 &= \frac{  \beta_0 \sigma_X^2\alpha_f^{-1}\chi +\sigma_{\hat{w}}z_{\hat{w}}  + \eta_0}{\lambda + \sigma_X^2\alpha_f^{-1}\chi}\\
\Delta y_0 &= \frac{\sigma_{\Delta y}z_{\Delta y} + \delta y^*_{\mathrm{NL}}(\vbx_0)  + \varepsilon_0 + \xi_0 }{1 + \sigma_X^2\nu }\\
\Delta \beta_0 &= \frac{\beta_0 \lambda -\sigma_{\hat{w} }z_{ \hat{w}}  - \eta_0}{\lambda + \sigma_X^2\alpha_f^{-1}\chi}.\label{eq:linreg_zeroselfcon}
\end{aligned}
\end{equation}
Note that all random variables within each of the above equations are statistically independent from one another.

Next, we make the approximation  that in the thermodynamic limit, each of the unknown quantities is ``self-averaging.''
In other words, we assume that an average over a set of non-$0$-indexed variables is equivalent to taking an ensemble average of the single corresponding $0$-indexed variable.

This allows us to use Eq.~\eqref{eq:linreg_zeroselfcon} to find a set of self-consistent equations for the scalar susceptibilities by evaluating the appropriate derivatives with respect to the $0$-indexed auxiliary fields and performing ensemble averages,
\begin{equation}
\begin{aligned}
\chi  &= \frac{1}{M}\sum_b\chi_{bb}^{\Delta y} \approx  \E\qty[\chi_{00}^{\Delta y}] =\E\qty[ \pdv{\Delta y_0 }{\xi_0}]= \frac{1}{1 + \sigma_X^2\nu }\\
\nu &= \frac{1}{N_f}\sum_k \nu_{kk}^{\hat{w}} \approx \E\qty[\nu^{\hat{w}}_{00}] = \E\qty[\pdv{\hat{w}_0 }{\eta_0}] = \frac{1}{\lambda + \sigma_X^2\alpha_f^{-1}\chi}.\label{eq:linreg_susceptselfcon}
\end{aligned}
\end{equation}
Furthermore, we find the following self-consistent equations for the quantities, $\expval*{\hat{w}^2}$, $\expval*{\Delta y^2}$, and $\expval*{\Delta \beta^2}$ by taking the appropriate expectation values of the $0$-indexed quantities, plugging in the forms of the scalar susceptibilities, and setting the auxiliaries fields to zero:
\begin{equation}
\begin{aligned}
\expval*{\hat{w}^2} &= \frac{1}{N_f}\sum_k \hat{w}_{k\setminus 0}^2 \approx \E\qty[\hat{w}_0^2] = \nu^2\qty(\sigma_\beta^2\sigma_X^4 \alpha_f^{-2}\chi^2 + \sigma_X^2\alpha_f^{-1}\expval*{\Delta y^2})\\
\expval*{\Delta y^2}  &= \frac{1}{M}\sum_b \Delta y_{b\setminus 0}^2 \approx \E\qty[\Delta y^2_0] =  \chi^2 \qty(\sigma_X^2 \expval*{\Delta \beta^2} + \sigma_{\delta y^*}^2 + \sigma_\varepsilon^2  )\\
\expval*{\Delta \beta^2} &= \frac{1}{N_f}\sum_k \Delta \beta_{k\setminus 0}^2 \approx \E\qty[\Delta \beta_0^2] = \nu^2\qty(\sigma_\beta^2 \lambda^2 + \sigma_X^2\alpha_f^{-1} \expval*{\Delta y^2}).\label{eq:linreg_squareselfcon}
\end{aligned}
\end{equation}
Note that we have also defined the mean squared fit parameter size $\expval*{\hat{w}^2}$.
In addition, each of the three mean squared quantities can be interpreted as a full ensemble average.
These self-consistent equations, along with those for the scalar susceptibilities, capture almost all behavior of our model of linear regression in the thermodynamic limit.

\subsubsection{Solutions with Finite Regularization ($\lambda \sim \order{1}$)}\label{sec:linreg_finite}

Next, we derive the solutions when the regularization parameter $\lambda$ is finite.
By combing the two scalar susceptibilities in Eq.~\eqref{eq:linreg_susceptselfcon}, we find a quadratic equation for $\chi$,
\begin{equation}
 \chi^2 + \qty[ \qty(\alpha_f-1) + \bar{\lambda}\alpha_f]\chi - \bar{\lambda}\alpha_f = 0,\label{eq:linreg_poly}
\end{equation}
where we have defined the dimensionless regularization parameter
\begin{equation}
\bar{\lambda} = \frac{\lambda}{\sigma_X^2}.
\end{equation}
Solving Eq.~\eqref{eq:linreg_poly}, we find two solutions:
\begin{equation}
\chi = \frac{1}{2} \qty[ 1-\alpha_f\qty(1+\bar{\lambda} ) \pm \sqrt{\qty[1-\alpha_f\qty(1+ \bar{\lambda} )]^2 + 4\alpha_f\bar{\lambda}}].\label{eq:linreg_nuexact}
\end{equation}
Using these solutions we can also find similar solutions for $\nu$. 
Next, we solve Eq.~\eqref{eq:linreg_squareselfcon} to find closed-form solutions for $\expval*{\hat{w}^2}$, $\expval*{\Delta y^2}$, and $\expval*{\Delta \beta^2}$:
\begin{equation}
\begin{aligned}
\mqty(\expval*{\hat{w}^2} \\ \expval*{\Delta y^2} \\ \expval*{\Delta \beta^2})
 =  \mqty(1 & -\sigma_X^2\alpha_f^{-1}\nu^2 & 0\\
 0 & 1 & -\sigma_X^2\chi^2\\
 0 & -\sigma_X^2\alpha_f^{-1}\nu^2 & 1)^{-1}
 \mqty(\sigma_\beta^2\sigma_X^4\alpha_f^{-2}\chi^2\nu^2\\ (\sigma_\varepsilon^2 + \sigma_{\delta y^*}^2)\chi^2 \\ \bar{\lambda}^2\sigma_\beta^2\sigma_X^2 \nu^2).
\end{aligned}
\end{equation}
In combination with the solutions for $\chi$ and $\nu$, these solutions are exact in the thermodynamic limit.

\subsubsection{Solutions in Ridge-less Limit ($\lambda\rightarrow 0$)}

In order to make the solutions in the previous section easier to interpret, we take the ridge-less limit where $\lambda\rightarrow 0$. 
Based on the form of Eq.~\eqref{eq:linreg_poly}, we make the ansatz that the lowest order contribution to $\chi$ is  $\order*{1}$ in small $\bar{\lambda}$. Accordingly, we expand $\chi$ in small $\bar{\lambda}$ up to $\order*{\bar{\lambda}}$ as
\begin{equation}
\chi \approx \chi_0 + \bar{\lambda}  \chi_1.
\end{equation}
Substituting this approximation into the formula for $\chi$ in Eq.~\eqref{eq:linreg_poly}, we find the following equation at $\order*{1}$:
\begin{equation}
\begin{aligned}
 0 &= \chi_0^2 + (\alpha_f-1)\chi_0.
\end{aligned}
\end{equation}
Solving this equation, we find two solutions for $\chi_0$,
\begin{equation}
\chi_0^{(1)} = 1-\alpha_f \qqc \chi_0^{(2)} = 0.
\end{equation}
We label each set of solutions for all quantities with a superscript $(1)$ or $(2)$.
These two solutions correspond to the two solutions in the exact formula for $\chi$ in Eq.~\eqref{eq:linreg_nuexact}.
Next, we collect terms in Eq.~\eqref{eq:linreg_poly} at $\mathcal{O}(\bar{\lambda})$,
\begin{equation}
0 = 2   \chi_0\chi_1 +   \qty(\alpha_f-1)  \chi_1 +   \alpha_f(\chi_0 - 1).
\end{equation}
Solving, we obtain an equation for $\chi_1$ in terms of $\chi_0$,
\begin{equation}
\chi_1 = \frac{\alpha_f(1 - \chi_0)}{  2 \chi_0 +\alpha_f - 1  }.
\end{equation}
Combining this equation with $\chi_0^{(2)}=0$, we obtain the leading order term for solution (2),
\begin{equation}
\chi_1^{(2)} =  \frac{\alpha_f}{ \alpha_f-1}.
\end{equation}

Next, we solve for the two solutions for $\nu$. By inspecting the equation for $\nu$ in terms of $\chi$ in Eq.~\eqref{eq:linreg_susceptselfcon}, we make the ansatz that the lowest contribution of $\nu$ is $\mathcal{O}(1/\lambda)$,
\begin{equation}
\nu \approx \frac{1}{\bar{\lambda}}\nu_{-1} + \nu_0.
\end{equation}
Substituting the solutions for $\chi_0$ and $\chi_1$ into the equation for $\nu$, we find that the solutions for $\nu_{-1}$ are
\begin{equation}
\nu_{-1}^{(1)} = 0 \qqc
\nu_{-1}^{(2)} = \frac{1}{\sigma_X^2 + \frac{\sigma_X^2}{\alpha_f}\chi_1} = \frac{1}{\sigma_X^2} \frac{\alpha_f-1}{ \alpha_f}.
\end{equation}
Since $\nu_{-1}^{(1)}$ is zero, we also solve for the next order term for solution $(1)$,
\begin{equation}
\nu_0^{(1)} = \frac{1}{\sigma_X^2} \frac{\alpha_f}{ (1- \alpha_f)}.
\end{equation}
For completion, we also find that we can continue with this procedure to derive 
\begin{equation}
\nu_0^{(2)} = \frac{1}{\sigma_X^2} \frac{1}{(\alpha_f-1)}.
\end{equation}

We also expand each of the ensemble-averaged quantities $\expval*{\hat{w}^2}$, $\expval*{\Delta y^2}$, and $\expval*{\Delta \beta^2}$ in small $\bar{\lambda}$.
We make the ansatz that each of these quantities is $\order*{1}$ to lowest order with the next terms in the expansion at $\order*{\bar{\lambda}^2}$:
\begin{equation}
\begin{aligned}
\expval*{\hat{w}^2} &\approx \expval*{\hat{w}^2}_0 + \bar{\lambda}^2\expval*{\hat{w}^2}_2\\
\expval*{\Delta y^2} &\approx \expval*{\Delta y^2}_0 + \bar{\lambda}^2 \expval*{\Delta y^2}_2\\
\expval*{\Delta \beta^2} &\approx \expval*{\Delta \beta^2}_0 + \bar{\lambda}^2 \expval*{\Delta \beta^2}_2.
\end{aligned}
\end{equation}

\paragraph*{Solution (1):}
For the first set of solutions, the self-consistent equations, Eq.~\eqref{eq:linreg_squareselfcon}, to lowest order, are
\begin{equation}
\begin{aligned}
\expval*{\hat{w}^2}_0^{(1)} &= (\nu_0^{(1)})^2\qty[\sigma_\beta^2\sigma_X^4 \alpha_f^{-2}(\chi_0^{(1)})^2 + \sigma_X^2\alpha_f^{-1}\expval*{\Delta y^2}_0^{(1)}]\\
\expval*{\Delta y^2}_0^{(1)}  &= (\chi_0^{(1)})^2 \qty[\sigma_X^2 \expval*{\Delta \beta^2}_0^{(1)} + \sigma_{\delta y^*}^2 + \sigma_\varepsilon^2  ]\\
\expval*{\Delta \beta^2}_0^{(1)} &=  (\nu_0^{(1)})^2 \sigma_X^2\alpha_f^{-1} \expval*{\Delta y^2}_0^{(1)}. 
\end{aligned}
\end{equation}
Substituting the solutions for the susceptibilities into these equations and solving, we find
\begin{equation}
\begin{aligned}
\expval*{\hat{w}^2}_0^{(1)} &=  \sigma_\beta^2  + \frac{1}{\sigma_X^2}(\sigma_\varepsilon^2 + \sigma_{\delta y^*}^2) \frac{\alpha_f}{(1-\alpha_f)}\\
\expval*{\Delta y^2}_0^{(1)} &=  (\sigma_\varepsilon^2 + \sigma_{\delta y^*}^2)(1-\alpha_f)\\
\expval*{\Delta \beta^2}_0^{(1)} &=  \frac{1}{\sigma_X^2}(\sigma_\varepsilon^2 + \sigma_{\delta y^*}^2)\frac{\alpha_f}{(1-\alpha_f)}.
\end{aligned}
\end{equation}

\paragraph*{Solution (2):} For the second set of solutions, the self-consistent equations to lowest order, are
\begin{equation}
\begin{aligned}
\expval*{\hat{w}^2}_0^{(2)} &= (\nu_{-1}^{(2)})^2\qty[\sigma_\beta^2\sigma_X^4\alpha_f^{-2}(\chi_1^{(2)})^2 + \sigma_X^2\alpha_f^{-1}\expval*{\Delta y^2}_2^{(2)}]\\
\expval*{\Delta y^2}_0^{(2)}  &= 0\\
\expval*{\Delta \beta^2}_0^{(2)} &=  (\nu_{-1}^{(2)})^2 \qty[\sigma_\beta^2\sigma_X^4 +  \sigma_X^2\alpha_f^{-1} \expval*{\Delta y^2}_2^{(2)}].
\end{aligned}
\end{equation}
We see that we also need the solution for $\expval*{\Delta y^2}$ to next lowest order,
\begin{equation}
\expval*{\Delta y^2}_2^{(2)}  = (\chi_1^{(2)})^2 \qty[\sigma_X^2 \expval*{\Delta \beta^2}_0^{(2)} + \sigma_{\delta y^*}^2 + \sigma_\varepsilon^2  ].
\end{equation}
Substituting the solutions for the susceptibilities into these equations and solving, we find
\begin{equation}
\begin{aligned}
\expval*{\hat{w}^2}_0^{(2)} &=  \sigma_\beta^2\frac{1}{\alpha_f} + \frac{1}{\sigma_X^2}(\sigma_\varepsilon^2 + \sigma_{\delta y^*}^2) \frac{1}{(\alpha_f-1)}\\
\expval*{\Delta y^2}_2^{(2)} &=  \sigma_\beta^2\sigma_X^2 \frac{\alpha_f}{(\alpha_f-1)} + (\sigma_\varepsilon^2 + \sigma_{\delta y^*}^2) \frac{\alpha_f^3}{(\alpha_f-1)^3}\\
\expval*{\Delta \beta^2}_0^{(2)} &= \sigma_\beta^2 \frac{(\alpha_f-1)}{\alpha_f} +   \frac{1}{\sigma_X^2}(\sigma_\varepsilon^2 + \sigma_{\delta y^*}^2) \frac{1}{(\alpha_f-1)}.
\end{aligned}
\end{equation}

\paragraph*{Combined solutions:}
To determine when each of the two solutions applies, we use the fact that each of the ensemble-averaged quantities $\expval*{\hat{w}^2}$, $\expval*{\Delta y^2}$, and $\expval*{\Delta \beta^2}$ must always be positive by definition.
Imposing this constraint, we find that solution (1) only applies when $\alpha_f < 1$, while solution (2) only applies when $\alpha_f > 1$. 
Combining these solutions, we arrive at the final forms for the three ensemble-averaged quantities in the $\lambda\rightarrow 0$ limit,
\begin{equation}
\begin{aligned}
\expval*{\hat{w}^2} &= \left\{
\begin{array}{cl}
\sigma_\beta^2  + \frac{(\sigma_\varepsilon^2 + \sigma_{\delta y^*}^2)}{\sigma_X^2} \frac{\alpha_f}{(1-\alpha_f)}& \qif N_f < M\\
\sigma_\beta^2\frac{1}{\alpha_f} + \frac{(\sigma_\varepsilon^2 + \sigma_{\delta y^*}^2)}{\sigma_X^2}\frac{1}{(\alpha_f-1)} & \qif N_f > M
\end{array}
\right.\\
\expval*{\Delta y^2} &= \left\{\begin{array}{cl}
  (\sigma_\varepsilon^2 + \sigma_{\delta y^*}^2)(1-\alpha_f) & \qif N_f < M\\
\frac{\lambda^2}{\sigma_X^4}\qty[ \sigma_\beta^2\sigma_X^2 \frac{\alpha_f}{(\alpha_f-1)} +  (\sigma_\varepsilon^2 + \sigma_{\delta y^*}^2) \frac{\alpha_f^3}{(\alpha_f-1)^3}] & \qif N_f > M
\end{array} \right.\\
\expval*{\Delta\beta^2} &= \left\{\begin{array}{cl}
\frac{ (\sigma_\varepsilon^2 + \sigma_{\delta y^*}^2)}{\sigma_X^2}\frac{\alpha_f}{(1-\alpha_f)} & \qif N_f < M\\
 \sigma_\beta^2 \frac{(\alpha_f-1)}{\alpha_f} +   \frac{ (\sigma_\varepsilon^2 + \sigma_{\delta y^*}^2) }{\sigma_X^2}\frac{1}{(\alpha_f-1)}  & \qif N_f > M.
\end{array} \right. 
\end{aligned}
\end{equation}

For completeness, we also report solutions for the two scalar susceptibilities,
\begin{equation}
\chi = \left\{
\begin{array}{cl}
1 - \alpha_f & \qif N_f < M\\
\frac{\lambda}{\sigma_X^2} \frac{\alpha_f}{(\alpha_f-1)} & \qif N_f > M
\end{array}
\right.\qqc
\nu = \left\{
\begin{array}{cl}
 \frac{1}{\sigma_X^2}\frac{\alpha_f}{(1-\alpha_f)} & \qif N_f < M\\
 \frac{1}{\lambda} \frac{(\alpha_f-1)}{\alpha_f} + \frac{1}{\sigma_X^2} \frac{1}{(\alpha_f - 1)}& \qif N_f > M.
\end{array}
\right.
\end{equation}

Finally, we use these expressions to derive the training and test error according to the general solutions in Eqs.~\eqref{eq:SIgen_train} and \eqref{eq:SIgen_test},
\begin{align}
\etrain &=  \left\{\begin{array}{cl}
  (\sigma_\varepsilon^2 + \sigma_{\delta y^*}^2)(1-\alpha_f) & \qif N_f < M\\
\frac{\lambda^2}{\sigma_X^4}\qty[ \sigma_\beta^2\sigma_X^2 \frac{\alpha_f}{(\alpha_f-1)} +  (\sigma_\varepsilon^2 + \sigma_{\delta y^*}^2) \frac{\alpha_f^3}{(\alpha_f-1)^3}] & \qif N_f > M
\end{array} \right.\\
\etest &= \left\{\begin{array}{cl}
(\sigma_\varepsilon^2+\sigma_{\delta y^*}^2)\frac{1}{(1-\alpha_f)} & \qif N_f < M\\
\sigma_\beta^2 \sigma_X^2\frac{(\alpha_f-1)}{\alpha_f} + (\sigma_\varepsilon^2+\sigma_{\delta y^*}^2) \frac{\alpha_f}{(\alpha_f-1)}  & \qif N_f > M.
\end{array} \right.
\end{align}

\subsubsection{Bias-Variance Decomposition}\label{sec:linreg_bv}

Next, we derive the bias and variance. According to the general solutions in Eqs.~\eqref{eq:SIgen_bias} and \eqref{eq:SIgen_var}, we only require the quantity  $\expval*{\Delta\beta_1\Delta\beta_2}$ since $\sigma_{\delta z}^2 = 0$.
To find $\expval*{\Delta\beta_1\Delta\beta_2}$, we use the formula for $\Delta\beta_0$ in Eq.~\eqref{eq:linreg_zeroselfcon} to characterize its behavior when trained separately on two independent training sets, $\mathcal{D}_1$ and $\mathcal{D}_2$,
\begin{equation}
\begin{aligned}
\Delta \beta_{1, 0} &= \nu\qty( \beta_0 \lambda -\sigma_{\hat{w} }z_{\hat{w}_1} )\\
\Delta \beta_{2, 0} &= \nu\qty( \beta_0 \lambda-\sigma_{\hat{w} }z_{\hat{w}_2} ).
\end{aligned}
\end{equation}
As a reminder, we use subscripts $1$ and $2$ to denote quantities that depend on one of the training sets.
Note that while the random variables $z_{\hat{w}_1}$ and $z_{\hat{w}_2}$ are defined separately for the two regression problems, both equations share the same $\beta_0$.
Multiplying these two equations together and using the self-averaging approximation, we find an expression for $\expval*{\Delta\beta_1\Delta\beta_2}$,
\begin{equation}
\expval*{\Delta\beta_1\Delta\beta_2} = \frac{1}{N_f}\sum_k \Delta\beta_{1, k}\Delta\beta_{2, } \approx \E\qty[\Delta\beta_{1,0} \Delta\beta_{2,0}] = \nu^2\qty(\sigma_\beta^2 \lambda^2+ \E\qty[\sigma_{\hat{w} }^2z_{\hat{w_1}}z_{\hat{w}_2}]). \label{eq:linreg_corrselfcon}
\end{equation}
Next, we calculate the expectation value of the product $z_{\hat{w_1}}z_{\hat{w}_2}$ and find that it evaluates to zero as a result of the statistical independence of the two design matrices, $X_1$ and $X_2$,
\begin{equation}
\begin{aligned}
\E\qty[\sigma_{\hat{w} }^2z_{\hat{w}_1}z_{\hat{w}_2}] &\approx \E\qty[\sum_{ab} \Delta y_{1, a\setminus 0}\Delta y_{2, b\setminus 0}X_{1, a 0}X_{2, b 0}]\\
&= \sum_{ab}\E\qty[ \Delta y_{1, a\setminus 0}\Delta y_{2, b\setminus 0}]\E\qty[X_{1, a 0}X_{2, b 0}] \\
&= 0.
\end{aligned}
\end{equation}
Substituting this solution into Eq.~\eqref{eq:linreg_corrselfcon}, we find an expression for $\expval*{\Delta\beta_1\Delta\beta_2}$,
\begin{equation}
\expval*{\Delta\beta_1\Delta\beta_2} = \sigma_\beta^2 \lambda^2 \nu^2.
\end{equation}
Inserting the solutions for $\nu$, we find in the $\lambda\rightarrow 0$ limit that
\begin{equation}
\expval*{\Delta\beta_1\Delta\beta_2}  = \left\{
\begin{array}{cl}
\frac{\lambda^2}{\sigma_X^2} \sigma_\beta^2 \frac{\alpha_f^2}{(1-\alpha_f)^2} 
 & \qif N_f < M\\
 \sigma_\beta^2 \frac{(\alpha_f-1)^2}{\alpha_f^2} & \qif N_f > M.
\end{array}
\right.
\end{equation}
From this, we arrive at the final expressions for the model bias and variance in the $\lambda\rightarrow 0$ limit,
\begin{equation}
\begin{aligned}
\expval*{\Bias^2[\hat{y}(\vbx)]} &= \left\{
\begin{array}{cl}
 \sigma_{\delta y^*}^2 & \qif N_f < M\\
\sigma_\beta^2\sigma_X^2 \frac{(\alpha_f-1)^2}{\alpha_f^2} +  \sigma_{\delta y^*}^2  & \qif N_f > M
\end{array}
\right.\\
\expval*{\Var[][ \hat{y}(\vbx)]} &= \left\{\begin{array}{cl}
(\sigma_\varepsilon^2+ \sigma_{\delta y^*}^2) \frac{\alpha_f}{(1-\alpha_f)} & \qif N_f < M\\
\sigma_\beta^2\sigma_X^2 \frac{(\alpha_f-1)}{\alpha_f^2} +   (\sigma_\varepsilon^2+ \sigma_{\delta y^*}^2)  \frac{1}{(\alpha_f-1)}  & \qif N_f > M.
\end{array} \right.
\end{aligned}
\end{equation}

\subsection{Random Nonlinear Features Model (Two-layer Nonlinear Neural Network)}

In the random nonlinear features model, the student model takes the form
\begin{equation}
\vbz(\vbx) = \frac{1}{\expval*{\varphi'}}\frac{\sigma_W\sigma_X}{\sqrt{N_p}}\varphi\qty(\frac{\sqrt{N_p}}{\sigma_W\sigma_X}W^T\vbx),
\end{equation}
where the elements of the random transformation matrix $W$ are identically and independently distributed, drawn from a normal distribution with zero mean and variance $\sigma_W^2/N_p$,
\begin{equation}
\E\qty[W_{jJ}] = 0 \qqc \Cov\qty[W_{jJ}, W_{kK}] = \frac{\sigma_W^2}{N_p} \delta_{jk}\delta_{JK}.
\end{equation}
We also assume that the elements of $W$ are statistically independent of the ground truth parameters $\vec{\beta}$, the label noise $\vec{\varepsilon}$, the features $X$, etc.

For this model, we again decompose the equation for the gradient in Eq.~\eqref{eq:SIgen_bias} for the above set of hidden features. 
To perform the cavity method, our aim is to construct a set of equations that are linear in the random matrices $W$ and $X$. This results in four different sets of linear equations,
\begin{equation}
\begin{aligned}
\lambda \hat{w}_J &=  \sqrt{M}\alpha_p^{-\frac{1}{2}} \mu_z\expval{\Delta y} + \sum_k\hat{u}_k W_{kJ} + \sum_b \Delta y_b \delta z_{\mathrm{NL}, J}(\vbx_b) + \eta_J\\
\hat{u}_j &= \sum_b \Delta y_b X_{bj} + \psi_j\\
\Delta y_a &=- \sqrt{N_p}\mu_z \expval{\hat{w}} +  \sum_k\Delta \beta_k X_{ak} - \sum_K\hat{w}_K\delta z_{\mathrm{NL}, K}(\vbx_a) + \delta y^*_{\mathrm{NL}}(\vbx_a) + \varepsilon_a   + \xi_a\\
\Delta \beta_j &= \beta_j - \sum_K \hat{w}_K W_{jK} + \zeta_j,
\end{aligned}
\end{equation}
where we have decomposed the training labels according to Eq.~\eqref{eq:SIydecomp} and the hidden features according to Eq.~\eqref{eq:SIzdecomp}.
We have also added a different auxiliary field to each equation and have defined the means of the residual label errors and fit parameter, respectively, as
\begin{equation}
\expval*{\Delta y} = \frac{1}{M}\sum_b \Delta y_b\qqc
\expval*{\hat{w}} = \frac{1}{N_p}\sum_K \hat{w}_K .
\end{equation}
Furthermore, we have included an additional set of variables $\hbu = X^T\Delta \vby$,
which we will also have to solve for to obtain closed form solutions.

\subsubsection{Cavity Expansion}

Next, we add an additional variable of each type, resulting in a total of $M+1$ data points, $N_f+1$ input features and $N_p+1$ fit parameters/hidden features. 
Each additional variable is represented using an index value of $0$, written as $\hat{w}_0$, $\hat{u}_0$, $\Delta y_0$, and $\Delta \beta_0$.
After including these new unknown quantities, the four equations become
\begin{equation}
\begin{aligned}
\lambda \hat{w}_J &= \sqrt{M}\alpha_p^{-\frac{1}{2}} \expval{\Delta y} + \sum_k\hat{u}_k W_{kJ} + \sum_b \Delta y_b \delta z_{\mathrm{NL}, J}(\vbx_b) + \eta_J + \hat{u}_0 W_{0J} +  \Delta y_0 \delta z_{\mathrm{NL}, J}(\vbx_0)\\
\hat{u}_j &= \sum_b \Delta y_b X_{bj} + \psi_j +\Delta y_0 X_{0j}  \\
\Delta y_a &=- \sqrt{N_p}\mu_z \expval{\hat{w}} +  \sum_k\Delta \beta_k X_{ak} - \sum_K\hat{w}_K\delta z_{\mathrm{NL}, K}(\vbx_a) + \delta y^*_{\mathrm{NL}}(\vbx_a) + \varepsilon_a   + \xi_a + \Delta \beta_0 X_{a0} - \hat{w}_0\delta z_{\mathrm{NL}, 0}(\vbx_a) \\
\Delta \beta_j &= \beta_j - \sum_K \hat{w}_K W_{jK} + \zeta_j -  \hat{w}_0 W_{j0},\label{eq:rnlfm_plus0}
\end{aligned}
\end{equation}
with each new variable described by a new equation,
\begin{equation}
\begin{aligned}
\lambda \hat{w}_0 &= \sqrt{M}\alpha_p^{-\frac{1}{2}}  \expval{\Delta y} + \sum_k\hat{u}_k W_{k0} + \sum_b \Delta y_b \delta z_{\mathrm{NL}, 0}(\vbx_b) + \eta_J + \hat{u}_0 W_{00} +  \Delta y_0 \delta z_{\mathrm{NL}, 0}(\vbx_0)\\
\hat{u}_0 &= \sum_b \Delta y_b X_{b0} + \psi_0 +\Delta y_0 X_{00}  \\
\Delta y_0 &=-  \sqrt{N_p}\mu_z \expval{\hat{w}} +  \sum_k\Delta \beta_k X_{0k} - \sum_K\hat{w}_K\delta z_{\mathrm{NL}, K}(\vbx_0) + \delta y^*_{\mathrm{NL}}(\vbx_0) + \varepsilon_0   + \xi_0 + \Delta \beta_0 X_{00} - \hat{w}_0\delta z_{\mathrm{NL}, 0}(\vbx_0) \\
\Delta \beta_0 &= \beta_0 - \sum_K \hat{w}_K W_{0K} + \zeta_0 -  \hat{w}_0 W_{00}.\label{eq:rnlfm_zeroeq}
\end{aligned}
\end{equation}

Now we take the thermodynamic limit in which $M$, $N_f$, and $N_p$ tend towards infinity, but their ratios, $\alpha_f=N_f/M$ and $\alpha_p=N_p/M$, remain fixed.
We interpret the extra terms in Eq.~\eqref{eq:rnlfm_plus0} as small perturbations to the auxiliary fields,
\begin{equation}
\begin{gathered}
\delta \eta_J = \hat{u}_0 W_{0J}  +  \Delta y_0 \delta z_{\mathrm{NL}, J}(\vbx_0)\qqc  \delta \psi_j = \Delta y_0  X_{0 j},\\
\delta \xi_a = \Delta \beta_0 X_{a 0}- \hat{w}_0\delta z_{\mathrm{NL}, 0}(\vbx_a)\qqc  \delta \zeta_j = - \hat{w}_0W_{j0},
\end{gathered}
\end{equation}
allowing us to expand each unknown quantity about its solution in the absence of the $0$-indexed variables, which correspond to the solutions for $M$ data points, $N_f$ input features, and $N_p$ fit parameters,
\begin{equation}
\begin{aligned}
\hat{w}_J &\approx \hat{w}_{J\setminus 0} + \sum_K \nu^{\hat{w}}_{JK}\delta \eta_K + \sum_k\phi^{\hat{w}}_{Jk}\delta \psi_k + \sum_b \chi^{\hat{w}}_{Jb}\delta \xi_b + \sum_k\omega^{\hat{w}}_{Jk}\delta \zeta_k
\\
\hat{u}_j &\approx \hat{u}_{j\setminus 0} + \sum_K \nu^{\hat{u}}_{jK}\delta \eta_K + \sum_k\phi^{\hat{u}}_{jk}\delta \psi_k + \sum_b \chi^{\hat{u}}_{jb}\delta \xi_b + \sum_k\omega^{\hat{u}}_{jk}\delta \zeta_k
\\
\Delta y_a &\approx \Delta y_{a\setminus 0} + \sum_K \nu^{\Delta y}_{aK}\delta \eta_K + \sum_k\phi^{\Delta y}_{ak}\delta \psi_k + \sum_b \chi^{\Delta y}_{ab}\delta \xi_b + \sum_k\omega^{\Delta y}_{ak}\delta \zeta_k
\\
\Delta \beta_j&\approx\Delta \beta_{j\setminus 0} + \sum_K \nu^{\Delta \beta}_{jK}\delta \eta_K + \sum_k\phi^{\Delta \beta}_{jk}\delta \psi_k + \sum_b \chi^{\Delta \beta}_{jb}\delta \xi_b + \sum_k\omega^{\Delta \beta}_{jk}\delta \zeta_k.\label{eq:rnlfm_cavexp}
\end{aligned}
\end{equation}
We define each of the susceptibility matrices as a derivative of a variable with respect to an auxiliary fields,
\begin{equation}
\begin{alignedat}{4}
\nu^{\hat{w}}_{JK} &= \pdv{\hat{w}_J}{\eta_K}\qqc & \phi^{\hat{w}}_{Jk} &= \pdv{\hat{w}_J}{\psi_k}\qqc& \chi^{\hat{w}}_{Jb} &= \pdv{\hat{w}_J}{\xi_b}\qqc & \omega^{\hat{w}}_{Jk} &= \pdv{\hat{w}_J}{\zeta_k},\\
\nu^{\hat{u}}_{jK} &= \pdv{\hat{u}_j}{\eta_K}\qqc & \phi^{\hat{u}}_{jk} &= \pdv{\hat{u}_j}{\psi_k}\qqc& \chi^{\hat{u}}_{jb} &= \pdv{\hat{u}_j}{\xi_b}\qqc & \omega^{\hat{u}}_{jk} &= \pdv{\hat{u}_j}{\zeta_k},\\
\nu^{\Delta y}_{aK} &= \pdv{\Delta y_a}{\eta_K}\qqc  & \phi^{\Delta y}_{ak} &= \pdv{\Delta y_a}{\psi_k}\qqc & \chi^{\Delta y}_{ab} &= \pdv{\Delta y_a}{\xi_b}\qqc &\omega^{\Delta y}_{ak} &= \pdv{\Delta y_a}{\zeta_k},\\
\nu^{\Delta \beta}_{jK} &= \pdv{\Delta \beta_j}{\eta_K}\qqc & \phi^{\Delta\beta}_{jk} &= \pdv{\Delta \beta_j}{\psi_k}\qqc & \chi^{\Delta \beta}_{jb} &= \pdv{\Delta \beta_j}{\xi_b}\qqc & \omega^{\Delta\beta}_{jk} &= \pdv{\Delta\beta_j}{\zeta_k}.
\end{alignedat}
\end{equation}
Next, we substitute the expansions in Eq.~\eqref{eq:rnlfm_cavexp} into the $0$-indexed equations in Eq.~\eqref{eq:rnlfm_zeroeq}.
We then aim to approximate each of the resulting sums in these expanded equations.

\subsubsection{Central Limit Approximations}

We approximate each of the sums containing one of the unperturbed quantities, $\hat{w}_{J\setminus 0}$, $\hat{u}_{j\setminus 0}$, $ \Delta y_{a\setminus 0}$, or $\Delta \beta_{j\setminus 0}$, using the central limit theorem.
Because the unperturbed quantities in each of these sums are statistically independent of all elements of both $X$ and $W$ with a $0$-valued index, we are able to apply the identity in Eq.~\eqref{eq:clt_vec} to find
\begin{equation}
\begin{alignedat}{2}
\sum_k  \hat{u}_{k\setminus 0} W_{k0} + \sum_b\Delta y_{b\setminus 0} \delta z_{\mathrm{NL}, 0}(\vbx_b) &\approx \sigma_{\hat{w}}z_{\hat{w}} \qqc& \sigma_{\hat{w}}^2 &= \sigma_W^2 \frac{\alpha_f}{\alpha_p}\expval*{\hat{u}^2} + \sigma_{\delta z}^2\alpha_p^{-1}\expval*{\Delta y^2}  \\
\sum_b \Delta y_{b\setminus 0} X_{b 0}  &\approx \sigma_{ \hat{u} }z_{ \hat{u}} \qqc&
\sigma_{ \hat{u}}^2 &= \sigma_X^2\alpha_f^{-1}\expval*{\Delta y^2}  \\
\sum_k \Delta \beta_{k\setminus 0} X_{0 k} - \sum_K \hat{w}_{K\setminus 0} \delta z_{\mathrm{NL}, K}(\vbx_0)  &\approx \sigma_{\Delta y}z_{\Delta y} \qqc&
\sigma_{\Delta y}^2 &= \sigma_X^2 \expval*{\Delta \beta^2} +  \sigma_{\delta z}^2  \expval*{\hat{w}^2}  \\
\sum_K \hat{w}_{K\setminus 0}  W_{0 K} &\approx \sigma_{\Delta \beta}z_{\Delta \beta} \qqc&
\sigma_{\Delta \beta}^2 &= \sigma_W^2  \expval*{\hat{w}^2},
\end{alignedat}
\end{equation}
where $z_{\hat{w}}$, $z_{\hat{u}}$, $z_{\Delta y}$, and $z_{\Delta \beta}$ are all independent random variables with zero mean and unit variance.
We also define the following averages:
\begin{equation}
 \expval*{\hat{w}^2}  =\frac{1}{N_p}\sum_K \hat{w}_{K\setminus 0}^2\qqc \expval*{\hat{u}^2} = \frac{1}{N_f}\sum_k\hat{u}_{k\setminus 0}^2\qqc \expval*{\Delta y^2} = \frac{1}{M}\sum_b  \Delta y_{b\setminus 0}^2\qqc \expval*{\Delta\beta^2} = \frac{1}{N_f}\sum_k \Delta \beta_{k\setminus 0}^2.
\end{equation}

Next, we approximate each of the sums containing a square susceptibility matrix.
Using the fact that all of the susceptibility matrices  are statistically independent of all elements of both $X$, $W$, and $\delta \vbz_{\mathrm{NL}}(\vbx)$ with a $0$-valued index, we apply the identity in Eq.~\eqref{eq:clt_square} to find
\begin{equation}
\begin{alignedat}{2}
\sum_{jk}\omega^{\hat{u}}_{jk}W_{j0}W_{k0} &\approx \sigma_W^2 \frac{\alpha_f}{\alpha_p}\omega \qqc & \omega &= \frac{1}{N_f}\sum_k \omega^{\hat{u}}_{kk}\\
\sum_{ab}\chi^{\Delta y}_{ab}X_{a0}X_{b0} &\approx \sigma_X^2 \alpha_f^{-1}\chi \qqc & \chi &= \frac{1}{M}\sum_b\chi^{\Delta y}_{bb}\\
\sum_{jk}\phi^{\Delta \beta}_{jk}X_{0j}X_{0k} &\approx \sigma_X^2 \phi \qqc & \phi &= \frac{1}{N_f}\sum_k \phi^{\Delta \beta}_{kk}\\
\sum_{JK}\nu^{\hat{w}}_{JK}W_{0J}W_{0K} &\approx \sigma_W^2 \nu \qqc & \nu &=\frac{1}{N_p}\sum_K\nu^{\hat{w}}_{KK}.
\end{alignedat}
\end{equation}
Similarly, we approximate the two additional sums containing the nonlinear components of hidden features,
\begin{equation}
\begin{aligned}
\sum_{JK}\nu^{\hat{w}}_{JK} \delta z_{\mathrm{NL}, J}(\vbx_0) \delta z_{\mathrm{NL}, K}(\vbx_0) &\approx \sigma_{\delta z}^2\nu\\
\sum_{ab}\chi^{\Delta y}_{ab} \delta z_{\mathrm{NL}, 0}(\vbx_a) \delta z_{\mathrm{NL}, 0}(\vbx_b) &\approx \sigma_{\delta z}^2\alpha_p^{-1}\chi,
\end{aligned}
\end{equation}
with $\sigma_{\delta z}$ defined in Eq.~\eqref{eq:dznl}.

Finally, each of the remaining sums contains a rectangular susceptibility matrix, so according to the identity in Eq.~\eqref{eq:clt_rect}, is approximately zero in the thermodynamic limit.

\subsubsection{Self-consistency Equations}

Next, we substitute the expansions in Eq.~\eqref{eq:rnlfm_cavexp} into  Eq.~\eqref{eq:rnlfm_zeroeq} and apply the approximations from the previous sections,
resulting in a set of self-consistent equations for $\hat{w}_0$, $\hat{u}_0$, $\Delta y_0$, and $\Delta \beta_0$,
\begin{equation}
\begin{aligned}
\lambda \hat{w}_0 &\approx  \sqrt{M}\alpha_p^{-\frac{1}{2}} \mu_z \expval*{y} + \sigma_{\hat{w}}z_{\hat{w}} -\hat{w}_0\qty(\sigma_W^2 \frac{\alpha_f}{\alpha_p}\omega + \sigma_{\delta z}^2 \alpha_p^{-1}\chi) + \eta_0 \\
\hat{u}_0 &\approx \sigma_{ \hat{u} }z_{ \hat{u}} +  \Delta \beta_0\sigma_X^2 \alpha_f^{-1}\chi + \psi_0 \\
\Delta y_0 &\approx  -  \sqrt{N_p}\mu_z\expval{\hat{w}}  + \sigma_{\Delta y}z_{\Delta y} + \Delta y_0\qty(\sigma_X^2 \phi -\sigma_{\delta z}^2\nu) + \delta y^*_{\mathrm{NL}}(\vbx_0) + \varepsilon_0 + \xi_0 \\
\Delta \beta_0 &\approx \beta_0 -  \sigma_{\Delta\beta}z_{\Delta\beta} - \hat{u}_0\sigma_W^2 \nu  +\zeta_0.
\end{aligned}
\end{equation}
We have also made use of the fact that the terms including $X_{00}$ or $W_{00}$ are infinitesimally small in the thermodynamic limit with zero mean and variances of $\order{1/N_f}$ and $\order{1/N_p}$, respectively.
Solving these equations for the $0$-indexed variables, we find
\begin{equation}
\begin{aligned}
\hat{w}_0  &=  \frac{\sqrt{M}\alpha_p^{-\frac{1}{2}} \mu_z\expval*{y}  + \sigma_{\hat{w}}z_{\hat{w}}  + \eta_0}{\lambda + \sigma_W^2 \frac{\alpha_f}{\alpha_p}\omega + \sigma_{\delta z}^2 \alpha_p^{-1}\chi}\\
\hat{u}_0 &= \frac{\sigma_{\hat{u} }z_{\hat{u}} + \psi_0 + \sigma_X^2 \alpha_f^{-1}\chi\qty(\beta_0 - \sigma_{\Delta \beta}z_{\Delta \beta}   + \zeta_0)}{1 + \sigma_W^2\sigma_X^2\alpha_f^{-1}\chi\nu} \\
\Delta y_0 &= \frac{ - \sqrt{N_p}\mu_z \expval{\hat{w}}  +  \sigma_{\Delta y}z_{\Delta y}+ \delta y^*_{\mathrm{NL}}(\vbx_0) + \varepsilon_0 + \xi_0 }{1-\sigma_X^2 \phi + \sigma_{\delta z}^2\nu} \\
\Delta \beta_0 &= \frac{\beta_0 - \sigma_{\Delta \beta}z_{\Delta \beta}   +\zeta_0 -\sigma_W^2\nu^2\qty(\sigma_{\hat{u} }z_{\hat{u}} + \psi_0)}{1 + \sigma_W^2\sigma_X^2\alpha_f^{-1}\chi\nu}.\label{eq:rnlfm_zeroselfcon}
\end{aligned}
\end{equation}

We then derive a set of self-consistent equations for the scalar susceptibilities by taking appropriate derivatives of these variables with respect to the auxiliary fields,
\begin{equation}
\begin{aligned}
\nu &= \frac{1}{N_p}\sum_K\nu^{\hat{w}}_{KK} \approx \E\qty[\nu^{\hat{w}}_{00}] = \E\qty[\pdv{\hat{w}_0}{\eta_0}] = \frac{1}{\lambda + \sigma_W^2 \frac{\alpha_f}{\alpha_p}\omega+ \sigma_{\delta z}^2 \alpha_p^{-1}\chi}\\
\omega &= \frac{1}{N_f}\sum_k \omega^{\hat{u}}_{kk} \approx \E\qty[\omega^{\hat{u}}_{00}] = \E\qty[\pdv{\hat{u}_0}{\zeta_0}] = \frac{\sigma_X^2\alpha_f^{-1}\chi}{1+\sigma_W^2\sigma_X^2\alpha_f^{-1}\chi\nu}\\
\chi &=\frac{1}{M}\sum_b\chi^{\Delta y}_{bb}\approx  \E\qty[\chi^{\Delta y}_{00}] = \E\qty[\pdv{\Delta y_0}{\xi_0}] = \frac{1}{1-\sigma_X^2\phi+ \sigma_{\delta z}^2\nu}\\
\phi &= \frac{1}{N_f}\sum_k \phi^{\Delta \beta}_{kk} \approx  \E\qty[\phi^{\Delta \beta}_{00}] = \E\qty[\pdv{\Delta\beta_0}{\psi_0}] = -\frac{\sigma_W^2\nu}{1+\sigma_W^2\sigma_X^2\alpha_f^{-1}\chi\nu}.
\end{aligned}
\end{equation}
For convenience, we also introduce a fifth scalar susceptibility,
\begin{equation}
\kappa = \frac{1}{N_f}\sum_k \phi^{\hat{u}}_{kk} = \frac{1}{N_f}\sum_k \omega^{\Delta\beta}_{kk}\approx \E\qty[\pdv{\hat{u}_0}{\psi_0}]  = \E\qty[\pdv{\Delta\beta_0}{\zeta_0}] = \frac{1}{1+\sigma_W^2\sigma_X^2\alpha_f^{-1}\chi\nu} .\label{eq:rnlfm_susceptselfconkappa}
\end{equation}
Using this formula for $\kappa$, we re-express the four other susceptibilities as
\begin{equation}
\begin{aligned}
\omega &= \sigma_X^2\alpha_f^{-1}\chi\kappa \\
\phi &=  -\sigma_W^2\nu\kappa \\
\nu &=\frac{1}{\lambda + \sigma_W^2\sigma_X^2  \alpha_p^{-1}\chi\qty(\kappa + \Delta \varphi)}\\
\chi &= \frac{1}{1+ \sigma_W^2\sigma_X^2\nu\qty(\kappa + \Delta \varphi)}.\label{eq:rnlfm_susceptselfcon}
\end{aligned}
\end{equation}

Next, we find self-consistent equations for the averages of the fit parameter and residual label errors,
\begin{equation}
\begin{aligned}
\expval*{\hat{w}} &= \frac{1}{N_p}\sum_K \hat{w}_K \approx \E\qty[\hat{w}_0 ] = \nu\sqrt{M}\alpha_p^{-\frac{1}{2}} \mu_z \expval*{\Delta y}\\
\expval*{\Delta y} &= \frac{1}{M}\sum_ b \Delta y_b \approx \E\qty[\Delta y_0] = -\chi \sqrt{N_p}\mu_z \expval*{\hat{w}},
\end{aligned}
\end{equation}
where we have set the auxiliary fields to zero.
Solving these equations, it is clear that both averages are zero,
\begin{equation}
\expval*{\hat{w}} = 0 \qqc \expval*{\Delta y}  = 0.
\end{equation}

Finally, we square and average each of Eq.~\eqref{eq:rnlfm_zeroselfcon} to find self-consistent equations for the four ensemble-averaged squared quantities (again setting the auxiliary fields to zero),
\begin{equation}
\begin{aligned}
\expval*{\hat{w}^2} &= \frac{1}{N_p}\sum_K \hat{w}_{K\setminus 0}^2 \approx \E\qty[\hat{w}_0^2] = \nu^2\qty(\sigma_W^2 \frac{\alpha_f}{\alpha_p}\expval*{\hat{u}^2} + \sigma_{\delta z}^2\alpha_p^{-1}\expval*{\Delta y^2}) \\
\expval*{\hat{u}^2} &= \frac{1}{N_f}\sum_k \hat{u}_{k\setminus 0}^2 \approx\E\qty[\hat{u}_0^2] =  \kappa^2\sigma_X^2\alpha_f^{-1}\expval*{\Delta y^2}  + \omega^2\qty(\sigma_\beta^2 + \sigma_W^2\expval*{\hat{w}^2})\\
\expval*{\Delta y^2} &= \frac{1}{M}\sum_b \Delta y_{b\setminus 0}^2 \approx\E\qty[\Delta y_0^2] = \chi^2 \qty(\sigma_X^2\expval*{\Delta\beta^2} + \sigma_{\delta z}^2\expval*{\hat{w}^2} + \sigma_{\delta y^*}^2 + \sigma_\varepsilon^2) \\
\expval*{\Delta\beta^2} &=\frac{1}{N_f}\sum_k \Delta \beta_{k\setminus 0}^2 \approx \E\qty[\Delta \beta_0^2] = \kappa^2\qty( \sigma_\beta^2 + \sigma_W^2\expval*{\hat{w}^2}) + \phi^2\sigma_X^2\alpha_f^{-1} \expval*{\Delta y^2}.\label{eq:rnlfm_squareselfcon}
\end{aligned}
\end{equation}

\subsubsection{Solution with Finite Regularization ($\lambda \sim \order{1}$)}

We start by solving the equations for $\chi$ and $\nu$ in Eq.~\eqref{eq:rnlfm_susceptselfcon} for $\kappa$ and setting them equal,
\begin{equation}
\kappa = \frac{1-\chi- \sigma_W^2\sigma_X^2\Delta \varphi \nu}{\sigma_W^2\sigma_X^2\chi\nu} = \frac{\alpha_p(1-\lambda\nu)-\sigma_W^2\sigma_X^2\Delta \varphi \nu}{ \sigma_W^2\sigma_X^2\chi\nu},
\end{equation}
giving us a relation between $\nu$ and $\chi$,
\begin{equation}
\nu = \frac{\chi+\alpha_p-1}{\lambda\alpha_p}.\label{eq:rnlfm_chinu}
\end{equation}
Substituting $\kappa$ from Eq.~\eqref{eq:rnlfm_susceptselfconkappa} into $\chi$ from Eq.~\eqref{eq:rnlfm_susceptselfcon}, inserting the expression for $\nu$ we just found, and then multiplying out the denominators, 
we find a quartic equation for $\chi$,
\begin{equation}
\begin{aligned}
0 &= \Delta \varphi \chi^4 +  \qty[2 \Delta\varphi (\alpha_p-1) + \alpha_p\bar{\lambda}]\chi^3 +  \qty[ \Delta\varphi (\alpha_p-1)^2 + \qty(\qty(1+ \Delta\varphi )\alpha_f + \alpha_p - 2)\alpha_p\bar{\lambda}]\chi^2\\
&\qquad +\qty[ \qty(\qty(1+ \Delta\varphi )\alpha_f-1)\qty(\alpha_p-1) +\alpha_f\alpha_p \bar{\lambda} ] \alpha_p\bar{\lambda}\chi  -\alpha_f\alpha_p^2\bar{\lambda}^2, \label{eq:rnlfm_poly}
\end{aligned}
\end{equation}
where we have defined the dminesionless regularization parameter
\begin{align}
\bar{\lambda} &= \frac{\lambda}{ \sigma_W^2\sigma_X^2}.
\end{align}
Solving the quartic equation and solving for the remaining susceptibilities, we find exact solutions in the thermodynamic limit by solving Eq.~\eqref{eq:rnlfm_squareselfcon},
\begin{equation}
\mqty(\expval*{\hat{w}^2}\\  \expval*{\hat{u}^2} \\ \expval*{\Delta y^2} \\ \expval*{\Delta \beta^2}) =
\mqty(1 & -\sigma_W^2  \frac{\alpha_f}{\alpha_p}\nu^2 &  -\sigma_{\delta z}^2 \alpha_p^{-1}\nu^2 & 0\\
- \sigma_W^2 \omega^2   & 1  & -\sigma_X^2\alpha_f^{-1}\kappa^2 & 0\\
- \sigma_{\delta z}^2 \chi^2  & 0 & 1 &  -\sigma_X^2\chi^2 \\
-  \sigma_W^2\kappa^2 & 0 &  -   \sigma_X^2 \alpha_f^{-1}\phi^2 & 1)^{-1}
\mqty(0 \\    \sigma_\beta^2 \omega^2\\ (\sigma_\varepsilon^2 + \sigma_{\delta y^*}^2)\chi^2 \\  \sigma_\beta^2\kappa^2). \label{eq:rnlfm_mateq}
\end{equation}

\subsubsection{Solutions in Ridge-less Limit ($\lambda\rightarrow 0$)}

In the ridge-less limit ($\lambda\rightarrow 0$), we make the ansatz that $\chi$ is $\order{1}$ in small $\bar{\lambda}$,
\begin{equation}
\chi \approx \chi_0 + \bar{\lambda} \chi_1.
\end{equation}
Using this approximation, Eq.~\eqref{eq:rnlfm_poly} gives us the following equation at $\order{1}$:
\begin{equation}
\begin{aligned}
0 &=  \Delta\varphi\chi_0^4 + 2\Delta\varphi(\alpha_p-1)\chi_0^3 + \Delta \varphi(\alpha_p-1)^2 \chi_0^2.
\end{aligned}
\end{equation}
This equation has two solutions for $\chi_0$,
\begin{equation}
\chi_0^{(1)} = 1-\alpha_p \qqc \chi_0^{(2)} = 0,
\end{equation}
labeled by superscript $(1)$ and $(2)$.

At $\order{\bar{\lambda}}$, we find the resulting equation to be uninformative after inserting either of the solutions for $\chi_0$.
However, the $\order{\bar{\lambda}^2}$ equation does provide unique solutions,
\begin{equation}
\begin{aligned}
0 &= \Delta\varphi\qty(4\chi_0^3\chi_2 + 6\chi_0^2\chi_1^2) +  2\Delta\varphi(\alpha_p-1)(3\chi_0^2\chi_2 + 3\chi_0\chi_1^2) + 3\alpha_p \chi_0^2\chi_1 + \Delta\varphi(\alpha_p-1)^2(2\chi_0\chi_2 + \chi_1^2)\\
&\qquad + 2\qty(\qty(1+\Delta\varphi)\alpha_f + \alpha_p - 2)\alpha_p\chi_0\chi_1 +  \qty(\qty(1+\Delta\varphi)\alpha_f-1)\qty(\alpha_p-1)\alpha_p\chi_1 +\alpha_f\alpha_p^2\chi_0  -\alpha_f\alpha_p^2.
\end{aligned}
\end{equation}
Inserting $\chi_0^{(1)}$, this equation becomes
\begin{equation}
0 = \Delta\varphi(1-\alpha_p)^2\chi_1^2 - \qty[\alpha_p - \qty(1+\Delta\varphi)\alpha_f](1-\alpha_p)\alpha_p\chi_1  - \alpha_f\alpha_p^3
\end{equation}
with a pair of solutions,
\begin{equation}
\chi_1^{(1)} = \frac{1}{ 2\Delta\varphi\frac{(1-\alpha_p)}{\alpha_p}}\qty[\alpha_p - \qty(1+\Delta\varphi)\alpha_f \pm \sqrt{\qty[\alpha_p - \qty(1+\Delta\varphi)\alpha_f]^2 + 4\Delta\varphi\alpha_f\alpha_p}].
\end{equation}
Similarly, inserting the second solution $\chi_0^{(2)}$, we find
\begin{equation}
0 = \Delta\varphi(\alpha_p-1)^2\chi_1^2 -  \qty[1-\qty(1+\Delta\varphi)\alpha_f]\qty(\alpha_p-1)\alpha_p\chi_1  - \alpha_f\alpha_p^2
\end{equation}
with a second pair of solutions,
\begin{equation}
\chi_1^{(2)} = \frac{1}{ 2\Delta\varphi\frac{(\alpha_p-1)}{\alpha_p}}\qty[1 - \qty(1+\Delta\varphi)\alpha_f \pm \sqrt{\qty[1 - \qty(1+\Delta\varphi)\alpha_f]^2 + 4\Delta\varphi\alpha_f}].
\end{equation}
We note that the two solutions for $\chi_1$ are qualitatively similar to the exact solution for $\chi$ in the model of linear regression, Eq.~\eqref{eq:linreg_nuexact}.
The implication is that the nonlinear nature of the activation function implicitly serves as a type of regularization via the quantity $ \Delta\varphi$ which evaluates to zero in the linear limit.

Next, we solve for the solutions to $\nu$. 
First, we make the ansatz
\begin{equation}
\nu \approx \frac{1}{\bar{\lambda}}\nu_{-1} + \nu_0.
\end{equation}
Using Eq.~\eqref{eq:rnlfm_chinu} and inserting the first solution for $\chi$, we find
\begin{equation}
\nu_{-1}^{(1)} = 0
\end{equation}
with the next order term
\begin{equation}
\nu_0^{(1)} = \frac{1}{\sigma_W^2\sigma_X^2}  \frac{1}{ 2\Delta\varphi (1-\alpha_p) }\qty[\alpha_p - \qty(1+\Delta\varphi)\alpha_f \pm \sqrt{\qty[\alpha_p - \qty(1+\Delta\varphi)\alpha_f]^2 + 4\Delta\varphi\alpha_f\alpha_p}].
\end{equation}
Similarly, the second solution for $\chi$ gives us
\begin{equation}
\nu_{-1}^{(2)} = \frac{1}{\sigma_W^2\sigma_X^2} \frac{(\alpha_p-1)}{\alpha_p}.
\end{equation}
For completion, we also find
\begin{equation}
\nu_0^{(2)} = \frac{\chi_1^{(2)}}{\alpha_p} = \frac{1}{ 2\Delta\varphi(\alpha_p-1)}\qty[1 - \qty(1+\Delta\varphi)\alpha_f \pm \sqrt{\qty[1 - \qty(1+\Delta\varphi)\alpha_f]^2 + 4\Delta\varphi\alpha_f}].
\end{equation}

None of the remaining scalar susceptibilities have simple forms. 
Therefore, we use their representations in terms of $\nu$ and $\chi$.
Similarly, the solutions for $\expval*{\hat{w}^2}$, $\expval*{\Delta y^2}$, $\expval*{\hat{u}^2}$, and $\expval*{\Delta \beta^2}$ do not simplify significantly,
but their limiting scaling behavior in terms of $\lambda$ can still be determined.
Using the fact that each of these quantities must be positive, it is straightforward to see that only two of the four solutions apply, depending on whether $\alpha_p > 1$ or $\alpha_p < 1$.
The resulting solutions for $\chi$ and $\nu$ are then
\begin{equation}
\begin{aligned}
\chi &= \left\{
\begin{array}{cl}
1-\alpha_p &\qif N_p < M\\
 \frac{\lambda}{2\Delta \varphi\sigma_X^2 \sigma_W^2 }\frac{\alpha_p}{ (\alpha_p-1)} \qty[ 1- \qty(1+ \Delta \varphi)\alpha_f + \sqrt{\qty[1-\qty(1+ \Delta \varphi)\alpha_f]^2   + 4 \Delta\varphi  \alpha_f }] &\qif N_p > M
\end{array}
\right.\\
\nu &= \left\{
\begin{array}{cl}
\frac{1}{2 \Delta \varphi  \sigma_X^2 \sigma_W^2 }\frac{1}{(1-\alpha_p)}\qty[\alpha_p -  \qty(1+\Delta \varphi) \alpha_f + \sqrt{  \qty[\alpha_p -   \qty(1+\Delta \varphi)\alpha_f]^2   + 4 \Delta \varphi\alpha_f\alpha_p }]& \qif N_p < M \\
\frac{1}{\lambda} \frac{(\alpha_p-1)}{\alpha_p} + \frac{1}{2\Delta\varphi  \sigma_X^2 \sigma_W^2 }\frac{1}{ (\alpha_p-1)} \qty[ 1- \qty(1+\Delta \varphi)\alpha_f + \sqrt{\qty[1-\qty(1+ \Delta \varphi)\alpha_f]^2   + 4  \Delta\varphi  \alpha_f }]  & \qif N_p > M
\end{array}
\right.
\end{aligned}
\end{equation}
In addition, $\expval*{\Delta y^2}$ is $\order{1}$ in small $\lambda$ when $\alpha_p < 1$ and $\order{\lambda^2}$ when $\alpha_p > 1$.
The training and test error can be determined by substituting these susceptibilities into the equations for $\expval*{\hat{w}^2}$, $\expval*{\Delta y^2}$, $\expval*{\hat{u}^2}$, and $\expval*{\Delta \beta^2}$ in Eq.~\eqref{eq:rnlfm_mateq} and then using the general solutions in Eqs.~\eqref{eq:SIgen_train} and \eqref{eq:SIgen_test}.

\subsubsection{Bias-Variance Decomposition}

To derive expressions for the bias and variance, according the general solutions in Eqs.~\eqref{eq:SIgen_bias} and \eqref{eq:SIgen_var}, we need to calculate the covariance of the residual parameter errors $\expval*{\Delta\beta_1\Delta\beta_2}$, as well as the covariance of the fit parameters $\expval*{\hat{w}_1\hat{w}_2}$.
As a reminder, the subscripts $1$ and $2$ refer to parameters resulting from fitting training sets $\mathcal{D}_1$ and $\mathcal{D}_2$ drawn independently from the same data distribution.
We apply the self-consistent equations for the $0$-indexed quantities, Eq.~\eqref{eq:rnlfm_zeroselfcon}, to the two data sets, giving us
\begin{equation}
\begin{aligned}
\hat{w}_{1,0}  &= \nu\sigma_{\hat{w}}z_{\hat{w}_1}\\
\hat{u}_{1,0} &= \kappa \sigma_{\hat{u}}z_{\hat{u}_1} + \omega\qty(\beta_0 - \sigma_{\Delta \beta}z_{\Delta \beta_1} )\\
\Delta y_{1,0} &= \chi\qty(\sigma_{\Delta y}z_{\Delta y_1} + \delta y^*_{\mathrm{NL}}(\vbx_{1, 0}) + \varepsilon_{1,0})\\
\Delta \beta_{1,0} &= \kappa \qty(\beta_0 - \sigma_{\Delta \beta}z_{\Delta \beta_1}) + \phi  \sigma_{\hat{u}}z_{\hat{u}_1}
\end{aligned}
\end{equation}
and
\begin{equation}
\begin{aligned}
\hat{w}_{2,0}  &= \nu\sigma_{\hat{w}}z_{\hat{w}_2}\\
\hat{u}_{2,0} &= \kappa \sigma_{\hat{u}}z_{\hat{u}_2} + \omega\qty(\beta_0 - \sigma_{\Delta \beta}z_{\Delta \beta_2} )\\
\Delta y_{2,0} &= \chi\qty(\sigma_{\Delta y}z_{\Delta y_2} + \delta y^*_{\mathrm{NL}}(\vbx_{2, 0}) + \varepsilon_{2,0})\\
\Delta \beta_{2,0} &= \kappa \qty(\beta_0 - \sigma_{\Delta \beta}z_{\Delta \beta_2}) + \phi \sigma_{\hat{u}}z_{\hat{u}_2}.
\end{aligned}
\end{equation}
Multiplying these equations and using the self-averaging approximation, we find
\begin{equation}
\begin{aligned}
\expval*{\hat{w}_1\hat{w}_2} &= \frac{1}{N_p}\sum_K \hat{w}_{1, K}\hat{w}_{2, K} \approx \E\qty[\hat{w}_{1, 0}\hat{w}_{2, 0}] = \nu^2\E\qty[\sigma_{\hat{w}}^2z_{\hat{w}_1}z_{\hat{w}_2}] \\
\expval*{\hat{u}_1\hat{u}_2} &= \frac{1}{N_f}\sum_k \hat{u}_{1, k}\hat{u}_{2, k} \approx \E\qty[\hat{u}_{1, 0}\hat{u}_{2, 0}] = \kappa^2\E\qty[\sigma_{\hat{u}}^2z_{\hat{u}_1}z_{\hat{u}_2}] + \omega^2\qty(\sigma_\beta^2 + \E\qty[\sigma_{\Delta \beta}^2z_{\Delta \beta_1}z_{\Delta \beta_2}] ) \\
\expval*{\Delta y_1 \Delta y_2} &= \frac{1}{M}\sum_b \Delta y_{1, b}\Delta y_{2, b} \approx \E\qty[\Delta y_{1, 0}\Delta y_{2, 0}] = \chi^2\E\qty[\sigma_{\Delta y}^2z_{\Delta y_1}z_{\Delta y_2}] \\
\expval*{\Delta\beta_1\Delta\beta_2} &= \frac{1}{N_f}\sum_k \Delta \beta_{1, k}  \Delta \beta_{2, k} \approx \E\qty[ \Delta \beta_{1, 0}  \Delta \beta_{2, 0}] = \kappa^2\qty(\sigma_\beta^2 + \E\qty[\sigma_{\Delta \beta}^2z_{\Delta \beta_1}z_{\Delta \beta_2}]) + \phi^2\E\qty[\sigma_{\hat{u}}^2z_{\hat{u}_1}z_{\hat{u}_2}] .\label{eq:rnlfm_corrselfcon}
\end{aligned}
\end{equation}

Next, we calculate each of the four resulting expectation values of products of random variables.
The average of the product $z_{\hat{w}_1}z_{\hat{w}_2}$ is
\begin{equation}
\begin{aligned}
\E\qty[\sigma_{\hat{w}}^2z_{\hat{w}_1}z_{\hat{w}_2}]
&= \E\qty[\qty(\sum_j \hat{u}_{1, j\setminus 0} W_{j0} + \sum_a \Delta y_{1, a\setminus 0}\delta z_{\mathrm{NL}, 0}(\vbx_{1, a}) )\qty(\sum_k \hat{u}_{2, k\setminus 0} W_{k0} + \sum_b \Delta y_{2, b\setminus 0}\delta z_{\mathrm{NL}, 0}(\vbx_{2, b}) )]\\
&=  \sum_{jk}\E\qty[\hat{u}_{1, j\setminus 0}\hat{u}_{2, k\setminus 0}]\E\qty[W_{j0}W_{k0}] + \sum_{ab} \E\qty[\Delta y_{1, a\setminus 0} \Delta y_{2, b\setminus 0}]\E\qty[\delta z_{\mathrm{NL}, 0}(\vbx_{1, a})]\E\qty[  \delta z_{\mathrm{NL}, 0}(\vbx_{2, b})]\\
&= \frac{\sigma_W^2}{N_p}\sum_k \E\qty[\hat{u}_{1, j\setminus 0}\hat{u}_{2, j\setminus 0}]\\
&\approx \sigma_W^2 \frac{\alpha_f}{\alpha_p} \expval*{\hat{u}_1\hat{u}_2},
\end{aligned}
\end{equation}
while the average of the product $z_{\Delta \beta_1}z_{\Delta \beta_2}$ results in 
\begin{equation}
\begin{aligned}
\E\qty[\sigma_{\Delta \beta}^2z_{\Delta \beta_1}z_{\Delta \beta_2}] &= \E\qty[\sum_{JK}\hat{w}_{1, J\setminus 0}\hat{w}_{2, K\setminus 0}W_{0J}W_{0K}]\\
&=  \sum_{JK}\E\qty[\hat{w}_{1, J\setminus 0}\hat{w}_{2, K\setminus 0}]\E\qty[W_{0J}W_{0K}]\\
&= \frac{\sigma_W^2}{N_p}\sum_K \E\qty[\hat{w}_{1, K\setminus 0}\hat{w}_{2, K\setminus 0}]\\
&\approx \sigma_W^2 \expval*{\hat{w}_1\hat{w}_2}.
\end{aligned}
\end{equation}
We find that the other two products average to zero due to the independence of $X_1$ and $X_2$, giving us
\begin{equation}
\begin{aligned}
\E\qty[\sigma_{\hat{u}}^2z_{\hat{u}_1}z_{\hat{u}_2}] &= \E\qty[\sum_{ab}\Delta y_{1, a\setminus 0}\Delta y_{2, b\setminus 0}X_{1, a0}X_{2, b0}]\\
&=  \sum_{ab}\E\qty[\Delta y_{1, a\setminus 0}\Delta y_{2, b\setminus 0}]\E\qty[X_{1, a0}]\E\qty[X_{2, b0}]\\
&= 0
\end{aligned}
\end{equation}
and
\begin{equation}
\begin{aligned}
\E\qty[\sigma_{\Delta y}^2z_{\Delta y_1}z_{\Delta y_2}] &= \E\qty[\qty(\sum_j \Delta \beta_{1, j\setminus 0} X_{1, 0j} - \sum_J \hat{w}_{1, J\setminus 0}\delta z_{\mathrm{NL}, J}(\vbx_{1, 0}))\qty(\sum_k \Delta \beta_{2, k\setminus 0} X_{2, 0k} - \sum_K \hat{w}_{2, K\setminus 0}\delta z_{\mathrm{NL}, K}(\vbx_{2, 0}))  ]\\
&=  \sum_{jk}\E\qty[\Delta \beta_{1, j\setminus 0}\Delta \beta_{2, k\setminus 0}]\E\qty[X_{1, 0j}]\E\qty[X_{2, 0k}] + \sum_{JK}\E\qty[\hat{w}_{1, J\setminus 0} \hat{w}_{2, K\setminus 0}]\E\qty[\delta z_{\mathrm{NL}, J}(\vbx_{1, 0})]\E\qty[\delta z_{\mathrm{NL}, K}(\vbx_{2, 0})]\\
&= 0.
\end{aligned}
\end{equation}
Substituting these results into Eq.~\eqref{eq:rnlfm_corrselfcon}, we find the self-consistent equations
\begin{equation}
\begin{aligned}
\expval*{\hat{w}_1\hat{w}_2} &= \nu^2\sigma_W^2\frac{\alpha_f}{\alpha_p} \expval*{\hat{u}_1\hat{u}_2}\\
\expval*{\hat{u}_1\hat{u}_2} &= \omega^2\qty(\sigma_\beta^2 + \sigma_W^2 \expval*{\hat{w}_1\hat{w}_2})\\
\expval*{\Delta y_1 \Delta y_2} &= 0\\
\expval*{\Delta\beta_1\Delta\beta_2} &= \kappa^2\qty( \sigma_\beta^2 + \sigma_W^2 \expval*{\hat{w}_1\hat{w}_2}).
\end{aligned}
\end{equation}

Solving these equations exactly in the thermodynamic limit, we find the expressions
\begin{equation}
\begin{aligned}
\expval*{\hat{w}_1\hat{w}_2}  &= \frac{\sigma_\beta^2}{\sigma_W^2} \frac{\sigma_W^4\frac{\alpha_f}{\alpha_p} \omega^2\nu^2}{\qty(1- \sigma_W^4\frac{\alpha_f}{\alpha_p} \omega^2\nu^2)}\\
\expval*{\hat{u}_1\hat{u}_2} &= \sigma_\beta^2 \frac{ \omega^2}{\qty(1- \sigma_W^4\frac{\alpha_f}{\alpha_p} \omega^2\nu^2)}\\
\expval*{\Delta\beta_1\Delta\beta_2} &=\sigma_\beta^2 \frac{\kappa^2}{\qty(1- \sigma_W^4\frac{\alpha_f}{\alpha_p} \omega^2\nu^2)}.
\end{aligned}
\end{equation}

\section{Spectral Densities of Kernel Matrices}

Here, we derive the spectral densities for the kernel matrix $Z^TZ$ for each model.
To do this, we use the technique laid out in Ref.~\onlinecite{Cui2020}.
For any symmetric matrix $A$ of size $N\times N$, the spectral density can be written in the form
\begin{align}
\rho(x) &= \frac{1}{\pi}\lim_{\varepsilon\rightarrow 0^+}\Im \frac{1}{N} \Tr G(x - i\varepsilon),
\end{align}
where 
\begin{equation}
G(z) = \qty[zI_N - A]^{-1}
\end{equation}
is the Green's function. 

In our case, we are interested in the case $A=Z^TZ$.
From the cavity calculations, we observe that the susceptibility  matrix 
\begin{align}
\nu^{\hat{w}}(\lambda) & = \qty[\lambda I_{N_p} + Z^TZ]^{-1}
\end{align}
is related to the Green's function via the relation $G(z) =  - \nu^{\hat{w}}(-z)$.
This allows us to the express the spectral density in terms of $\nu$,
\begin{align}
\rho(x) &= -\frac{1}{\pi}\lim_{\varepsilon\rightarrow 0^+}\Im \nu(-x + i\varepsilon).\label{eq:spectrum}
\end{align}
Therefore, all we will need to do is evaluate Eq.~\eqref{eq:spectrum} using the appropriate function $ \nu(\lambda)$ for each model.

Sometimes, there will be some fraction of eigenvalues at zero. 
While the weight of this contribution can be directly calculated via Eq.~\eqref{eq:spectrum}, sometimes it is easier to instead examine the susceptibility matrix $\chi^{\Delta y}$ in the limit $\lambda \rightarrow 0$, which becomes
\begin{equation}
\chi^{\Delta y} =I_M - Z\qty[\lambda I_{N_p} + Z^TZ]^{-1}Z^T \approx I_M - ZZ^+.
\end{equation}
The matrix $ZZ^+$ is a projector, so its trace is the rank of $Z^TZ$.
The trace of $\chi^{\Delta y}$ is then
\begin{equation}
\begin{aligned}
\chi &= \frac{1}{M}\Tr\chi^{\Delta y} = 1 - \frac{1}{M}\rank(Z^TZ).
\end{aligned}
\end{equation}
The fraction of eigenvalues at zero is then
\begin{align}
f_{\mathrm{zero}} = 1-\frac{1}{N_p}\rank(Z^TZ) =  \frac{\chi + \alpha_p -1}{\alpha_p}.\label{eq:fzero}
\end{align}

\subsection{Linear Regression}

For linear regression, the kernel  is a Wishart matrix of the form $A = X^TX$,
where the elements of the matrix $X$ are independent and identically distributed according to a normal distribution with zero mean, the expected eigenvalue spectrum is the Marchenko-Pastur distribution~\cite{Marchenko1967}. 
To show this, we start the self-consistency equations for the susceptibilities for this model,
\begin{equation}
\chi = \frac{1}{1 + \bar{\nu} } \qqc \bar{\nu} =  \frac{1}{\bar{\lambda} + \alpha_f^{-1}\chi},
\end{equation}
where we have non-dimensionalized $\nu$ and $\lambda$ by defining
\begin{equation}
\bar{\nu} = \sigma_X^2 \nu\qqc\bar{\lambda} = \frac{\lambda}{\sigma_X^2}.
\end{equation}
Plugging $\chi$ into $\bar{\nu}$ and rearranging, we find a quadratic equation for $\bar{\nu}$,
\begin{equation}
\bar{\lambda} \bar{\nu}^2 + \qty[\qty(\alpha_f^{-1}-1) +   \bar{\lambda}]\bar{\nu} - 1 = 0.
\end{equation}
Solving this equation, we find
\begin{equation}
\nu(\lambda) =  \frac{\sigma_X^2\qty(1-\alpha_f^{-1}) -    \lambda \pm \sqrt{D(\lambda)}}{2\sigma_X^2 \lambda},
\end{equation}
where we have defined the discriminant
\begin{equation}
D(\lambda) = \qty[\sigma_X^2\qty(\alpha_f^{-1}-1) +   \lambda]^2 + 4\sigma_X^2 \lambda.
\end{equation}
Next, we substitute the above solution for $\nu$ into Eq.~\eqref{eq:spectrum} and simplify to find to find
\begin{equation}
\rho(x) = \frac{1}{2\sigma_X^2}\qty[\sigma_X^2\qty(1 - \alpha_f^{-1}) \pm \Re\sqrt{D(-x)}] \frac{1}{\pi} \lim_{\epsilon\rightarrow 0^+} \frac{\epsilon}{\qty(x^2+\epsilon^2)} \pm \frac{\Im \sqrt{D(-x)}}{2\pi\sigma_X^2 x}.
\end{equation}
We see that the first term contains the definition of a delta function evaluated at zero,
\begin{equation}
\delta (x) = \frac{1}{\pi} \lim_{\epsilon\rightarrow 0^+} \frac{\epsilon}{\qty(x^2+\epsilon^2)}.
\end{equation}
This allows us to evaluate the the coefficient of this delta function at zero so that the first term in the spectrum becomes
\begin{equation}
\begin{aligned}
 \frac{1}{2\sigma_X^2}\qty[\sigma_X^2\qty(1- \alpha_f^{-1}) \pm \Re\sqrt{D(0)}] \delta(x)  &= \max\qty(0, 1 - \alpha_f^{-1})\delta(x).
 \end{aligned}
\end{equation}
We have chosen the signs of the solutions $(\pm)$ so that the spectral density at zero is always non-negative. 
To simplify the second term in Eq.~\eqref{eq:spectrum}, we need to find the interval over which $D(-x) < 0$.  Solving $D(-x) = 0$,
\begin{equation}
D(-x)  = x^2 - 2x\sigma_X^2\qty(\alpha_f^{-1}+1)x + \sigma_X^4\qty(\alpha_f^{-1}-1)^2 = 0,
\end{equation}
we find the limits of the interval to be
\begin{equation}
x_\pm = \sigma_X^2 \qty(1 \pm \sqrt{\alpha_f^{-1}})^2.
\end{equation}
The second term in the spectrum then becomes
\begin{equation}
\pm\frac{\Im \sqrt{-(x_+-x)(x-x_-)}}{2\pi\sigma_X^2 x} = \frac{\sqrt{(x_+-x)(x-x_-)}}{2\pi\sigma_X^2 x},
\end{equation}
where we have again chosen the plus sign so that the spectrum is always non-negative.

The complete spectrum is then written as
\begin{equation}
\begin{aligned}
\rho(x) &= \max\qty(0, 1 -  \alpha_f^{-1})\delta(x)   + \left\{ 
\begin{array}{cl}
\frac{1}{2\pi\sigma_X^2 x} \sqrt{(x_{\max}-x)(x-x_{\min})}& \qif x \in [x_{\min}, x_{\max}]\\
0 & \qotherwise
\end{array}
\right.
\end{aligned}
\end{equation}
with 
\begin{equation}
x_{\min} = \sigma_X^2 \qty(1 - \sqrt{\alpha_f^{-1}})^2\qqc
x_{\max} = \sigma_X^2 \qty(1 + \sqrt{\alpha_f^{-1}})^2.
\end{equation}
As expected, this is the Marchenko-Pastur distribution.

\subsection{Random Nonlinear Features Model}

Next, we derive the eigenvalue distribution for the kernel of the random nonlinear features model.
Previously this result was derived in Ref.~\onlinecite{Pennington2019}.
To reproduce this analytic result, we start with three of the susceptibilities from the cavity derivation,
\begin{equation}
\begin{gathered}
\bar{\nu} =\frac{1}{\bar{\lambda} +  \alpha_p^{-1}\chi\qty(\kappa + \Delta \varphi)}\qqc
\chi = \frac{1}{1+ \bar{\nu}\qty(\kappa + \Delta \varphi)}\qqc
\kappa = \frac{1}{1+\alpha_f^{-1}\chi\bar{\nu}}.
\end{gathered}
\end{equation}
where we have non-dimensionalized $\nu$ and $\lambda$ by defining
\begin{equation}
\bar{\nu} = \sigma_W^2\sigma_X^2\nu\qqc \bar{\lambda} = \frac{\lambda}{\sigma_W^2\sigma_X^2}.
\end{equation}
Solving each of the equations for $\chi$ and $\bar{\nu}$ for $\kappa$ and then setting them equal we find
\begin{equation}
\kappa = \frac{1-\chi-  \Delta \varphi\bar{\nu}}{ \chi\bar{\nu}} = \frac{\alpha_p(1-\bar{\lambda}\bar{\nu})- \Delta \varphi \bar{\nu}}{  \chi\bar{\nu}}.
\end{equation}
From here, we find the following relation between $\chi$ and $\bar{\nu}$:
\begin{equation}
\chi = \alpha_p\bar{\lambda}\bar{\nu} - \alpha_p + 1.\label{eq:rnlfm_spec_chinu}
\end{equation}
Next, we substitute $\kappa$ into original equation for $\bar{\nu}$, solve for $\chi$, and then substitute this result  into Eq.~\eqref{eq:rnlfm_spec_chinu}.
If we then eliminate any denominators, we find a quartic equation for $\bar{\nu}$,
\begin{equation}
\begin{aligned}
0 &= 
\Delta \varphi (\alpha_p \bar{\lambda}\bar{\nu})^4 + \qty[2\Delta \varphi (1-\alpha_p) + \alpha_p \bar{\lambda}](\alpha_p \bar{\lambda}\bar{\nu})^3\\
&\qquad + \qty[\Delta \varphi(1-\alpha_p)^2 + \alpha_p\qty(\alpha_f(1+\Delta\varphi)-\alpha_p + 1-\alpha_p )\bar{\lambda} ](\alpha_p \bar{\lambda}\bar{\nu})^2\\
&\qquad + \alpha_p\qty[(\alpha_f(1+\Delta\varphi)-\alpha_p)(1-\alpha_p) + \alpha_f\alpha_p \bar{\lambda}]\bar{\lambda}(\alpha_p \bar{\lambda}\bar{\nu}) - \alpha_f\alpha_p^3 \bar{\lambda}^2.
\end{aligned}
\end{equation}

Solving this quartic equation analytically is very involved, so instead we will solve this equation numerically for negative imaginary roots of  $\nu(\lambda)$ with $\lambda = -x$, according to Eq.~\eqref{eq:spectrum}.
However, to find the interval over which the spectrum is positive, we rewrite the equation in general form for $\alpha_p\bar{\lambda}\bar{\nu}$,
\begin{equation}
\begin{aligned}
0 &= 
a_4 (\alpha_p \bar{\lambda}\bar{\nu})^4 + a_3(\alpha_p \bar{\lambda}\bar{\nu})^3 + a_2(\alpha_p \bar{\lambda}\bar{\nu})^2 + a_1(\alpha_p \bar{\lambda}\bar{\nu}) +a_0,
\end{aligned}
\end{equation}
where the coefficients are
\begin{equation}
\begin{aligned}
a_0 &=  - \alpha_f\alpha_p^3 \bar{\lambda}^2\\
a_1 &=\alpha_p\qty[(\alpha_f(1+\Delta\varphi)-\alpha_p)(1-\alpha_p) + \alpha_f\alpha_p \bar{\lambda}]\bar{\lambda}\\
a_2 &=  \Delta \varphi(1-\alpha_p)^2 + \alpha_p\qty(\alpha_f(1+\Delta\varphi)-\alpha_p + 1-\alpha_p )\bar{\lambda}\\
a_3 &=  2\Delta \varphi (1-\alpha_p) + \alpha_p \bar{\lambda}\\
a_4 &= \Delta \varphi.
\end{aligned}
\end{equation}
The discriminant for a quartic equation is expressed in terms of these coefficients as
\begin{equation}
D(z) = R^2 - 4Q^3
\end{equation}
with
\begin{equation}
\begin{aligned}
R &= 2a_2^3 - 9a_1a_2a_3 + 27 a_0 a_3^2  + 27 a_1^2a_4 - 72 a_0a_2a_4\\
Q &=  a_2^2 - 3 a_1  a_3 + 12 a_0a_4.
\end{aligned}
\end{equation}
To find the limiting eigenvalues, we then solve the equation $D(\lambda) = 0$ (with $\lambda = -x$) numerically for the largest and smallest non-negative real roots.

To find the weight of the delta function component at zero, we use Eq.~\eqref{eq:fzero} and the solutions for $\nu$ we found previously this model, giving us
\begin{equation}
\begin{aligned}
f_{\mathrm{zero}} &=  \max\qty(0, 1 - \alpha_p^{-1} ).
\end{aligned}
\end{equation}

\section{Accuracy of Minimum Principal Component}\label{sec:acc_mpc}

In this section, we derive expressions for the predicted labels $\hat{y}$ as a function of projections of the data points along the minimum principal component $\hbh_{\min}\cdot\vbz(\vbx)$ used to assess model accuracy in Figs.~\ref{fig:narrow} and \ref{fig:narrow_sigvar}.
We seek two different predictions of the labels as a function of $\hbh_{\min}\cdot\vbz(\vbx)$ : the labels $\hat{y}_{\mathrm{train}}$ that result from a finite training set and the labels $\hat{y}_{\mathrm{test}}$ that result from fitting to an average test set, or equivalently, the full data distribution (the limit of a training set of size $M\rightarrow\infty$ for fixed $N_f$ and $N_p$).

Given a training set consisting of $M$ data points $\mathcal{D}=\{(y_b, \vbx_b)\}_{b=1}^M$ with corresponding  hidden features $\vbz_a = \vbz(\vbx_a)$, we start by decomposing the kernel matrix into $n$ principal components $\hbh_i$ with non-zero eigenvalues $\sigma_i^2$,
\begin{align}
Z^TZ = \sum_{i=1}^n \sigma_i^2  \hbh_i\hbh_i^T.
\end{align}
We define the principal components so that they form an orthonormal basis of (hidden) features,
\begin{align}
 \hbh_i\cdot\hbh_j &= \delta_{ij}.
\end{align}
We define the minimum component $\hbh_{\min}$ as the principal component with the smallest non-zero eigenvalue $\sigma^2_{\min}$.
Next, we define the empirical variance of $\hbh_i\cdot \vbz_a$ (holding $\hbh_i$ fixed) within the training set and derive its relationship to the eigenvalue $\sigma_i^2$,
\begin{align}
\begin{split}
\Var[\vbx\in \mathcal{D} ]\qty[\hbh_i\cdot \vbz(\vbx)| \hbh_i ] &=  \frac{1}{M}\sum_{a=1}^M(\hbh_i\cdot \vbz_a)^2\\
&=  \frac{1}{M}\hbh_i^TZ^TZ\hbh_i\\
&= \frac{\sigma_i^2}{M}.\label{eq:mincomp_var}
\end{split}
\end{align}
Similarly, we define the empirical covariance of $\hbh_i\cdot \vbz_a$ and the labels $y_a$ (again, holding $\hbh_i$ fixed),
\begin{align}
\begin{split}
\Cov[\vbx\in \mathcal{D} ]\qty[\hbh_i\cdot \vbz(\vbx), y(\vbx) | \hbh_i] &= \frac{1}{M}\sum_{a=1}^M  (\hbh_i\cdot \vbz_a)y_a\\
&= \frac{1}{M}\hbh_i^TZ^T\vby.
\end{split}
\end{align}
Using the expression for the predicted labels in Eq.~\eqref{eq:student} and the exact solution for the fit parameters in the ridge-less limit in Eq.~\eqref{eq:exactpseudo}, we express the predicted label for an arbitrary test data point $\vbz'$ in terms of the empirical variance and covariance as
\begin{align}
\begin{split}
\hat{y} &\approx   \vbz'\cdot(Z^TZ)^+ Z^T\vby\\
& =\vbz'\cdot \sum_{i=1}^n \frac{1}{\sigma_i^2} \hbh_i  \hbh_i^T Z^T\vby\\
& = \sum_{i=1}^n \frac{\Cov[\vbx\in \mathcal{D} ]\qty[\hbh_i\cdot \vbz(\vbx), y(\vbx)| \hbh_i] }{\Var[\vbx\in \mathcal{D} ]\qty[\hbh_i\cdot \vbz(\vbx)| \hbh_i ]} (\hbh_i\cdot \vbz').
\end{split}
\end{align}
Dropping all terms except for the one containing the minimum component, we find the expression for the predicted labels as a function of  $\hbh_{\min}\cdot \vbz'$ resulting from fitting the training data,
\begin{align}
\hat{y}_{\mathrm{train}}\qty(\hbh_{\min}\cdot \vbz') &= \frac{\Cov[\vbx\in \mathcal{D}]\qty[\hbh_{\min}\cdot \vbz, y(\vbx)  | \hbh_{\min}] }{\Var[\vbx\in \mathcal{D}]\qty[\hbh_{\min}\cdot \vbz  | \hbh_{\min}]} \qty(\hbh_{\min}\cdot \vbz').
\end{align}
To find this relationship for an average test set, we extend the empirical variance and covariance to consider an infinitely large data set,
or equivalently, average over all possible data points $(y, \vbx)$ with hidden features $\vbz(\vbx)$.
However, we still hold $\hbh_{\min}$ fixed since it is a result of the training set.
The resulting relationship is
\begin{align}
\hat{y}_{\mathrm{test}}\qty(\hbh_{\min}\cdot \vbz') &= \frac{\Cov[\vbx]\qty[\hbh_{\min}\cdot \vbz(\vbx), y(\vbx) |\hbh_{\min}] }{\Var[\vbx]\qty[\hbh_{\min}\cdot \vbz(\vbx) |\hbh_{\min}]} \qty(\hbh_{\min}\cdot \vbz').
\end{align}

According to Eq.~\eqref{eq:mincomp_var}, we calculate the spread of the training data points along the minimum principal component,
\begin{equation}
\sigma_{\mathrm{train}}^2 = \Var[\vbx\in \mathcal{D} ]\qty[\hbh_{\min}\cdot \vbz(\vbx) | \hbh_{\min}] = \frac{\sigma_{\min}^2}{M}.
\end{equation}
We find that it is related to the minimum eigenvalue of the kernel matrix.
We also derive the true variance of data points along $\hbh_{\min}$ for an average test set,
\begin{equation}
\sigma_{\mathrm{test}}^2 = \Var[\vbx]\qty[\hbh_{\min}\cdot \vbz(\vbx)|\hbh_{\min} ].
\end{equation}
In the next few sections, we derive this variance, along with the covariance with respect an average test set, or the full data distribution, for both models, along with expressions for $\hat{y}_{\mathrm{test}}$. 

\subsection{Linear Regression}
In  linear regression without basis functions, the input features and hidden features are identical such that $Z=X$.
Using the decomposition of the labels in Eq.~\eqref{eq:SIydecomp}, we find
\begin{align}
\Cov[\vbx]\qty[\hbh_{\min}\cdot \vbz(\vbx), y(\vbx)|\hbh_{\min} ]  &= \frac{\sigma_X^2}{N_f}\hbh_{\min}^T\vbbeta\\
\Var[\vbx]\qty[\hbh_{\min}\cdot \vbz(\vbx)|\hbh_{\min} ]  &= \frac{\sigma_X^2}{N_f}.
\end{align}
Using these results, the predicted labels for an average test set as a function of $\hbh_{\min}\cdot \vbz'$ are then
\begin{align}
\hat{y}_{\mathrm{test}}\qty(\hbh_{\min}\cdot \vbz') &=  \hbh_{\min}^T\vbbeta\qty(\hbh_{\min}\cdot \vbz'),
\end{align}
while the expected spread is
\begin{equation}
\sigma_{\mathrm{test}}^2 = \frac{\sigma_X^2}{N_f}.
\end{equation}

\subsection{Random Nonlinear Features Model}

In the random nonlinear features model, we use the label decomposition in Eq.~\eqref{eq:SIydecomp} and the hidden feature decomposition in Eq.~\eqref{eq:SIzdecomp}, to find
\begin{align}
\Cov[\vbx]\qty[\hbh_{\min}\cdot \vbz(\vbx), y(\vbx_ |\hbh_{\min}]  &= \frac{\sigma_X^2}{N_f} \hbh_{\min}^TW^T \vbbeta\\
\Var[\vbx]\qty[\hbh_{\min}\cdot \vbz(\vbx) |\hbh_{\min} ] &= \frac{\sigma_X^2}{N_f}\hbh_{\min}^TW^TW\hbh_{\min} + \Delta\varphi\frac{\sigma_W^2\sigma_X^2}{N_p}.
\end{align}
The predicted labels are then
\begin{align}
\hat{y}_{\mathrm{test}}\qty(\hbh_{\min}\cdot \vbz') &=  \frac{ \hbh_{\min}^TW^T \vbbeta}{\qty(\hbh_{\min}^TW^TW\hbh_{\min} + \Delta\varphi\sigma_W^2\frac{\alpha_f}{\alpha_p} )}\qty(\hbh_{\min}\cdot \vbz')
\end{align}
and the expected spread is
\begin{equation}
\sigma_{\mathrm{test}}^2 =\frac{\sigma_X^2}{N_f}\hbh_{\min}^TW^TW\hbh_{\min} + \Delta\varphi\frac{\sigma_W^2\sigma_X^2}{N_p}.
\end{equation}

\section{Numerical Simulation Details}

In this section, we explain our procedures for generating numerical results.

\subsection{General Details}\label{sec:numdetails}

In all plots of training error, test error, bias, and variance, each point (or pixel for 2$d$ plots)  is averaged over $1000$ independent simulations, unless located exactly at a phase transition, in which case, each point is averaged over $150000$ simulations.
Small error bars are shown each plot, representing the error on the mean.
We also scale the error in each plot by the variance of the labels $\sigma_y^2 = \sigma_\beta^2\sigma_X^2 + \sigma_{\delta y^*}^2 +\sigma_\varepsilon^2$.
In all simulations, we use training and test sets of size $M=M'=512$,  a signal-to-noise ratio of $(\sigma_\beta^2\sigma_X^2 + \sigma_{\delta y^*}^2)/\sigma_\varepsilon^2 = 10$, and a regularization parameter of $\lambda=10^{-6}$.
We also use a linear teacher model $y^*(\vbx) = \vbx\cdot\vbbeta$ ($\sigma_{\delta y^*}^2 = 0$) in most cases.
In Fig.~\ref{fig:narrow_sigvar}, we use a nonlinear teacher model with $f(h) = \tanh(h)$. In this case, we find that $\expval*{f'} = 0.6057$ and $\expval*{f^2} = 0.3943$, resulting in ${\sigma_{\delta y^*}^2/\sigma_\beta^2\sigma_X^2 =  \Delta f = 0.0747}$.

To find the solution for a particular regression problem, we solve a different (but equivalent) system of equations depending on whether $N_p < M$ or $N_p > M$, allowing us to reduce the size of the linear system we need to solve.
If $N_p < M$, we solve the system of $N_p$ equations
\begin{align}
\qty[\lambda I_{N_p} + Z^TZ]\hbw &=  Z^T\vby
\end{align}
for the  $N_p$ unknown fit parameters $\hbw$ where $I_{N_p}$ is the $N_p\times N_p$ identity matrix. 
This equation is identical to that in Eq.~\eqref{eq:exact} in the main text.

Alternatively, if $N_p > M$ we solve a system of $M$ equations,
\begin{align}
\qty[\lambda I_M + ZZ^T]\hba &= \vby,
\end{align}
for the $M$ unknowns $\hba$ where $I_M$ is the $M\times M$ identity matrix.
We then convert to fit parameters via the formula $\hbw = Z^T \hba$.

\subsection{Bias-Variance Decompositions}

To efficiently calculate the ensemble-averaged bias and variance, we take inspiration from Eq.~\eqref{eq:biastwodatasets}.
During each simulation, we independently generate two training data sets $\mathcal{D}_1$ and $\mathcal{D}_2$.
Using the results from the first training set we calculate the training and test error. 
To calculate the bias, we also calculate the label predictions for both training sets for an identical test set, $\hby_1$ and $\hby_2$, and record the residual label errors between these predictions and the true labels of the test set $\vby^{*\prime}$ and record the product $(\hby_1-\vby^{*\prime})\cdot(\hby_2-\vby^{*\prime})$. When averaged over many simulations, this quantity approximates the bias. We can then subtract this quantity from the average test error to find the variance. We follow an analogous procedure to find each contribution of the labels in Eq.~\eqref{eq:SIydecomp}  to the bias and variance in Fig.~\ref{fig:narrow_sigvar}. This is achieved by calculating the test error, bias and variance using only a single contribution from the labels at a time and setting the rest to zero.

\subsection{Eigenvalue Decompositions of Kernel Matrices}

For each of the numerical eigenvalue distributions for the kernel matrices presented in the main text, we choose $M = 4096$.
We then average over the distributions for 10 independently sampled matrices when  $\alpha_p = 1$ or $\alpha_p = 8$ and over 80 matrices when  $\alpha_p = 1/8$.
In this way, we ensure that the same number of non-zero eigenvalues is present in the part of the histograms corresponding to the bulk of the distributions (the distribution excluding the delta function at zero).
For $M < N_p$ we calculate the eigenvalues of $Z^TZ$, while for $M > N_p$ we instead calculate the eigenvalues of $ZZ^T$ since this matrix is smaller and contains the same non-zero eigenvalues. In the later case, we then manally append an additional $N_p -M$ zero-valued eigenvalues to the distribution.

\subsection{Spread Along Mimimum Principal Components}

For each the scatter plots in Figs.~\ref{fig:narrow} and \ref{fig:narrow_sigvar}, 
we consider training and test sets of size $M=M'=200$, with all other parameters specified in Sec.~\ref{sec:numdetails}.
We then calculate the principal component corresponding to the minimum eigenvalue numerically and use this to plot the relationship learned by the model for the training set, as detailed in Sec.~\ref{sec:acc_mpc}.
For the test set, we show the relationship for an average test set rather than the specific test shown, again using the formulas detailed in Sec.~\ref{sec:acc_mpc}.
We note that these formulas still require the minimum principal component calculated for the training set.

In Fig.~\ref{fig:narrow}, the spread of the the training set compared to a test set as a function of $\alpha_p$ is calculated using 100 simulations for each point.
For each simulation, we record the ratio $\sigma_{\mathrm{train}}/\sigma_{\mathrm{test}}$ (see Sec.~\ref{sec:acc_mpc}) and then average this quantity across simulations for each $\alpha_p$.

\section{Complete Numerical Results}\label{sec:numerics}

In this section, we provide complete comparisons between the analytic and numerical results when lacking from the main text.
For linear regression, comparisons to numerical results for the training error, test error, bias ,and variance are depicted in Fig.~\ref{fig:linreg} in the main text. 

\begin{figure}[h!]
\centering
\includegraphics[width=\linewidth]{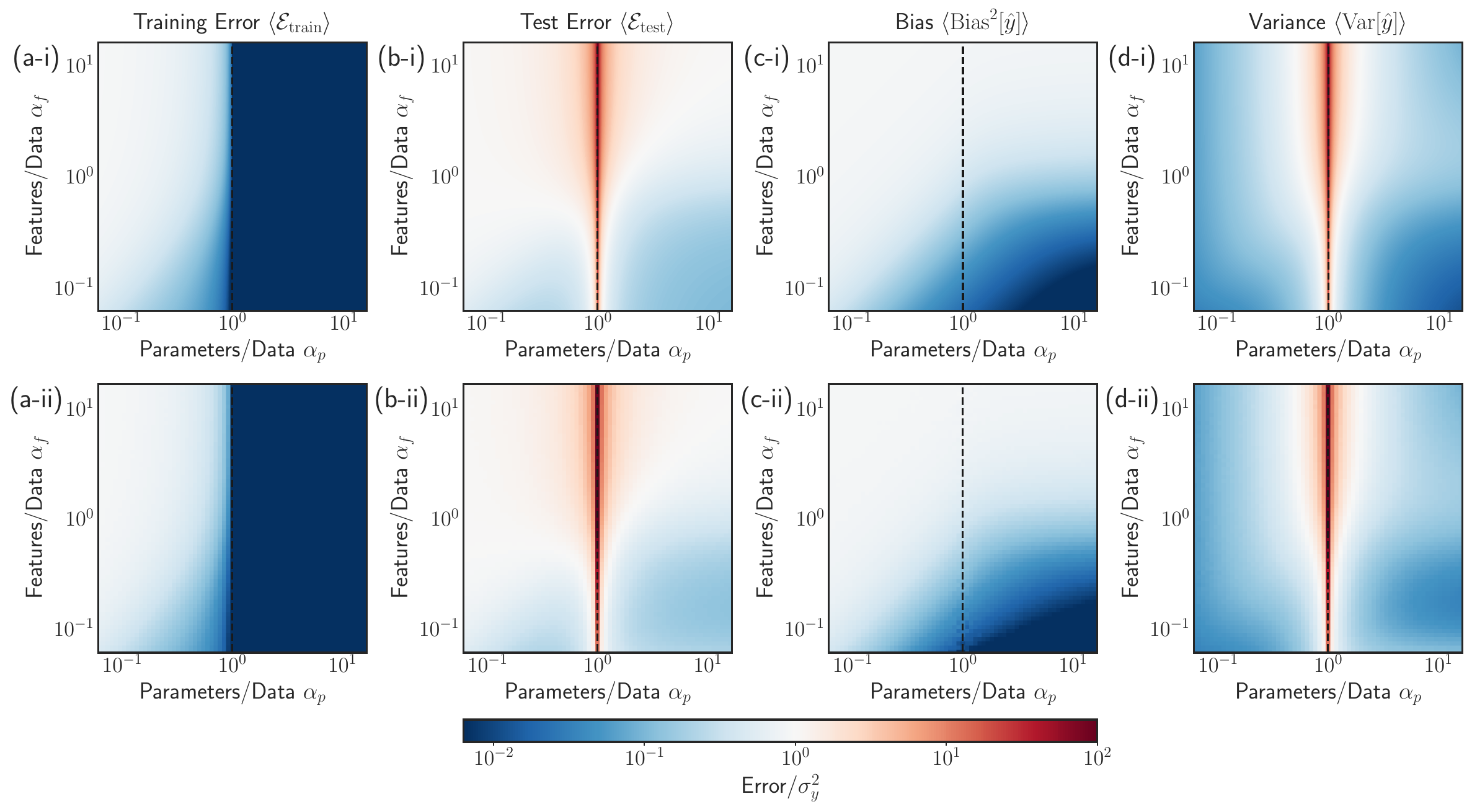} 
\caption{
{\bf Comparison of analytic and numerical results for the random nonlinear features model: Training error and bias-variance decomposition.}
(Top Row) Analytic solutions and (Bottom Row) numerical results are shown as a function of $\alpha_p=N_p/M$ and $\alpha_f=N_f/M$. 
Plotted are the ensemble-averaged (a) training error, (b) test error, (c) squared bias, and (d) variance. 
In each panel, a black dashed line marks the boundary between the under- and over-parameterized regimes at $\alpha_p = 1$. 
}\label{fig:rnlfm_numerics1}
\end{figure}

For the random nonlinear features model, Fig.~\ref{fig:rnlfm_numerics1} provides comparisons to numerical results for the training error, test error, bias, and variance.

%

\section{Non-standard Bias-Variance Decompositions}

\begin{figure}[h!]
\centering
\includegraphics[width=\linewidth]{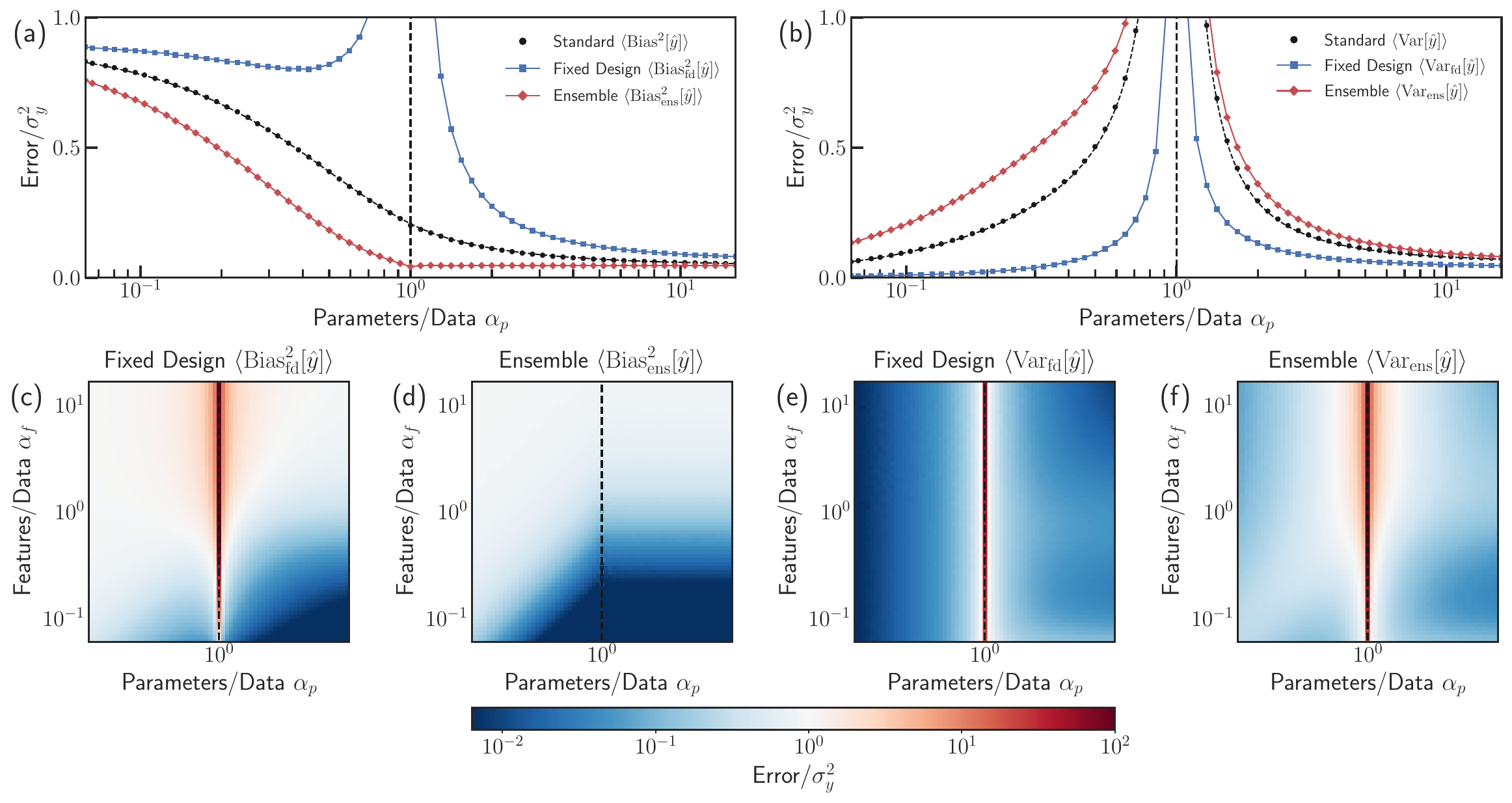} 
\caption{
{\bf Numerical comparison of the bias-variance decompositions using three different definitions.}
The {\bf (a)} squared bias and {\bf (b)} variance for the standard setting (black circles), fixed-design setting (blue squares) and ensemble setting (red diamonds) are shown for fixed $\alpha_f = 1/2$.
Results are also shown as a function of $\alpha_p=N_p/M$ and $\alpha_f=N_f/M$ for the {\bf (c)}  squared fixed design bias, {\bf (d)} squared ensemble bias, {\bf (e)}  fixed design variance, and {\bf (f)} ensemble variance. 
In each panel, a black dashed line marks the boundary between the under- and over-parameterized regimes at $\alpha_p = 1$. 
}\label{fig:bias_alt_defs}
\end{figure}

In Fig.~\ref{fig:bias_alt_defs}, we show numerical results for the alternative definitions of bias described in Sec.~VF for the random nonlinear features model.
In the fixed design setting, the bias and variance are defined as
\begin{align}
\begin{split}
\Bias_{\mathrm{fd}}[\hat{y}(\vbx)] &= \E[\vbeps][\hat{y}(\vbx)] - y^*(\vbx)\\
\Var[\mathrm{fd}][\hat{y}(\vbx)] &= \E[\vbeps][\hat{y}(\vbx)^2] - \E[\vbeps][\hat{y}(\vbx)]^2.
\end{split}
\end{align}
Alternatively, in the ensemble setting, the bias and variance are defined as
\begin{equation}
\begin{split}
\Bias_{\mathrm{ens}}[\hat{y}(\vbx)] &= \E[X, \vbeps, W][\hat{y}(\vbx)] - y^*(\vbx)\\
\Var[\mathrm{ens}][\hat{y}(\vbx)] &= \E[X, \vbeps, W][\hat{y}(\vbx)^2] -\E[X, \vbeps, W][\hat{y}(\vbx)]^2.
\end{split}
\end{equation}
We plot all four quantities with comparisons to the standard counterparts at fixed $\alpha_f$ in Figs.~\ref{fig:bias_alt_defs}(a) and (b).
We also show the full behavior as a function of both $\alpha_p$ and $\alpha_f$ for the four quantities in Figs.~\ref{fig:bias_alt_defs}(c)-(f).

